\documentclass[sn-basic,iicol]{sn-jnl}

\usepackage{color}
\usepackage{graphicx}  
\usepackage{float}  
\usepackage{subfigure}  

\jyear{2021}%
\theoremstyle{thmstyleone}%
\theoremstyle{thmstyletwo}%

\theoremstyle{thmstylethree}%

\begin{document}

\title[Article Title]{Simple, Effective and General: A New Backbone for Cross-view Image Geo-localization}


\author*{\fnm{Yingying} \sur{Zhu}}\email{zhuyy@szu.edu.cn}

\author{\fnm{Hongji} \sur{Yang}}

\author{\fnm{Yuxin} \sur{Lu}}

\author{\fnm{Qiang} \sur{Huang}}

\affil{\orgdiv{College of Computer Science and Software Engineering}, \orgname{Shenzhen University}, \orgaddress{Nanhai Ave 3688}, \city{Shenzhen}, \postcode{518060}, \state{Guangdong}, \country{China}}


\abstract{

In this work, we aim at an important but less explored problem of a simple yet effective backbone specific for cross-view geo-localization task. Existing methods for cross-view geo-localization tasks are frequently characterized by 1) complicated methodologies, 2) GPU-consuming computations, and 3) a stringent assumption that aerial and ground images are centrally or orientation aligned. To address the above three challenges for cross-view image matching, we propose a new backbone network, named Simple Attention-based Image Geo-localization network (SAIG). 


The proposed SAIG effectively represents long-range interactions among patches as well as cross-view correspondence with multi-head self-attention layers. The ”narrow-deep” architecture of our SAIG improves the feature richness without degradation in performance, while its shallow and effective convolutional stem preserves the locality, eliminating the loss of patchify boundary information.

Our SAIG achieves state-of-the-art results on cross-view geo-localization, while being far simpler than previous works. Furthermore, with only 15.9\% of the model parameters and half of the output dimension compared to the state-of-the-art, the SAIG adapts well across multiple cross-view datasets without employing any well-designed feature aggregation modules or feature alignment algorithms. In addition, our SAIG attains competitive scores on image retrieval benchmarks, further demonstrating its generalizability. As a backbone network, our SAIG is both easy to follow and computationally lightweight, which is meaningful in practical scenario. Moreover, we propose a simple Spatial-Mixed feature aggregation moDule (SMD) that can mix and project spatial information into a low-dimensional space to generate feature descriptors. In particular, SMD inherits the property of not being constrained by the strict assumption of model and further improves performance in cross-view tasks. The code is available at 
\href{https://github.com/yanghongji2007/SAIG}{https://github.com/yanghongji2007/SAIG} }

\keywords{Geo-localization, Cross-view image matching, Backbone, Feature aggregation}



\maketitle

\section{Introduction}\label{sec1}

The booming development of robot navigation, autonomous driving, and augmented reality has facilitated research on localization problems. Cross-view image geo-localization, which locates a ground view image by comparing it with a database of GPS-tagged aerial images and vice versa, is considered feasible to compensate for the unreliable positioning techniques, such as GPS. The task is extremely challenging due to drastic viewpoint changes and significant appearance changes between ground and aerial images.

\begin{figure}
\setlength{\abovecaptionskip}{0.2cm}
\centering
\includegraphics[height=5cm,width=7.159cm]{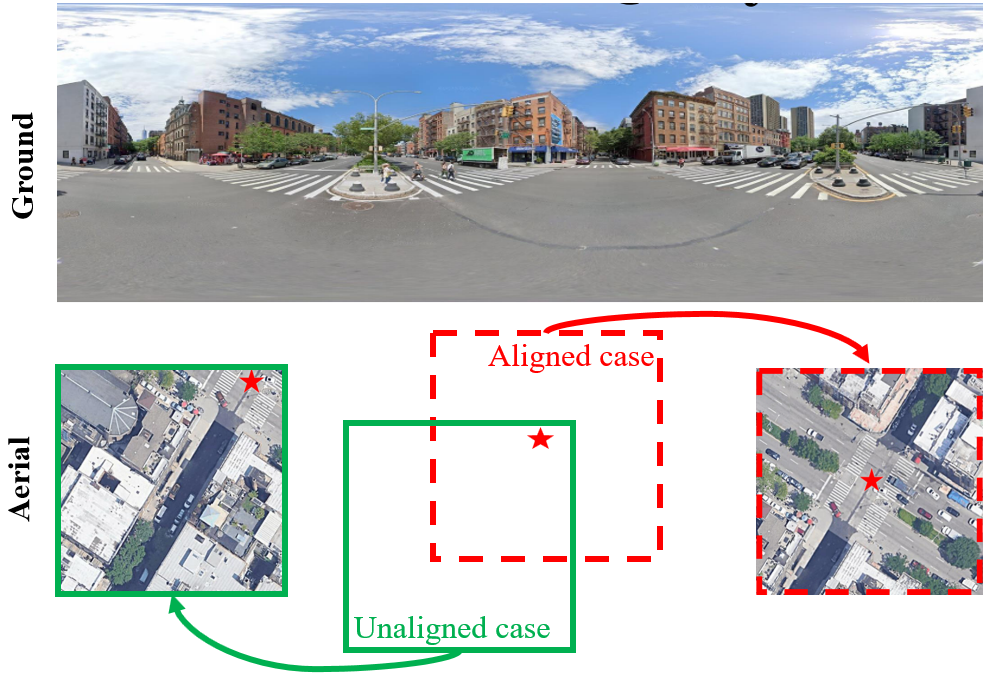}
\caption{Illustration of center-aligned image pairs. Existing studies mainly rely on a strong assumption that the query ground image must be exactly centered at the location of the aerial image. Thus, for one-to-one matching, the red border is considered as a correct match, while the green border is an incorrect match. In contrast, for one-to-many matching, both the red and green borders are correct matches.} 
\label{aligned image pairs}
\vspace{-12pt}
\end{figure}

There have been several attempts to develop large Siamese-like models with well-designed feature aggregation layers on top of a backbone network (e.g., VGG or ResNet) for cross-view geo-localization task, including CVM-Net \citep{hu2018cvm}, CVFT~\citep{CVFT2020}, GeoCapsNet~\citep{sun2019geocapsnet}, GeoNet~\citep{zhu2021geographic} etc. Moreover, previous work mainly rely on strong assumptions: ground panorama images are center-aligned~\citep{SAFA2019,DSM2020,wang2021LPN} as shown in Figure \ref{aligned image pairs} and orientation aligned~\citep{liu2019lending} with aerial images. Obviously, these assumptions largely hamper the performance of the methods in the practical cross-view geo-localization. Although L2LTR~\citep{L2LTR2021} and TransGeo~\citep{zhu2022transgeo}, the transformer-based models that can work without such strong assumptions, they lead to large memory and expensive computation when fitted into the Siamese-like dual-branch architecture.
Additionally, existing works mainly focus on one-to-one matching. That is, each ground query corresponds to only one reference aerial image. Nevertheless, in practical scenarios, one-to-many matching is more pervasive and practicable (Figure \ref{aligned image pairs}), since perfectly aligned one-to-one correspondence is hard to guarantee. 

Thus, instead of designing complicated networks for higher accuracy on specific datasets, how to explore a simple yet effective backbone network that can be adapted to more image geo-localization scenarios is more desirable and promising now.
In this work, we aim at an important but less explored problem of a simple and effective backbone specific for cross-view geo-localization task. So far, this problem has not been fully addressed. CNN-based models generally are unable to overcome the shortcoming that heavily rely on strong assumption, and Transformer-based models are limited by complex structures and computational costs. 

To this end, we propose a Simple Attention-based Image Geo-localization backbone (SAIG) consisting of an overlapping convolutional stem, a self-attention-based module, and a global pooling layer. Motivated by the favorable performance of the L2LTR model~\citep{L2LTR2021} in cross-view geo-localization tasks, we introduce the attention mechanism in feature extraction. The attention mechanism is skilled at capturing long-range information, which proves to be crucial for feature matching. Meanwhile, inspired by \citep{T2T-ViT2021}, we design a "narrow-deep" architecture that improves the feature richness without degradation in performance. In this manner, our network can bridge the domain gap and correspond to global features without additional techniques (i.e., polar transform algorithm or well-designed feature aggregation layers) that impose the strong assumptions mentioned before. Besides, we adopt a convolutional stem (conv stem)~\citep{earlyconv2021} on images to split them into overlapping patches and finally flatten the patches to aggregate them together. Specifically, the conv stem contains six convolutional layers and a linear projection layer. This shallow but effective conv stem enables our network to gather local features into patch embeddings with overlapping receptive fields to maintain locality and to avoid the loss of patchify boundary information. Comprehensive experiments are provided to verify that our SAIG can effectively learn cross-view features without additional techniques. Highly simple but effective, our network achieves superior or competitive performance with only 15.90\% of the model parameters and half of the output dimension compared to the state-of-the-art. Besides, it presents the potential of scalability and produce much better performance with various tricks. Moreover, inspired by NetVLAD \citep{hu2018cvm} and SAFA \citep{SAFA2019}, we design a new trainable aggregation layer for cross-view geo-localization. The layer is readily pluggable into our proposed SAIG architecture and achieve current state-of-the-art performance.


The contributions of this work can be summarized as follows:
\begin{itemize}
\item We propose a simple, effective and general backbone for cross-view geo-localization. It achieves favorable performance without incorporating other well-designed feature aggregation layers and feature alignment algorithms. Our highly compact model also performs competitively with only 15.9\% of the model parameters and half of the output dimension compared to existing methods.

\item We propose a new spatial-mixed feature aggregation module (SMD), which can efficiently aggregate local features in our network, facilitating the model training process and generating more discriminative descriptors.

\item We devise two losses (i.e., Triplet loss with semi-hard sample mining and InfoNCE loss) to improve our SAIG in one-to-many scenario. It alleviates the problem that traditional online batch hardest sample mining often lead to a collapse model. We show experimentally that above losses in our SAIG are better than the origin triplet loss in one-to-many dataset.

\item We conduct comprehensive experiments and ablation studies to demonstrate the simplicity and effectiveness of the SAIG and the SMD in various cross-view geo-localization and image retrieval settings. And our SAIG achieves competitive performance on six benchmarks compared to state-of-the-art methods, and is easy to follow.
\end{itemize}

\begin{figure*}
\centering
\includegraphics[height=8cm,width=13.085cm]{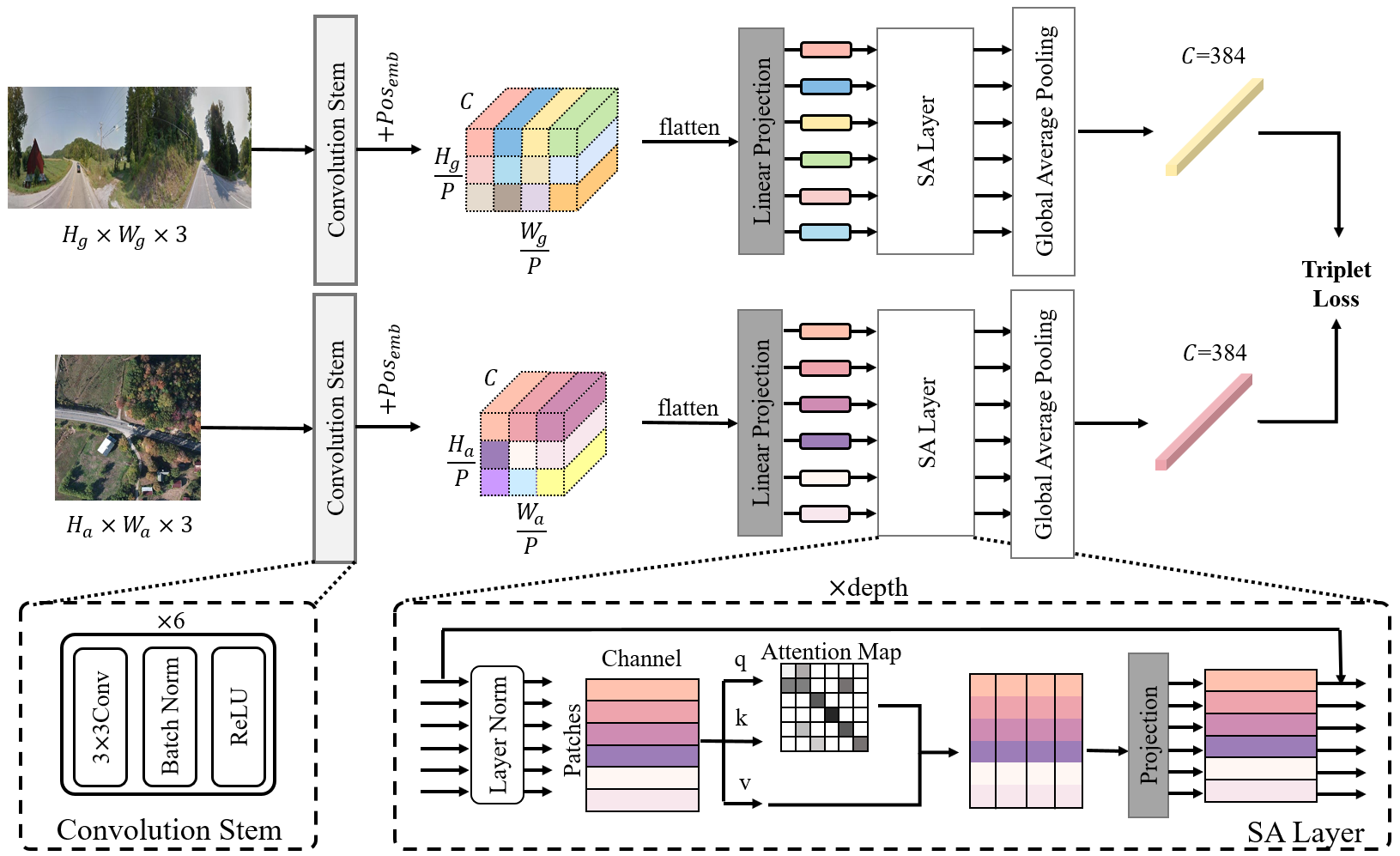}
\caption{\textbf{Overall structure of SAIG.} 
The network applies a Siamese-like architecture (no weight-shared) for extracting features from the two views. The convolutional stem captures some low-level features of each input and then projects each pixel to obtain the "$Patch\times Channel$" patch-based representation. These patches are further fed into the stacked SA layers and finally processed by global average pooling. \textbf{Bottom left:} A convolutional stem contains six layers of $3\times3$ convolution with Batch normalization and ReLU non-linearity. \textbf{Bottom right:} An SA layer contains layer norm, a self-attention module, and a linear projection, building the global relationship among patches.} 
\label{model_architecture}
\vspace{-3pt}
\end{figure*}

\section{Related Work}\label{sec2}
For locating a ground image from satellite database, most of the existing works draw on techniques from image retrieval. Many different models are designed to extract discriminative global features or to reduce the domain gap in cross-view geo-localization tasks. Several works investigate the effect of specific losses on cross-view image matching. In this section, we first present an overview of the development of the models in cross-view geo-localization and then review the specific loss designed for this task.

\subsection{CNN-based method}
\bmhead{Deep Features}
The first line of works aims at {\bf generating discriminative deep learning representations.} In the early stage, the hand-crafted features are commonly used in the cross-view image matching \citep{bansal2011geo,senlet2011framework,senlet2012satellite, viswanathan2014vision}. However, these methods with hand-crafted features fail to unify the vast differences in appearance between ground view images and aerial view images. Inspired by the success of deep learning on vision tasks, some effect are taken to introduce convolution neural networks into cross-view image matching. \cite{workman2015location} first apply CNN features for cross-view image matching and retrieval, and show its superiority over hand-crafted features. Then, \cite{workman2015wide} further improve the performance by performing model training on aerial branch using multi-scale images. \cite{vo2016localizing} introduce a novel soft-margin triplet loss to Siamese network and Triplet network for cross-view geo-localization. \cite{CVUSA} proposed a weakly supervised network to learn the semantic layout as the final image descriptors of satellite images. \cite{cai2019reweightingloss} integrate channel and spatial attention modules into a residual network trained with a Hard Exemplar Reweighting triplet loss, which is capable of highlighting salient features in both views.  Nevertheless, the inherent local properties of CNN networks hinder the analysis and extraction of geospatial information in images. As a result, these methods fail to generate descriptors discriminative enough to handle drastic viewpoint changes.
\bmhead{Feature Alignment}
Another line of studies aims at {\bf explicitly exploiting image or feature correspondence}. To this end, they focus on learning viewpoint shifting or aligning features to mitigate the domain discrepancy in an explicit manner. \cite{liu2019lending} and \cite{DSM2020} take the orientation of ground-view images as a complementary information for alignment. CVFT~\citep{CVFT2020} inserts a feature transport layer in the ground branch to learn feature transformation. Most notably, SAFA applies an efficient polar transform algorithm to warp aerial images to align them with ground images. LPN~\citep{wang2021LPN} adopts a square-ring feature partition strategy based on prior alignment of the datasets with neighbor areas as auxiliary information. \cite{regmi2019bridging} and \cite{coming2021} attempt to leverage the features learned in cross-view synthesis for retrieval. They utilize the generative adversarial networks (GANs) to synthesize the corresponding representation of a specified view from another. However, the majority of these efforts are exploited on the centre-aligned ground panoramas and aerial image pairs, and thus fail to generalize to the geo-localization beyond one-to-one retrieval~\citep{VIGOR2021}.

\subsection{Transformer-based method}
Inspired by the success of vision transformer, L2LTR~\citep{L2LTR2021} and TransGeo ~\citep{zhu2022transgeo} employ Transformer blocks for modeling global dependencies. Although such methods alleviate the heavy reliance on explicit feature alignment, it commonly comes with large computational cost and GPU-memory. Besides, TransGeo also introduces Sharpness-Aware Minimization (SAM) ~\citep{foret2020sharpness} to achieve the state-of-the-art results.
 
\subsection{Feature Aggregation}
To produce a more discriminative representation, there are some work focus on the design of feature aggregation. \cite{hu2018cvm} incorporate a dual-branch VGG16 network with independent NetVLAD layers to better encode local features. The feature representations from both views are embedded into a joint embedding space for further matching. ~\cite{sun2019geocapsnet} combines ResNet with Capsule layers for modeling high-level semantics. SAFA \citep{SAFA2019} employs a multi-head attention module as the feature aggregation layer to encode spatial information.


\subsection{Retrieval loss}
Metric learning via deep networks are highly related to cross-view image matching task. Contrast loss \citep{varior2016gated} and triplet loss \citep{schroff2015facenet}, which are widely used in image retrieval tasks, can attract similar (positive) satellite images and dispel different (negative) satellite images for ground view images. To decreasing the intra-class variation during training, the quadruplet \citep{chen2017beyond} and angular losses \citep{wang2017deep} are proposed to improve the models. The soft-margin triplet loss proposed by \cite{vo2016localizing} is effective in cross-view geo-localization. To improve the convergence, ~\cite{hu2018cvm} propose a weighted soft-margin triplet loss which scales the distances by a coefficient $\alpha$. Besides, the hard triplet loss with batch hard negative mining \citep{hermans2017defense} is also used to exclude the interference of massive easy negative samples, which may lead to a low generalization ability. In particular, there are certain potential positive samples in one-to-many dataset, which the traditional batch hardest sample mining strategy fails. In our work, we further investigate the losses of intrinsically focusing on hard samples in one-to-many dataset.

\section{Methodology}\label{sec3}

\subsection{Backbone Overview}\label{overview}
The overall structure of the SAIG backbone is shown in Figure \ref{model_architecture}. As Siamese-like architecture has been proved to be more adaptive to appearance variations of cross-view image pairs, we employ a domain-specific Siamese-like architecture following previous works~\citep{hu2018cvm, SAFA2019,liu2019lending,CVFT2020}.
The backbone processes images from a specific viewpoint (or domain), and extracts high-level features from the images, and projects these features into a shared embedding space. 
In each branch of the Siamese-like network, there is a {\bf convolutional stem} to extract local feature (see Sec. \ref{conv stem}), several {\bf multi-head self-attention layers} to model long-range relationship (see Sec. \ref{MSA}) and a {\bf global pooling layer} to generate final descriptors. 

For a training set ${D_{tr}}=\left\{\left(I^{g}_{1}, I^{a}_{1}\right), \ldots,\left(I^{g}_{N}, I^{a}_{N}\right)\right\}$ of $N$ image pairs, an image pair $\left(I^{g}_{i}, I^{a}_{i}\right)$ contains a query ground image $I^{g}_{i}$ and one single ground-truth aerial image $ I^{a}_{i}$ $\left(1 \leq i \leq N\right)$.
Take the ground view image $I^{g}_{i}$ as an example. $I^{g}_{i}$ is fed to the conv stem to get a feature embedding, which is attached with a learnable position embedding. After that, the MSA module (which will be discussed in Sec. \ref{MSA}) captures long-range interactions among patches. Finally, we apply a feature aggregation layer to acquire a ground descriptor $F^{g}_{i}$. The aerial branch network generates a descriptor $F^{a}_{i}$ analogously.

\subsection{Conv Stem}\label{conv stem}

We refer to the initial layers of a conventional CNN network as the stem, which is used to downsample an input image to an appropriate feature map size. The vanilla ViT~\citep{ViT2021} and MLP-Mixer~\citep{MLP-Mixer2021} employ a patch stem that performs stride-$p$ $p\times{p}$ non-overlapping convolution to images. The patch-like architecture that contains three unique properties: 1) features are decomposed into fixed-sized patches, 2) there is no downsampling operation after the stem, i.e., the channel dimension is fixed, and 3) the spatial and channel information are processed separately. The appealing performance of ViT and MLP-mixer indicates that the patch-like architecture is a robust template for deep learning. 

In spite of the effective patch stem, it ignores the local structures, making it difficult for the model to learn local features (e.g., detecting edge or corner), leading to a significant loss of patchify boundary information in visual perception. 
Unlike the ViT and MLP-mixer models, we employ a convolutional stem (conv stem) to produce patches with overlapping regions and introduce an inductive bias inherent to CNN for modeling invariance. In this way, our patch-based embedding features not only contain overlapping regions, which benefits the contextual connection and preserves the integrity of patch boundary features, but also introduce the inductive bias, resulting in less training data. Although using the majority of ResNet-50 as a stem (called ResNet stem) in the “hybrid ViT” model is an attractive alternative for gaining significant performance improvements, this stem also brings a large number of parameters. In contrast, the conv stem provides the best trade-off between performance and model size. Thus, a well-designed, lightweight yet effective convolution stem can work perfectly well for the cross-view task.

Concretely, our conv stem consists of 6 convolutional layers with kernel size of $3\times3$ followed by a batch normalization and ReLU to gradually downsample input images. At the end of the conv stem, a fully-connected layer is employed for feature projection. The strides of each CNN layer are [2, 2, 1, 2, 1, 2] and the output channels of conv stem and fully-connected layer are [64, 128, 128, 256, 256, 512] and [384], respectively. For an input image with size ${224 \times 224}$, the conv stem transforms it into $14\times14$ patches correspondingly. These patches will be fed into the SA layers with learnable position embedding.

\subsection{Multi-head Self-attention}\label{MSA}

Inspired by the success of self-attention~\citep{Transformer2017} that directly model long-range interactions among patches, we introduce the self-attention mechanism into our SAIG. We perform a multi-head self-attention module (MSA) to project patch features into different subspaces to capture information. Each layer contains a LayerNorm and a multi-head self-attention sub-layer. Given a sequence of $d$-dimensional input \{${x_1,x_2,...,x_p}$\}, the self-attention function (regarded as a head) applies three independent transformation to project the input into query, key, and value, respectively:
\begin{equation}
Z^{'}_{l} = LayerNorm(X_{l})\label{LN}
\end{equation}
\begin{equation}
Q = W_qZ^{'}_{l}, K = W_kZ^{'}_{l}, V=W_vZ^{'}_{l}\label{qkv}
\end{equation}
\begin{equation}
Z_l = Vsoftmax(\frac{QK^T}{\sqrt{d/H}})\label{score}
\end{equation}
where $W_q$, $W_k$, $W_v$ are the weight of the "query", "key", and "value", respectively. $H$ denotes as the number of heads and $X_{l}$ means the input feature of the layer $l$. 

In the SAIG, we do not employ the feed-forward network (FFN) that typically acts as a sublayer in the ViT or MLP-Mixer for "channel mixing”. First, the FFN is expensive to fit in the Siamese-like architecture. Second, with tremendous parameters, the FFN requires data augmentation and regularization techniques to avoid overfitting. Finally, although the FFN layer has exhibited its superiority in several models, we find it is less critical in cross-view geo-localization tasks, because of a minor performance gain while coming with large parameters. We also speculate the reason that the self-attention layers are capable of modeling global interactions from overlapping patch features without FFN. As a result, we remove the FFN layer to make a trade-off between performance and computational budget.

\bmhead{The effectiveness of SAIG for cross-view correspondence} Most previous works employ CNN to extract image features. However, the convolution operation ignores the global context which is robust to drastic viewpoint changes. As a result, CNNs have to rely on the strong assumptions, well-designed feature aggregation layers to compensate for this limitation.
In contrast, our SAIG utilizes the MSA layer, which excels in calculating relationships among patches and capturing non-local dependencies. Benefiting from this self-attention mechanism, the SAIG can extract critical global information without other well-designed feature aggregation modules or feature alignment (e.g., polar transform). The essential global information captured by the self-attention layer further facilitates the network to learn cross-view correspondence and achieves competitive performance compared to the state-of-the-art. CNNs are more aware of the adjacent patches or pixels, while SAIG can directly model the relationship among the patches by MSA. In addition, the global receptive field of SAIG can help focus on matching essential visual cues, which are often neglected by CNN-based models.

\begin{figure}
\centering
\includegraphics[height=3.5cm,width=7.2cm]{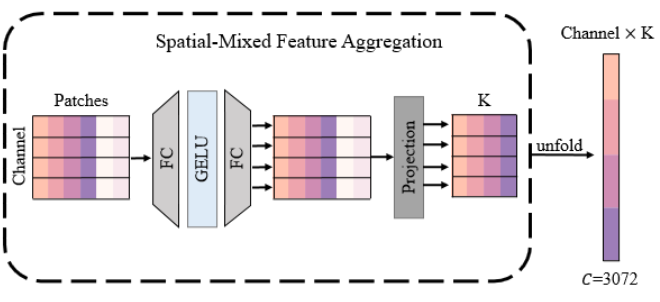}
\caption{Spatial-mixed feature aggregation module} 
\label{pool_layer}
\vspace{-12pt}
\end{figure}

\subsection{Spatial-mixed Feature Aggregation Module}
With some approaches demonstrating the validity of feature aggregation layers on cross-view tasks (e.g., NetVLAD, SAFA), we observe that most feature aggregation layers possess the following properties: 1) learnable pooling layer, 2) aggregated descriptors with increased length, and 3) operating along the spatial dimension to keep the network matching images without alignment. Since our backbone is effective and the images matching without alignment is taken into account, a simple aggregation layer, Spatial-Mixed feature aggregation Module (SMD), is presented for cross-view tasks. Specifically, our SMD first mixes spatial information by two fully-connected layers (first expanded to 4 times, then reduced to input size) with GELU. Given a local feature input $X$:


\begin{equation}
F^{'} = W_2 \sigma (W_1(X^T))
\end{equation}
Here $W_1$ and $W_2$ denote the weights of fully-connected layer and $\sigma$ denotes the GELU nonlinearity. Then we project $F^{'}$ into \emph{K}-d space through one fully-connected layer and unfold it along \emph{K} dimension to produce the final descriptor $F$.

\begin{equation}
F^K = W_3 F^{'}
\end{equation}

\begin{equation}
F = \{f^1_1, ..., f^1_c, f^2_1, ..., f^2_c, ..., f^K_1, ..., f^K_c\}
\end{equation}

First, in contrast to "self-attention" module, the spatial mixer freely model long-range information without any additional supervision. Superficially, this mixer is more like unconstrained inter-patch attention, allowing the free-flow of the information. Second, instead of utilizing global pooling to generate the final descriptor, we apply a fully-connected layer along the spatial dimension, allowing our model to freely learn how to aggregate critical information. Meanwhile, SMD expands the length of the descriptor, and the local features are projected into K-dim space rather than shrinking abruptly to a narrow feature embedding space in aggregation. The latter may leads to excessive feature loss at training. Consequently, our SMD facilitates the training for a better representation. Note that our SMD is a plug-and-play feature aggregation module and improves performance by replacing the GAP in original SAIG backbone.

\subsection{SAIG Variants}

To deeply investigate our backbone, we design two variants with different number of SA layers: SAIG-Shallow (SAIG-S) and SAIG-Deep (SAIG-D). Following the "narrow-deep" architecture design, SAIG-D contains 22 SA layers to extract deep features representations. Besides, we provide the shallower version, called SAIG-S, which has 11 SA layers in each branch and offers considerable performance at lower cost. Both variants have the same conv stem, which has been discussed in Sec. \ref{conv stem}. To create a lightweight model for both variations, the hidden dimension is set to 384 and the number of heads in an MSA layer is set to 12. Note that when exploring our backbone (SAIG), we adopt only global average pooling to explores the superior performance of SAIG as a backbone for handling cross-view tasks. To this end, we do not use any well-designed feature aggregation module to avoid its possible booster.


\subsection{Loss Function}\label{triplet loss}
Cross-view geo-localization can be cast as an image retrieval task. Thus, the triplet loss, which aims to attract positive samples and dispel negative samples, is widely used in this task. Following the previous works, we adopt a weighted soft-margin triplet loss~\citep{hu2018cvm} that can be expressed as:
\begin{equation}
L_{triplet} = log(1+e^{\alpha(d_{pos}-d_{neg})}) \label{loss}
\end{equation}
where $d_{pos}$ and $d_{neg}$ denote the $l2$ distance of the positive pairs and negative pairs in same embedding space, respectively. $\alpha$ is a hyperparameter used to speed up convergence, and the value is chosen empirically.

However, the VIGOR benchmark often comes with a large number of easy negative examples and thus induce no loss. It is difficult for the models to focus on the key differences among hard examples. Therefore, we first introduce the semi-hard negative mining strategy that proposed by FaceNet~\citep{schroff2015facenet}, which yields a fairly hard but not too hard example and thus does not lead to collapsed models~\citep{wu2017sampling}. We present its soft-margin format in this paper, which is given by
\begin{equation}
d^{*}_{neg} = \mathop{\arg\min}_{d_{pos}<d_{neg}} d_{neg} \label{semihardloss}
\end{equation}

\begin{equation}
L_{semi-triplet} = log(1+e^{\alpha(d_{pos}-d_{d^{*}_{neg}})}) \label{semihardloss1}
\end{equation}
where $d^{*}_{neg}$ is the $l2$ distance of the semi-hard negative pairs.



In light of the recent development in image retrieval, we introduce the infoNCE loss into one-to-many scenario for the first time and the experiments (see Table \ref{VIGOR}) show that the introduced loss is beneficial to improve the performance of SAIG in such scenarios.
The goal of the infoNCE loss is to construct sufficient noisy samples in a batch to ensure that the models can distinguish positive ones from them. In addition, the infoNCE loss can potentially mine the hard samples by adjusting the temperature, rather than being explicitly to modifying the inputs. 

 \begin{equation}
 L_{infoNCE} = -log\frac{\exp(S_{i,i}/\tau)}{\sum^N_{k\neq i} \exp(S_{i,k}/\tau) + \exp(S_{i,i}/\tau)} \label{infoNCEloss}
 \end{equation}
where $S_{i,k}$ denote the $l2$ distance between i-th and k-th instances in a batch $N$ and $\tau$ is the temperature.

\section{Experiment}\label{sec4}

In this work, we intend to seek a backbone with a combination of advantages, which is reflected in simplicity, effectiveness and generalization, not just in the best performance. To this end, we not only evaluate the proposed methods on the basis of SAIG on several well-known benchmark datasets compared with other methods, but also perform comprehensive experiments to verify our SAIG backbone in various cross-view geo-localization settings and image retrieval tasks. Additionally, we conduct experiments to demonstrate our proposed SMD and losses compared with existing aggregation approaches and loss fucncitons.
We give the detail of datasets, metrics, experiment implementation in the following sections. Extensive experiments demonstrate that our SAIG brings consistent and significant performance improvements in different cross-view geo-localization settings for a wide range of cross-view matching tasks.

\subsection{Dataset and Metric}

To verify the simplicity, effectiveness and generalizability of our backbone network, we evaluate it on four well-known cross-view geo-localization benchmarks: CVUSA~\citep{CVUSA}, CVACT~\citep{liu2019lending}, VIGOR~\citep{VIGOR2021}, and University-1652~\citep{2020University-1652}. Besides, experiments conducted on $\mathcal{R}$Oxford and $\mathcal{R}$Paris~\citep{RoxfordAndParis} further illustrate the generalizability on retrieval tasks.  


\bmhead{CVUSA and CVACT} Image pairs in CVUSA and CVACT are ground panoramas and satellite images from USA and Canberra, and the correct match of a ground image is a corresponding aerial image. They both consist of 35,532 image pairs for training. The CVUSA dataset provide 8,884 image pairs for testing, and the CVACT contains 8,884 pairs for validation (denoted as CVACT\_val ) for standard cross-view geo-localization. Additionally, to support fine-grained city-scale geo-localization, the CVACT also provides 92,802 image pairs with accurate geo-tags for testing (denoted as CVACT\_test). Namely, for the CVACT\_test, a retrieved aerial image is considered correct as long as it is within the distance d=5m from the ground-truth location of the ground image.
There exists a strong assumption that the center of aerial images is exactly aligned with the corresponding ground view images. We refer to this assumption as the assumption of alignment (AA). 

\bmhead{VIGOR} VIGOR is a novel cross-view benchmark that retrieving beyond one-to-one matching and without the assumption of alignment, which gathers 90,618 ground panorama images and 105,124 aerial images. It makes sure that each aerial image has no more than 2 positive panoramas and around 4\% of the aerial images have no corresponding panoramas. To evaluate the capability of cross-view geo-localization, VIGOR calculates the IOU (Intersection Over Union) of all positive reference aerial images with the exactly aligned aerial images. A positive reference image has an IOU greater than 0.39, and the IOU of a semi-positive image is between $ [{1}/{7},{9}/{23}]$.


\bmhead{University-1652} University-1652 is a drone-based cross-view geo-localization benchmark, which selects satellite-view images and drone-view images from 1652 buildings in 72 universities. The training set has 50218 images in total, which contains 701 buildings and each building contains multiple drone-view images and one satellite view image. This type of benchmark introduces two new tasks: drone-view target geo-localization and drone navigation. Unlike other corss-view benchmarks, University-1652 can be regarded as a classification task, so the overall framework and loss are slightly different from the above datasets. Note that the assumption of alignment still exists in University-1652.

\bmhead{ILSVRC-2012 ImageNet} It selects 1000 high-level object categories from ImageNet, with 1.3 million training images, 50,000 validation images and 100,000 test images without overlapping. In this work, we follow standard transfer learning to pre-train SAIG on ILSVRC-2012 ImageNet-1k dataset and then fine-tune on the downstream tasks.

\bmhead{$\mathcal{R}$Oxford and $\mathcal{R}$Paris} 
They revisit the classical Oxford5k and Paris6k landmark datasets and are both image retrieval benchmarks, which addressing the issues including annotation errors, the size of the datasets, and the level of challenge. Most previous image retrieval methods conduct experiments on $\mathcal{R}$Oxford and $\mathcal{R}$Paris. In particular, $\mathcal{R}$Oxford and $\mathcal{R}$Paris present new annotations, introduce 15 new challenging queries and divide both datasets into Easy (E), Medium (M), and Hard (H) difficulties according to three new protocols. The revisited $\mathcal{R}$Oxford and $\mathcal{R}$Paris datasets comprising 4993 and 6322 images respectively, and are widely used to evaluate the image retrieval models.

\bmhead{Metric}
Similar to the prior works~\citep{CVUSA,hu2018cvm,SAFA2019}, we deploy the recall accuracy at top-$K$ (known as r@K) as our evaluation metric. For a query ground view image, r@K measures whether the retrieved top-K images contain the corresponding satellite image. Besides, r@1 is more challenging than other metrics (e.g. r@5, r@10, etc.), and it requires that the model must return the corresponding satellite image as the first retrieved image. For the VIGOR dataset, the hit rate is also introduced according to ~\citep{VIGOR2021}, which considers the retrieval at top-1 accuracy when both positive and semi-positive samples are considered as correct results. For image classification and image retrieval, top-1 accuracy and mAP (mean average precision) are adopted for evaluation.

\subsection{Implementation Detail}

\bmhead{Pre-training on ImageNet-1k} Our models are pre-trained on ImageNet-1k. We follow ~\citep{DeiT} for most data augmentation and apply EMA (for inference, we found it facilitates convergence in the early stages). The model is trained with an AdamW optimizer, with a batch size of 512 and a weight decay of 0.01. We use an initial learning rate of 0.001, with a linear warm-up and cosine decay schedule. Our models are implemented in PyTorch and are trained on 4 V100 GPUs for 300 epochs in four to six days.


\bmhead{Fine-tuning on Cross-view dataset} To fit the pre-trained model into a cross-view geo-localization task, we remove the final classifier module and reserve the feature extraction part. Besides, a Siamese-like dual-branch architecture is introduced to extract features from images of different views. For fine-tuning, we use AdamW with cosine learning rate decay. The initial learning rate is set to 0.0001 and weight decay is set to 0.03, with a batch size of 32 and a grad clipping at global norm 1. The $\alpha$ in Triplet loss is set to 10 and the $\tau$ in InfoNCE loss is set to 0.02, respectively. Our models are trained for 200 epochs on CVUSA and CVACT datasets, and 90 epochs on VIGOR large-scale dataset. If not mentioned otherwise, the resolution of ground and aerial images is $128\times512$ and $256\times256$ for CVUSA and CVACT, while $320\times640$ and $320\times320$ for VIGOR, respectively. Besides, we compare the result on VIGOR with different loss and training strategy. If SAM is enabled, the rho is set to 2. For University-1652, we follow ~\citep{wang2021LPN} but remove the first fully-connected layer in classifier module. Our models are trained for 120 epochs using SGD optimizer with a momentum of 0.9 and a weight decay of 0.005. The learning rate is set to 0.01 for the new layers and 0.001 for the backbone, and decayed by 0.1 after 80 epochs. 

\bmhead{Fine-tuning on Image Retrieval dataset} To further investigate the capability of the proposed model, experiments are also conducted on the image retrieval task. We fine-tune SAIG-D on the SFM120k dataset with the same setting and optimization strategies as in ~\citep{TrainViTforImageRetri}. For a fair comparison, all the models are trained for 100 epochs with images in fixed-size of $224\times224$ and $384\times384$ using Adam optimizer. 

\subsection{Experiment for Cross-view Geo-localization}

\subsubsection{Comparison with State-of-the-art Models}
\bmhead{Compared Methods} To verify the generalizability of our SAIG, we compare it with the previous methods on following cross-view geo-localization benchmarks: standard cross-view geo-localization~\citep{CVUSA,liu2019lending}, fine-grained cross-view geo-localization~\citep{liu2019lending}, geo-localization beyond one-to-one matching~\citep{VIGOR2021}, drone-based geo-localization ~\citep{2020University-1652} and localizing with unknown orientation and limited FoV~\citep{DSM2020}. In terms of these compared methods that combine with extra feature aggregation layers or well-designed modules, CVM-Net \citep{hu2018cvm} encodes global descriptors by the NetVLAD aggregation layer to counteract viewpoint variance and gains performance from weighted soft-margin ranking loss. Liu and Li~\citep{liu2019lending} lend a trainable orientation embedding into feature maps for guiding the alignment of different viewpoints. SAFA~\citep{SAFA2019} and ~\cite{DSM2020} exploit the assumption of central alignment and further reduce the domain gap across views by polar-transformation. To better encode spatial information, SAFA~\citep{SAFA2019} also applies attention mechanism for spatial-aware aggregation. CVFT \citep{CVFT2020} inserts a feature transport module to convert the features from ground-view to aerial-view for matching. LPN~\citep{wang2021LPN} proposes a feature partition strategy to mitigates the visual discrepancy on multi-view images matching. \cite{coming2021} leaves the issue of bridging the gap between different views to cross-view image synthesis. \cite{L2LTR2021} and \cite{zhu2022transgeo} employ Transformer-based models to overcome the shortcomings of previous methods that rely heavily on strong assumptions, but are still limited by complex model structures and high computational costs. To fairly compare the above methods, we retain the assumption-based SAIG models and compare them to the assumption-free SAIG. Note that TransGeo~\citep{zhu2022transgeo} employs an additional training method SAM \citep{foret2020sharpness} to mitigate overfitting of Transformer. For a fair comparison, we include SAM in part of our experiments and verify its effectiveness on our SAIG. For those compared methods, where not specifically stated, we directly cite the reported results from their papers. For ease of reference, these methods are indicated with their corresponding references in Table \ref{ref}.

\begin{table}[ht]\scriptsize
\resizebox{1.05\linewidth}{!}{
\begin{tabular}{|l|c|l|c|}
\hline
\bf{Methods} & \bf{References} & \bf{Methods} & \bf{References} \\
\hline
CVM-Net & \cite{hu2018cvm} & Toker & \cite{coming2021}\\
CVFT & \cite{CVFT2020} & LPN & \cite{wang2021LPN} \\ 
SAFA & \cite{SAFA2019} & L2LTR & \cite{L2LTR2021} \\ 
Liu\&Li & \cite{liu2019lending} & VIGOR & \cite{VIGOR2021}\\ 
Shi & \cite{DSM2020} &  TransGeo  &\cite{zhu2022transgeo} \\ 
Zheng & \cite{2020University-1652} & & \\
\hline

\end{tabular}
}
\caption{The references of compared methods}
\label{ref}
\vspace{-10pt}
\end{table}

\begin{table*}\scriptsize
\centering
\begin{tabular}{lcccccccccc}
\multirow{3}{*}{Models}& & & \multicolumn{4}{c}{CVUSA}&\multicolumn{4}{c}{CVACT\_val}\\ \cmidrule(lr){4-7} \cmidrule(lr){8-11} 
~ &Backbone& Dim& r@1 & r@5& r@10& r@1\%&  r@1 & r@5& r@10& r@1\%\\ 
 ~ &~ & ~& (\%) & (\%) & (\%) & (\%)& (\%) & (\%) & (\%) & (\%) \\
\toprule
\multicolumn{5}{l}{\emph{(A) w/o assumption}}&&&&&&\\
CVM-Net & VGG16 & 4096 &22.47&49.98&63.18&93.62&20.15&45.00&56.87&87.57\\
CVFT& VGG16 & 4096 & 61.43&84.69&90.49&99.02&61.05&81.33&86.52&95.93\\
SAFA& VGG16 & 4096 & 81.15&94.23&96.85&99.49&78.28&91.60&93.79&98.15\\
Liu \& Li& VGG16 & 1536 &40.79&66.82&76.36&96.12&46.96&68.28&75.48&92.01\\
L2LTR~& HybridViT  & 768 & 91.99 & 97.68&98.65&99.75&83.14&93.84&95.51&98.40\\ 
TransGeo~& DeiT-S/16  & 1000 & {94.08} & 98.36&99.04&99.77&84.95&94.14&95.78&98.37\\ 
\toprule
SAIG-S & SAIG-S  & \textbf{384} & 88.82 & 97.17 & 98.27 & 99.74 &81.39 &93.88&95.53&98.44 \\
SAIG-D & SAIG-D  & \textbf{384} & 90.29 & 97.71 & 98.74& 99.76 & 82.40 &93.94&95.54& 98.49 \\  
SAIG-S + SAM & SAIG-S  & \textbf{384} & 92.69 & 98.13 & 98.95 & 99.84 & 85.39 & 95.09 & 96.52 & 98.53 \\
SAIG-D + SAM & SAIG-D  & \textbf{384} & {93.97} & {98.47} & {99.09} & \textbf{99.86} & {86.65} & {95.25} & {96.53} & {98.61} \\
\toprule
SAIG-S + SMD & SAIG-S  & 3072 & 91.77 & 97.60 & 98.60 & 99.64 & 83.54  & 93.89 & 95.53 & 98.47 \\
SAIG-D + SMD & SAIG-D  & 3072 & {92.71}  & {97.92}  & {98.89} & 99.71 & {84.42} & {94.09} & {95.57} & {98.49} \\   
SAIG-S + SMD + SAM & SAIG-S  & 3072 & {95.37}  & {98.68}  & \textbf{99.25} & {99.85} & {88.44} & {95.61} & {96.80} & {98.73} \\
SAIG-D + SMD + SAM& SAIG-D  & 3072 & \textbf{96.08}  & \textbf{98.72}  & {99.22} & \textbf{99.86} & \textbf{89.21} & \textbf{96.07} & \textbf{97.04} & \textbf{98.74} \\
\midrule
\multicolumn{5}{l}{\emph{(B) w assumption}}&&&&&&\\
Polar-SAFA& VGG16  & 4096 &89.84&96.93&98.14&99.64&81.03&92.80&94.84&98.17\\
Shi& VGG16  & 4096 &91.93&97.50&98.54&99.67&82.49&92.44&93.99&97.32\\
Toker& U-net  & 4096 &92.56&97.55&98.33&99.57&83.28&93.57&95.42&98.22\\
LPN& VGG16  & 4096 &92.83&98.00&98.85&99.78&83.66&94.14&95.92&98.41\\
L2LTR & HybridViT  & 768 & 94.05 & 98.27&98.99&99.67&{84.89}&94.59&95.96&98.37\\ 
\toprule
SAIG-S  & SAIG-S & \textbf{384} & 92.21 & 98.26 & 99.16 & 99.84 &81.53 &94.38&96.33&98.68 \\
SAIG-D & SAIG-D & \textbf{384} & 93.87 & 98.60 & {99.30} & {99.85} & 83.34 & {95.18} & {96.54} & {98.75}\\

SAIG-S + SAM & SAIG-S & \textbf{384} & 93.39 & 98.45 & 99.10 & 99.83 & 84.50 & 94.84 & 96.25 & 98.58 \\
SAIG-D + SAM & SAIG-D & \textbf{384} & {94.90} & {98.82} & {99.38} & 99.84 & {85.95} & {95.18} & 96.41 & 98.58 \\ 
\toprule
SAIG-S + SMD & SAIG-S & 3072 & 94.45 & 98.51 & 99.18 & 99.83 & 83.94 & 94.93 & 96.47 & 98.69 \\
SAIG-D + SMD  & SAIG-D & 3072 &{95.00}&{98.66}&99.20&99.83&84.71&{95.25}&{96.68}&\textbf{98.89}\\ 
SAIG-S + SMD + SAM  & SAIG-S & 3072 &{95.91}&{98.96}& {99.43} & {99.84} &{88.34}&{95.80}&{96.80}&{98.83}\\
SAIG-D + SMD + SAM & SAIG-D  & 3072 &\textbf{96.34}&\textbf{99.10}& \textbf{99.50} & \textbf{99.86} &\textbf{89.06}&\textbf{96.11}&\textbf{97.08}&\textbf{98.89}\\
\bottomrule
\end{tabular}
\caption{Comparison with state-of-the-art methods on CVUSA and CVACT. The assumption of alignment denotes that if the query image centre at the reference image. According to this geometric assumption, some recent works~\protect\citep{SAFA2019,wang2021LPN} has applied polar-transformation or feature-level partition strategy to achieve the state-of-the-art performance. However, our proposed method is a strong backbone for cross-view task and exhibits competitive performance even without this assumption of the alignment (the polar-transform strategy is employed as the way to enables this assumption of alignment).}
\label{CVUSA and CVACT}
\end{table*}

\bmhead{Standard Cross-view Geo-localization} 

We compare our SAIG with state-of-the-art methods on CVUSA and CVACT datasets from two perspectives.

On one hand, with or without the assumption of alignment, using our backbone (e.g., SAIG-S or SAIG-D) alone outperforms most of compared methods that are combined with some kind of feature aggregation algorithm or aggregation layers. 
For the case without the assumption of alignment, with similar parameters, SAIG-D backbone gains more than 9\% improvement at r@1 compared to the best-performed CNN method SAFA~\citep{SAFA2019}. For more information about parameters, please refer to Table \ref{model complexity}. 
Moreover, our SAIG-D outperforms TransGeo by 1.7\% at r@1 on the CVACT dataset with fewer parameters and shorter dimension after adopting the same training strategy SAM as TransGeo.
Although there is a narrow margin of 0.74\% at r@1 between SAIG-D (82.40\%) and L2LTR (83.14\%) on CVACT, SAIG-D is simpler and more compact than L2LTR, with only 15.9\% of the model parameters and half of the output dimension. It verifies our SAIG is a strong but simple backbone for cross-view tasks. 
For the case with the assumption of alignment, SAIG-D backbone shows remarkable performances on both CVUSA (93.87\% at r@1) and CVACT (83.34\% at r@1). It demonstrates that the proposed SAIG backbone can still benefit from such additional alignment algorithm from the previous works ~\citep{SAFA2019,DSM2020}. 
Particularly, only trained with SAM, our SAIG-D backbone surpasses these compared methods and achieves state-of-the-art results in all recall metrics on two benchmarks.
Meanwhile, the superiority in terms of model size and dimensionality further illustrates our simplicity of backbone, which facilitates storage and retrieval speed. Results in Table \ref{CVUSA and CVACT} verify the effectiveness of SAIG, which attributes to MSA module in learning global information for cross-view image feature correspondence. 

On the other hand, combining with the feature aggregation layers or training strategy (i.e., SAIG+SMD or SAIG+SMD+SAM), SAIG achieves the competitive performance compared to state-of-the-art across two datasets whether or not to use the assumption of alignment.
The proposed SMD enables SAIG to achieve better performance, e.g., compared to the SAIG-D backbone, SMD achieves 2.42\% and 2.02\% performance gains at r@1 on CVUSA and CVACT without alignment, respectively, versus 1.17\% and 1.37\% performance gains at r@1 with alignment. It suggests that our SMD can aggregate more discriminative global features to facilitate the backbone in cross-view tasks. Additionally, SAM mitigates the gap between assumption-free and assumption-based retrieval for SAIG. With SAM, our networks (SAIG+SMD+SAM) without alignment even outperform those models using assumption on CVUSA and CVACT. The network (SAIG-D+SMD+SAM) can reach the r@1 accuracy of 96.08\% on CVUSA and 89.21\% on CVACT without alignment, as well as the r@1 accuracy of 96.34\% on CVUSA and 89.06\% on CVACT with alignment, respectively. To the best of our knowledge, they are new state-of-the-art results. 

\begin{table}[ht]
\centering
\begin{tabular}{lcccc}
\toprule
\multirow{3}{*}{Models}& \multicolumn{4}{c}{CVACT\_test}\\ \cmidrule(lr){2-5} 
~ &  r@1 & r@5& r@10& r@1\% \\ 
~ &  (\%) & (\%) & (\%) & (\%) \\ \midrule
CVM-Net & 5.41 & 14.79 & 25.63 & 54.53 \\
Liu \& Li & 19.21 & 35.97 & 43.30 & 60.69 \\
CVFT & 26.12 & 45.33 & 53.80 & 71.69 \\
SAFA & 55.50 & 79.94 & 85.08 & 94.49 \\
Shi & 35.63 & 60.07 & 69.10 & 84.75 \\
L2LTR & 60.72 & 85.85 & 89.88 & 96.12 \\
\midrule
SAIG-S & 50.67 & 82.79 & 88.61 & 96.26 \\
SAIG-D & 53.30 & 84.35 & 89.44 & 96.27 \\
\midrule
SAIG-S+SMD & 56.40 & 85.48 & 90.00 & 96.54 \\
SAIG-D+SMD & 57.61 & 85.93 & 90.34 & 96.55 \\
SAIG-S+SMD+SAM & 65.77 & 88.66 & 91.83 & 96.67 \\
SAIG-D+SMD+SAM & \textbf{67.49} & \textbf{89.39} & \textbf{92.30} & \textbf{96.80} \\

\bottomrule
\end{tabular}
\caption{Comparison with state-of-the-art methods on CVACT\_test.  }
\vspace{-3pt}
\label{CVACT_test}
\end{table}

\bmhead{Fine-grained Cross-view Geo-localization} To evaluate the generalizability and effectiveness of our model, we verify the SAIG on the fine-grained cross-view geo-localization task. Specifically, the models are evaluated on a more challenging large-scale CVACT\_test dataset. The experimental results are showed in Table \ref{CVACT_test}. 

Even compared to these existing methods that incorporated with extra feature aggregation layers or well-designed modules, our SAIG-D and SAIG-S, as the backbone, surpass most compared methods, respectively. In the case of r@1\%, our SAIG-S and SAIG-D achieves state-of-the-art performance.

Besides, SMD further boost the r@1 accuracy by 5.73\% and 4.31\% on the top of SAIG-S and -D respectively. 
Our network (SAIG-D+SMD+SAM) exceeds the state-of-the-art methods in all recall metrics, especially by almost 7\% at r@1.
These results further demonstrate the excellent generalizability of our SAIG.

\begin{table*}[htb]\scriptsize
\setlength{\abovecaptionskip}{0.2cm}
\centering
\begin{tabular}{l|c|cccc|cccc}
\toprule
\multirow{3}{*}{Models} & \multirow{3}{*}{Loss} & \multicolumn{4}{c|}{Same-Area}&\multicolumn{4}{c}{Cross-Area}\\ \cmidrule(lr){3-6} \cmidrule(lr){7-10} 
~ & ~ & r@1 & r@5& r@1\%& Hit Rate &  r@1 & r@5& r@1\%&Hit Rate\\ 
~ & ~ &  (\%) & (\%) & (\%) & (\%)& (\%) & (\%) & (\%) & (\%) \\\midrule
Siamese-VGG & Triplet &18.59&43.64&97.55&21.90&2.77&8.61&62.64&3.16 \\
SAFA & Triplet & 33.93&58.42&98.24&36.87&8.20&19.59&77.61&8.85 \\ 
VIGOR & Hybrid & 41.07&65.81&98.37&44.71&11.00&23.56&80.22&11.64 \\ 
TransGeo& Triplet & 61.48&87.54&98.56&73.09&18.99&38.24&88.94&21.21 \\ 
\midrule
SAIG-S & Triplet & 40.38 & 70.29 & 98.97& 47.57& 10.22 &23.93 & 81.32& 11.61\\
SAIG-D & Triplet& 42.15 & 72.59& 99.13& 50.08&11.88 & 26.98& 84.43& 13.23\\
\midrule
SAIG-S + SMD   & Triplet& 45.92 & 74.26 & 99.22 & 52.26 & 14.50 & 30.80 & 85.63 & 15.86 \\
SAIG-D + SMD &Triplet & 51.50 & 78.67& 99.37& 58.87&17.58 & 35.99& 89.21& 19.32\\
SAIG-D + SMD & InfoNCE & 55.37 & 81.16& 99.50& 62.70& 23.47 & 43.57 & 91.19 & 25.97 \\
SAIG-D + SMD &Semi-Triplet & 55.60 & 81.63& 99.43& 63.57& 22.35 & 42.43 & 90.83 & 24.69 \\
SAIG-S + SMD + SAM & Triplet& 57.57 & 82.45 & 99.63 & 64.56 & 25.44 & 46.30 & 93.26 & 27.53 \\
SAIG-S + SMD + SAM & Semi-Triplet & 62.28 & 85.92 & 99.66 & 70.21 & 30.14 & 52.23 & 93.87 & 33.34 \\
SAIG-D + SMD + SAM &Triplet & 61.27 & 85.69& 99.68& 69.26& 27.61 & 48.87 & 93.14 & 30.37 \\
SAIG-D + SMD + SAM &InfoNCE & 62.92 & 86.67 & 99.66 & 71.10 & 32.77 & 55.52 & 94.60 & 36.11 \\
SAIG-D + SMD + SAM &Semi-Triplet & \textbf{65.23} & \textbf{88.08}& \textbf{99.68}& \textbf{74.11}&\textbf{33.05} & \textbf{55.94}& \textbf{94.64}& \textbf{36.71}\\
\bottomrule

\end{tabular}
\caption{Comparison with state-of-the-art methods on VIGOR. The assumption of alignment is not available on the VIGOR dataset, and thus most methods using alignment are not applicable. Note that we do not adopt the hybrid loss that proposed by \protect\cite{VIGOR2021}.}
\label{VIGOR}
\end{table*}

\bmhead{Geo-localization Beyond One-to-one Matching} Experiments are conducted on the newly proposed VIGOR dataset, which is a more challenging benchmark that many previous methods on one-to-one matching fail to adapt to this dataset. As reported in Table \ref{VIGOR}, SAIG-D backbone gains the competitive performance without using the hybrid loss that ~\cite{VIGOR2021} proposed and is effective in such a one-to-many matching scenario. Our SAIG-D improves the hit rate by 5.37\% in the Same-Area testing and 1.59\% in the Cross-Area testing respectively compared to the second best method ~\citep{VIGOR2021}. These results verify that our SAIG effectively captures significant global information, since the geographical areas of these images is not overlapping. 

To better investigate the scalability of our SAIG backbone, we deploy the above proposed SMD for SAIG, as a new network to conduct experiments on VIGOR. SMD brings clear performance gains on two SAIG variants (5.54\% on SAIG-S and 9.35\% on SAIG-D in Same-Area at r@1). 
Furthermore, with SAM, our network (SAIG-D+SMD+SAM) outperforms TransGeo by 3.75\% in Same-Area and by 14.06\% in Cross-Area at r@1.
This demonstrates the robustness of our proposed SAIG, which not only attains favorable scores on its own, but also produces the state-of-the-art results. 

Besides, we found that on VIGOR, different losses have a significant impact on our SAIG network. Compared to the vanilla triplet loss, InfoNCE loss brings an improvement of up to 3.87\%, and Semi-Triplet loss gains up to 4.1\%. It reveals that getting the models to focus specifically on hard samples can better improve the performance in one-to-many image matching.

 

\begin{table*}\scriptsize
\setlength{\abovecaptionskip}{0.2cm}
\centering
\begin{tabular}{lccccccc}
 & & & &\multicolumn{4}{c}{University\-1652} \\\cmidrule{5-8}
Models & Backbone & Params & Dim & \multicolumn{2}{c}{Drone $\rightarrow$ Satellite} & \multicolumn{2}{c}{ Satellite $\rightarrow$ Drone} \\
 & & (M) & & R@1 & AP & R@1 & AP \\ \midrule
\emph{(A) w/o assumption of alignment} & & & & && &\\
Zheng & ResNet-50 &24.9 & 512 &58.23&62.91&74.47&59.45 \\
SAIG-S \textbf{[ours]}& SAIG-S & \textbf{18.2} &\textbf{384} &74.70&77.99&83.59&73.22 \\
SAIG-D \textbf{[ours]}& SAIG-D&31.2 &\textbf{384} &\textbf{75.31}&\textbf{78.52}&\textbf{86.31}&\textbf{75.82} \\\midrule
\emph{(B) w assumption of alignment} &&&&& && \\
LPN& ResNet-50 & 29.2 & 2048 & 75.93&79.14&\textbf{86.45}&74.79 \\
SAIG-S \textbf{[ours]}& SAIG-S &\textbf{18.2} &\textbf{1536}&76.78&79.84&86.02&77.18 \\
SAIG-D \textbf{[ours]}& SAIG-D&31.2 &\textbf{1536}&\textbf{78.85}&\textbf{81.62}&\textbf{86.45}&\textbf{78.48} \\
\bottomrule
\end{tabular}
\caption{Comparison with state-of-the-art methods on University\-1652. Note that this benchmark does not utilize the Siamese-like architecture nor the triplet loss, but is regarded as a classification task. Therefore, we only consider SAIG as the backbone to replace the original ResNet-50.}
\label{University1652}
\end{table*}

\bmhead{Drone-based Geo-localization} We also compare the proposed SAIG with other competitive approaches on the University-1652 dataset, which offers the drone-view images and the geo-tagged satellite-view images for drone localization. 
With fewer obstacles, the drone-view images can provide more reasonable information compared to ground-view panoramas, which are easily obscured by trees or buildings.
In this experiment, the feature-level partition strategy~\citep{wang2021LPN} is considered as enabling the assumption of alignment. As shown in Table \ref{University1652}, SAIG-D outperforms the LPN by 2.48\% in AP for the drone-view target localization task (Drone $\rightarrow$ Satellite), and by 3.69\% for the drone navigation task (Satellite $\rightarrow$ Drone). Moreover, the proposed SAIG models get the best performance in the experiment without the assumption of alignment. For example, compared to ~\cite{2020University-1652}, SAIG-S and SAIG-D increase by 15\% and 16\% in AP, respectively. These results further validate the generalizability of SAIG for various cross-view geo-localization tasks.

\begin{table*}[htb]\scriptsize
\setlength{\abovecaptionskip}{0.2cm}
\centering
\setlength{\tabcolsep}{0.5mm}{
\begin{tabular}{c|cccc|cccc|cccc|cccc}
\toprule
\multirow{3}{*}{Models}&\multicolumn{4}{c|}{FoV=$360^{\circ}$}&\multicolumn{4}{c|}{FoV=$180^{\circ}$}&\multicolumn{4}{c|}{FoV=$90^{\circ}$}&\multicolumn{4}{c}{FoV=$70^{\circ}$}\\ \cmidrule{2-5} \cmidrule{6-9} \cmidrule{10-13} \cmidrule{14-17} 
~&r@1 &r@5 &r@10 &r@1\% &r@1 &r@5 &r@10 &r@1\% &r@1 &r@5 &r@10 &r@1\% &r@1 &r@5 &r@10 &r@1\%\\
~&(\%) &(\%)&(\%)&(\%)&(\%)&(\%)&(\%)&(\%)&(\%)&(\%)&(\%)&(\%) &(\%)&(\%)&(\%)&(\%)\\ \midrule
CVM-Net &16.25 &38.86 &49.41 &88.11 &7.38 &22.51 &32.63 &75.38 &2.76 &10.11 &16.74 &55.49 &2.62 &9.30 &15.06 &21.77 \\
CVFT &23.38& 44.42& 55.20& 86.64& 8.10& 24.25& 34.47& 75.15& 4.80& 14.84& 23.18& 61.23& 3.79& 12.44& 19.33& 55.56 \\
Shi &\textbf{78.11}& 89.46& 92.90& 98.50& 48.53& 68.47& 75.63& 93.02& 16.19& 31.44& 39.85& 71.13& 8.78& 19.90& 27.30& 61.20 \\
L2LTR & - & - & - & - & \textbf{56.69} & \textbf{80.86}& \textbf{87.75}& \textbf{98.01}& \textbf{26.92}& \textbf{50.49}& \textbf{60.41}& \textbf{86.88}& 13.95& 33.07& 43.86& 77.65 \\
SAIG-D&71.95&\textbf{90.15}&\textbf{94.03}&\textbf{99.18}&{52.45}&{78.11}&{85.83}&{97.71}&{26.71}&{50.24}&{59.82}&{86.63}&\textbf{20.90}&\textbf{41.40}&\textbf{51.23}&\textbf{80.38} \\ \hline
\end{tabular}}
\caption{Comparisons with existing methods for localizing ground images with unknown orientation and limited field of view on CVUSA.}
\label{FoV}
\end{table*}

\bmhead{Geo-localization in Limited FoV with Unknown Orientation} Additionally, we evaluate some methods on the CVUSA dataset in a more realistic localization scenario, where the ground images have a limited Field-of-View (FoV) view rather than a panoramic view. Since the FoV view contains less semantic information compared to the panoramic view, it is more challenging to match the ground-level FoV view with satilite-level view.
Those methods that use the alignments or require orientation information like ~\citep{liu2019lending} can not be applied to this setting, while the performances of these methods that are applicable for the scenario drop significantly.
In Table \ref{FoV}, the images with limited FoV and unknown orientation are followed the same settings as \citep{DSM2020}. Clearly, the results show our method extensively outperforms other methods, with better performance improvement on a narrower FoV. Remarkably, when in the 70° FoV, our method obtains a 2.38$\times$ improvement compared with \citep{DSM2020} on CVUSA in terms of r@1. The experiments verify that our SAIG can benefit from the MSA module, reducing the ambiguity of geospatial information and achieving better performance under a narrower FoV condition.

\begin{table}[ht]
\setlength{\abovecaptionskip}{0.2cm}
\centering
\begin{tabular}{cc|c|c}
\toprule
Models & Backbone & GFLOPs & \#Params (M) \\\midrule
CVM-Net &\multirow{6}{*}{VGG16}& - & 160.3\\
Liu \& Li & ~ & - & 30.7\\
CVFT & ~ & - & 26.8\\
SAFA& ~ & 40.2 & 29.5 \\ 
Shi& ~ & 39.3 & \textbf{14.7} \\
LPN & ~ & 40.2 & 29.5 \\ \midrule
L2LTR& HybridViT & 57.1 & 195.9 \\
TransGeo& DeiT-S/16 & 12.3 & 44.9 \\
\midrule
SAIG-S & SAIG-S & \textbf{8.8} & 18.2 \\
SAIG-D & SAIG-D & 13.3 & 31.2 \\\bottomrule
\end{tabular}
\caption{Comparison of FLOPs and parameters with several methods in cross-view geo-localization. All results are computed in a Siamese-like architecture. “GFLOPs” is calculated under the input size of 256$\times$256 or 128$\times$512 in PyTorch.}
\label{model complexity}
\vspace{-3pt}
\end{table}

\subsubsection{Model Analysis}

\bmhead{Model Complexity} We compare FLOPs and parameters with other existing methods using different backbone in cross-view geo-localization. Since we use PyTorch to calculate FLOPs, those methods implemented with TensorFlow are not computed (marked with $'-'$ in the Table \ref{model complexity}). As shown in Table \ref{model complexity}, our model is more compact with less parameters and FLOPs, which embodies the simplicity of our SAIG backbone. This is attributed to the patch representation obtained by conv stem and the capability of modeling global relationships through the self-attention layers.

\begin{figure*}[htbp] 
\begin{minipage}[t]{0.335\linewidth} 
\centering

\subfigure[CVUSA]{
\label{pool examples:A}
\includegraphics[width=2.3in]{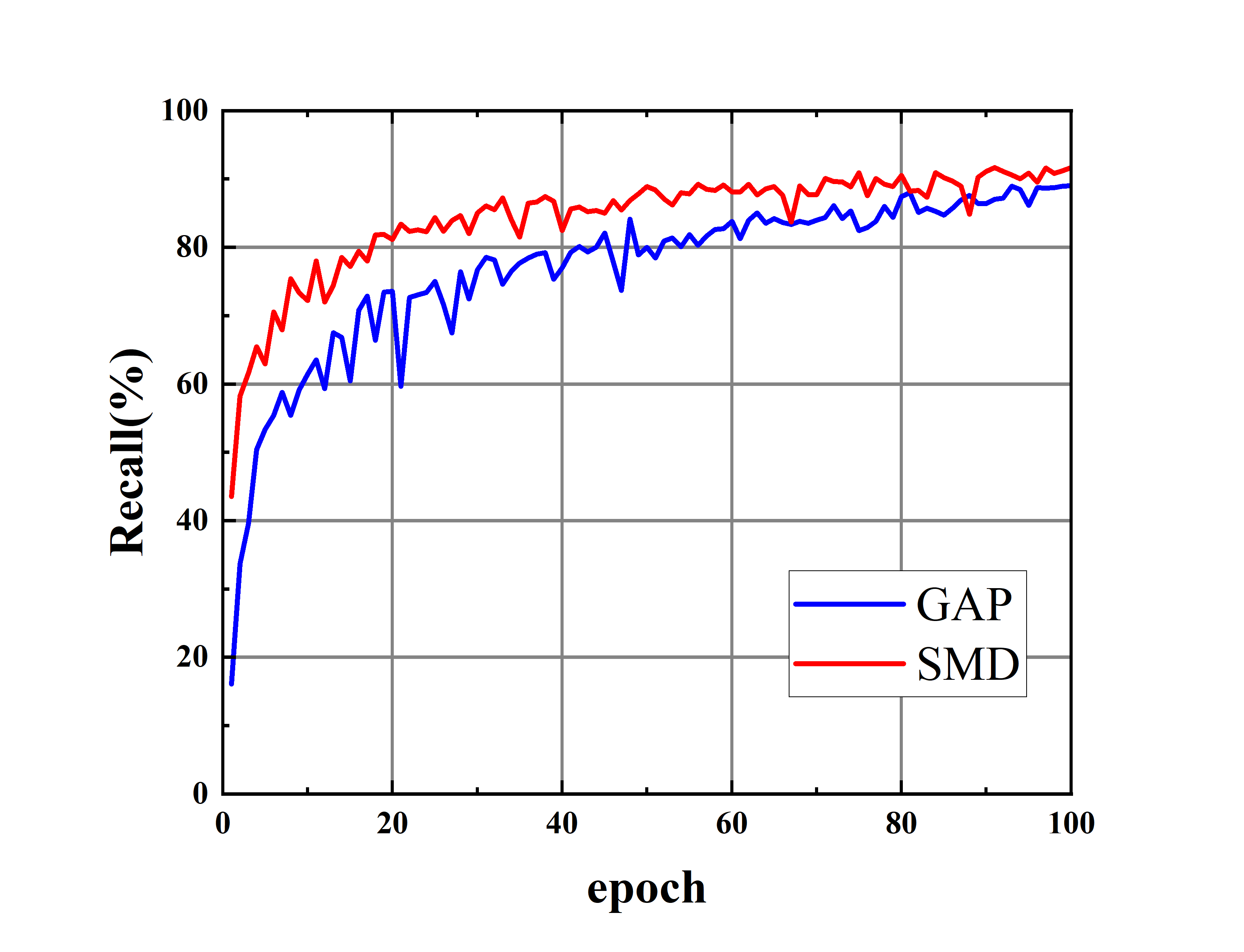} 
}
\end{minipage}%
\begin{minipage}[t]{0.335\linewidth}
\centering
\subfigure[CVAVT]{
\label{pool examples:B}
\includegraphics[width=2.3in]{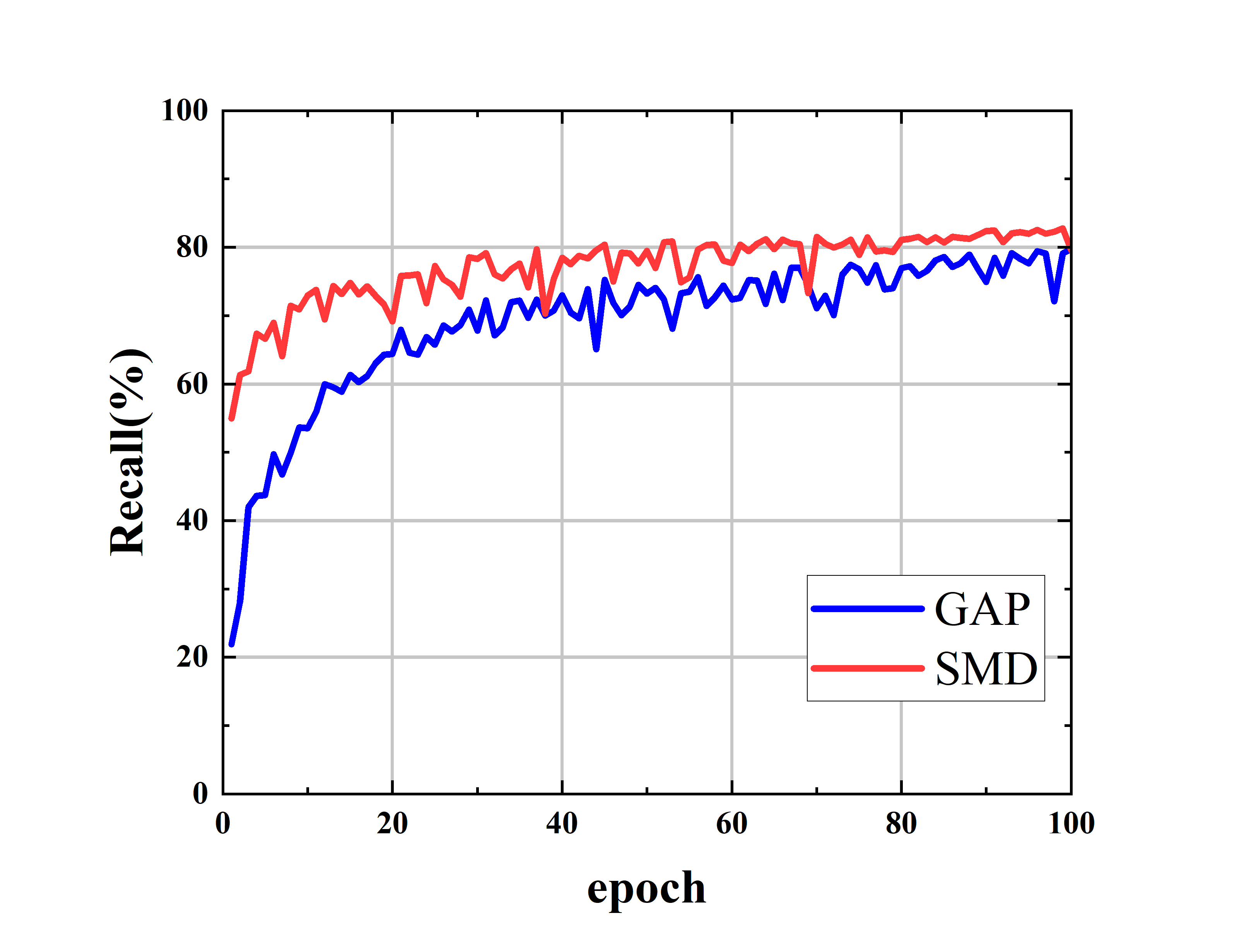}
}
\end{minipage}%
\begin{minipage}[t]{0.33\linewidth}
\centering
\subfigure[VIGOR]{
\label{pool examples:C}
\includegraphics[width=2.3in]{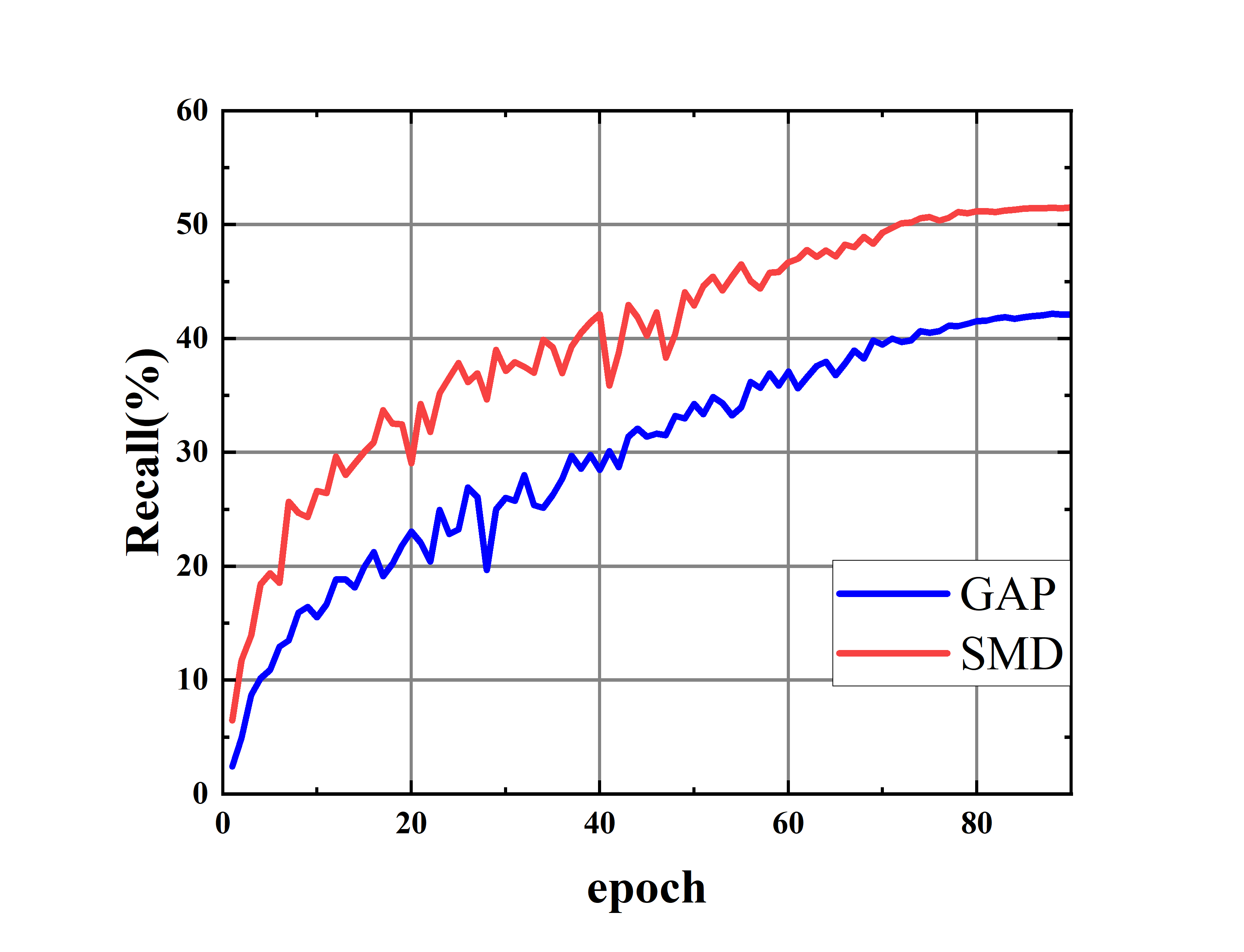}
}
\end{minipage}

\caption{Visualization of training curve (r@1) on (a) CVUSA, (b) CVACT and (c) VIGOR. The blue lines show the curve of the SAIG-D backbone training with GAP and the red lines show the the curve of the SAIG-D network training with our SMD. Compared to GAP, our SMD significantly improves the performance and saturation rate of SAIG.}
\label{pool}
\vspace{-10pt}
\end{figure*}

\begin{figure}[ht] 
\begin{minipage}[t]{0.33\linewidth} 
\centering

\subfigure[CVUSA]{
\includegraphics[width=1in]{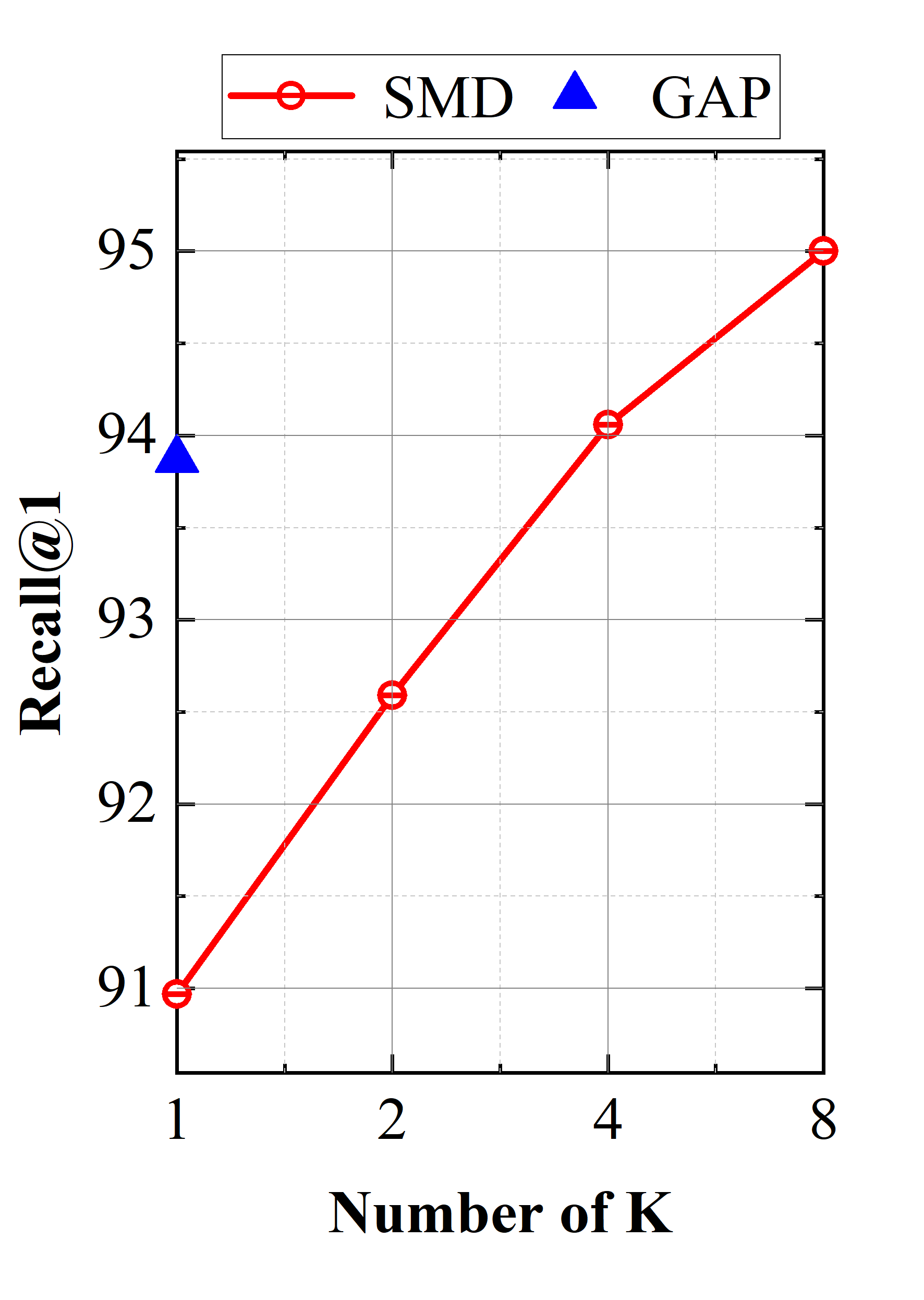} 
}
\label{fig1} 
\end{minipage}%
\begin{minipage}[t]{0.33\linewidth}
\centering
\subfigure[CVAVT]{
\includegraphics[width=1in]{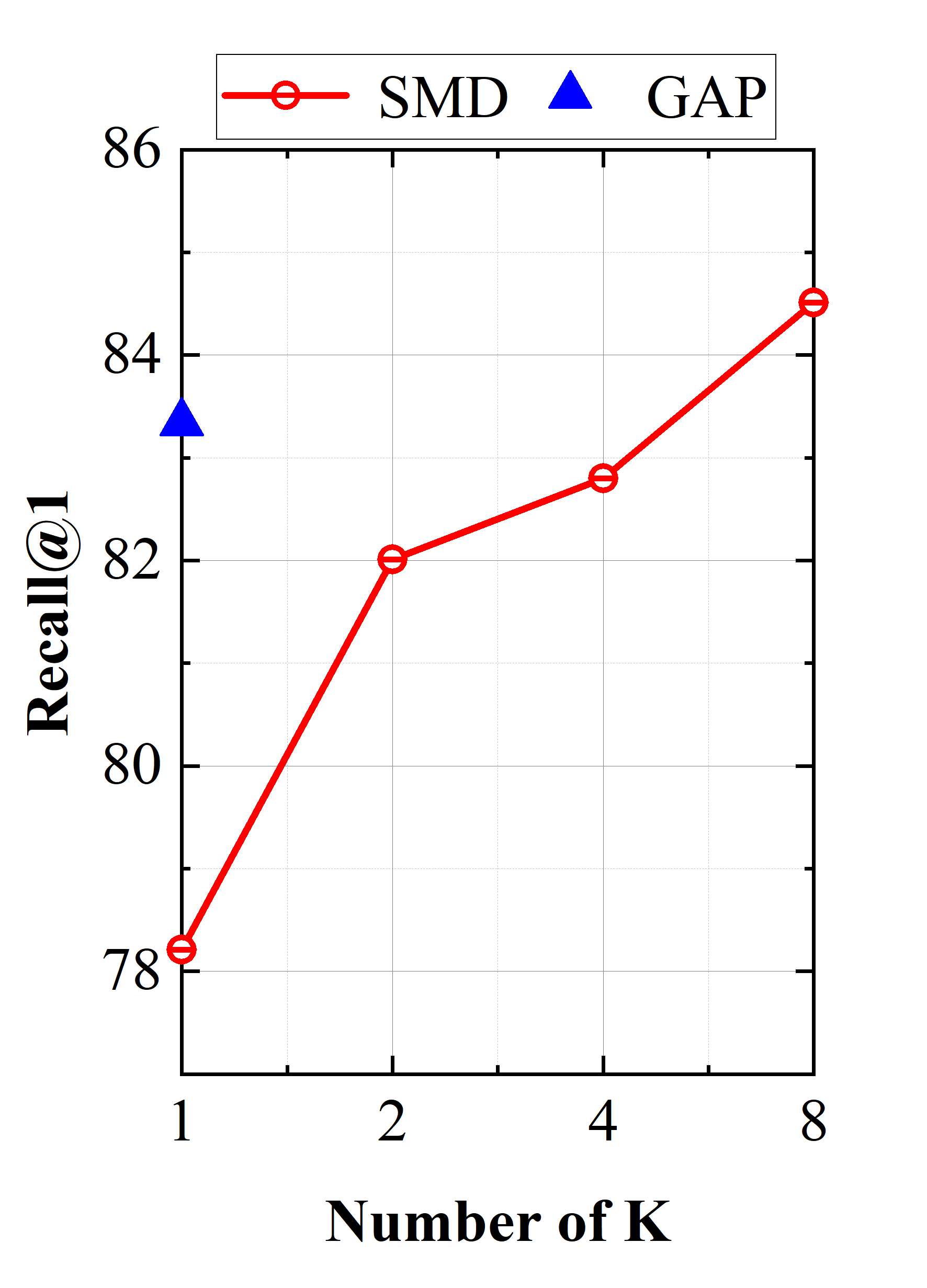}
}
\label{fig2}
\end{minipage}%
\begin{minipage}[t]{0.33\linewidth}
\centering
\subfigure[VIGOR]{
\includegraphics[width=1in]{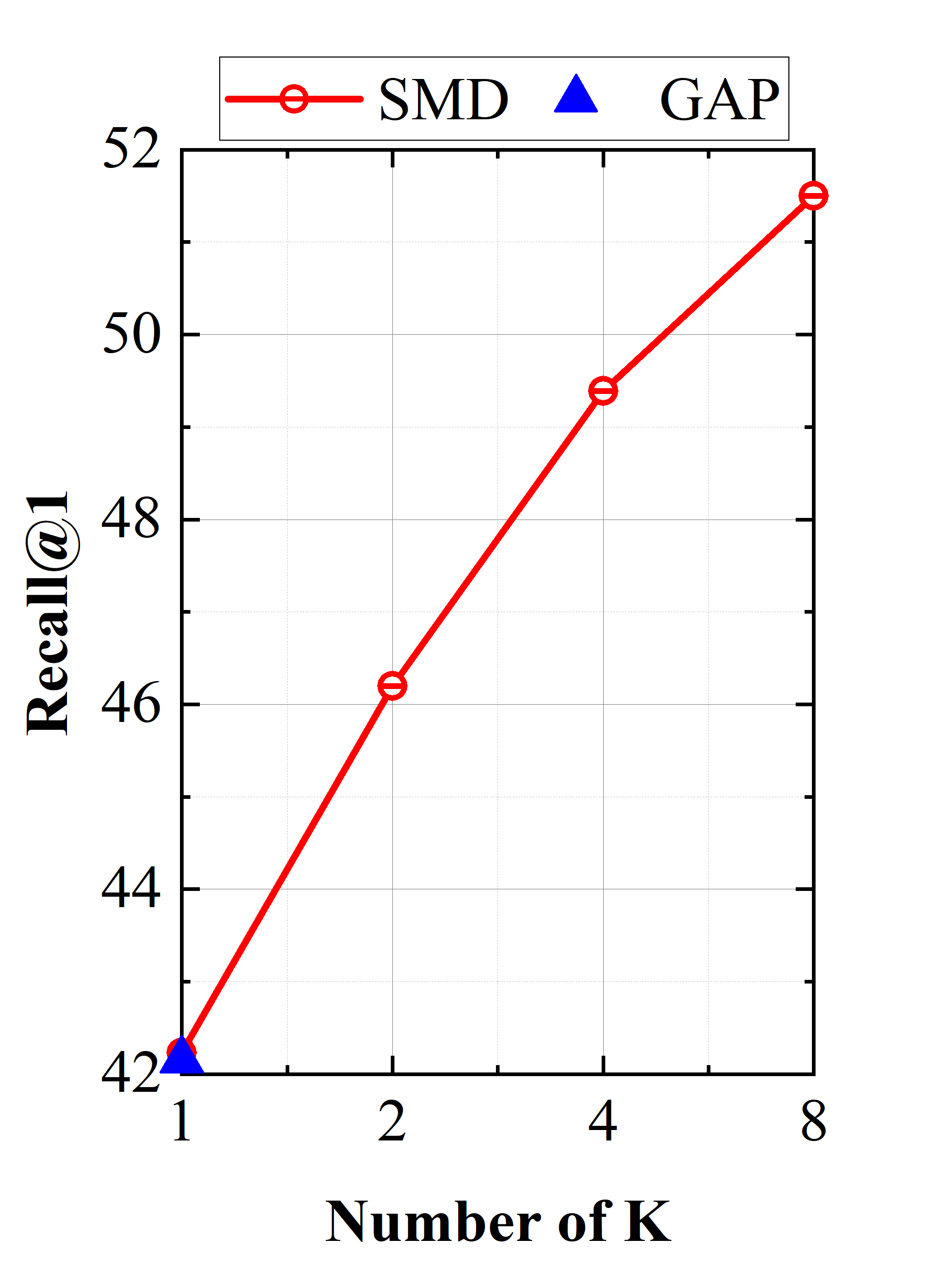}
}
\label{fig2}
\end{minipage}%

\caption{Cross-view geo-localizaion accuracy versus the number of \emph{K} in SMD training with SAIG-D. Note that the log scale of the x-axis.}
\label{pool_k}
\vspace{-10pt}
\end{figure}

\bmhead{Different pooling layer} We then replace the GAP with our SMD in our SAIG-D backbone on three datasets, respectively. And this method yields an improvement of 9\% at r@1 on VIGOR, indicating that our SMD is able to aggregate a more discriminative representation. In addition, as shown in figure \ref{pool examples:A} and figure \ref{pool examples:B}, the SAIG-D using SMD starts saturation after roughly 50 epochs, whereas the backbone with GAP takes longer to reach the same performance and ends up worse than SMD. This demonstrates that our SMD can capture key information rapidly at the early stage of training.

\begin{table}[!htbp]
\setlength{\abovecaptionskip}{0.2cm}
\centering
\begin{tabular}{c|c|c|c|c|c|c}
\toprule
\multirow{2}{*}{AA} & \multirow{2}{*}{Feature}& \multirow{2}{*}{Dim} & r@1 & r@5 & r@10 & r@1\% \\ 
~ & ~& ~ & (\%) & (\%) & (\%) & (\%) \\\midrule
\multirow{3}{*}{w/o} & Local & 3072 &85.05&96.56&98.10&99.74\\
~& GAP &\textbf{384} &90.29&97.71&98.74&\textbf{99.76}\\ 
~& SMD &3072 &\textbf{92.71}&\textbf{97.92}&\textbf{98.89}&99.71\\ \midrule
\multirow{3}{*}{w} & Local & 3072 &94.91&98.57&99.17&99.82 \\
~& GAP & \textbf{384} &93.87&98.60&\textbf{99.30}&\textbf{99.85} \\ 
~& SMD &3072 &\textbf{95.00}&\textbf{98.66}&99.20&99.83\\ \bottomrule
\end{tabular}
\caption{Comparison with SAIG-D using different features on the CVUSA dataset. Local features with Assumption of Alignment are better at precision, while global features are better at universality. SMD module is proposed to better aggregates global features. }
\label{global vs local}
\end{table}

\bmhead{Global feature vs. Local feature} We compare the SAIG-D on the CVUSA dataset using the global and local features\footnote{For local feature, the resolution of feature maps is reduced to $4\times16\times48$ by one average pooling layer and one fully-connected layer, and is finally unfolded along the spatial dimension to output a 3072-d descriptor}, respectively. In Table \ref{global vs local}, the global features (GAP and our proposed SMD) perform well with and without the alignment information, while the local feature only adapts well with the assumption of alignment, with 5.24\% drop at r@1 compared to global features in the non-aligned case. We conclude from the results that local features are more precise with aligned features, and global features have better generalization at different settings. Compared to GAP, SMD achieves 2.42\% improvement at r@1.
%

%

\subsection{Ablation Study}
%

%
\subsubsection{Backbone} Ablation study are also conducted to validate the contribution of each component in our SAIG, including five variants: (a) an overlapping conv stem with MSA module (i.e., our SAIG-D backbone); (b) replacing conv stem with non-overlapping patch stem; (c) integrating the FFN sub-layer and an add-norm operation follow with each MSA layer; (d) a counterpart of baseline which shares network weights in two branches. 
All ablation experiments are conducted on SAIG-D baseline. Table \ref{ablation experiment} reports their number of parameters and r@1 on the CVUSA dataset.

\bmhead{Conv stem vs. Patch stem} Comparing Table \ref{ablation experiment} (a) with (b), we conclude that the conv stem greatly facilitates modeling global interactions by generating overlapping patch features. There is a 34.5\% drop at r@1 for the model with a patch stem against one with a conv stem. It implies that the conv stem is a more effective module to compress the input patch and project it into patch embedding for cross-view tasks. The patch embedding provided by conv stem could facilitate the self-attention mechanism for better modeling global relationships and obtaining more representational global contents. 

\bmhead{FFN vs. No FFN} For the SAIG with FFN layer in our ablation, each MSA layer is followed by a FFN layer and an add-norm operation. From Table \ref{ablation experiment} (c), although the addition of the FFN layers improves the model slightly, it leads to an extra cost (up by 52.1M) in terms of parameters.


\bmhead{Weight sharing} We argue that the features of images from different domain should be extracted by using domain-specific branch with independent weights. From Table \ref{ablation experiment} (a) and (d), we can observe that the model with shared weights drops about 5\% at r@1. It verifies that the features from images of different domains/views should be extracted by using domain-specific branches with independent weights. 

\begin{table}[htbp]
\setlength{\abovecaptionskip}{0.2cm}
\centering
\begin{tabular}{cl|c|c}
\toprule
\# & Models & Params (M)& r@1(\%) \\ \midrule
(a) & \textbf{Baseline} & 31.2& 93.87 \\
(b) & {- w/ Patch stem} & 26.8 & 59.79\\
(c) & {- w/ FFN} & 83.3 & 94.23\\
(d) & {- shared weight} &15.6 & 88.95\\
\bottomrule
\end{tabular}
\caption{Ablation study of the SAIG-D backbone on CVUSA.}
\label{ablation experiment}
\vspace{-5pt}
\end{table}


 \begin{table*}
\setlength{\abovecaptionskip}{0.2cm}
\centering
\begin{tabular}{clc|r|cccc}
\toprule
\multirow{2}{*}{Input Size} & \multirow{2}{*}{Backbone} & \multirow{2}{*}{Pooling Layer} & \multirow{2}{*}{Dim} & \multicolumn{2}{c}{$\mathcal{R}$Oxford} & \multicolumn{2}{c}{$\mathcal{R}$Paris}\\ \cmidrule{5-8}
~&~&~ &~&M&\multicolumn{1}{c|}{H}&M&H\\ \midrule 
\multirow{5}{*}{$224\times224$}&${}^{\star}$R50&GeM&\multirow{4}{*}{2048}&28.7&\multicolumn{1}{c|}{10.9}&61.2&35.9 \\
~&${}^{\star}$R50&R-MAC&~&25.6&\multicolumn{1}{c|}{7.3}&60.6&35.4 \\
~&${}^{\star}$R101&GeM&~&\textbf{31.7}&\multicolumn{1}{c|}{11.1}&63.4&37.3 \\
~&${}^{\star}$R101&R-MAC&~&31.0&\multicolumn{1}{c|}{9.3}&62.6&36.5 \\
~&SAIG-D&GAP&384& 31.4&\multicolumn{1}{c|}{\textbf{11.4}}& \textbf{65.9}& \textbf{41.2}\\ \midrule
\multirow{3}{*}{$384\times384$}&${}^{\star}$R101&GeM&\multirow{2}{*}{2048}&38.1&\multicolumn{1}{c|}{12.5}&\textbf{69.4}&\textbf{45.8} \\
~&${}^{\star}$R101&R-MAC&~&37.1&\multicolumn{1}{c|}{10.6}& 66.0& 41.4 \\
~&SAIG-D&GAP&384& \textbf{38.3}&\multicolumn{1}{c|}{\textbf{19.6}} &68.2 &42.3  \\
\bottomrule
\end{tabular}
\caption{Image retrieval mAP performance comparison with different convolution backbones ResNet-50 (R50) and ResNet-101 (R101) on the well acknowledged Medium (M) and Hard (H) difficulty metrics. $^{\star}$: the results in the table are from \protect\citep{TrainViTforImageRetri}}
\label{Image retrieval}
\end{table*}
 
\subsubsection{SMD and Losses} Figure \ref{pool_k} reports the correlations between our SMD performance and the number of \emph{K}. \emph{K} denotes the SMD project feature maps into \emph{K}-d space along the spatial dimension. 
As \emph{K} increases, our network obtain better performance. It is noteworthy that simply assigning the pool layer as a learnable layer may exert a negative effect on improving network (e.g., \emph{K}=1). Instead, a larger \emph{K}, i.e., 4 and 8, is required for the maximum performance of SMD in our network. In short, the projection of local feature into the larger k-dim space in SMD is critical for cross-view geo-localization. Besides, Table \ref{CVUSA and CVACT}, \ref{CVACT_test} and \ref{VIGOR} demonstrate the effectiveness of our SMD, which brings favorable improvement in both SAIG-S and SAIG-D. Figure \ref{pool} presents the training curve of network with SMD.

Moreover, we compare the performance of our SAIG on different losses in one-to-many cross-view image matching in Table \ref{VIGOR}. As can be seen, both semi-hard triplet loss (Eq. \ref{semihardloss}) and InfoNCE loss (Eq. \ref{infoNCEloss}) can improve our SAIG in one-to-many tasks. Compared to the origin soft-margin triplet loss, hard samples can be identified for training either explicitly through semi-hard negative mining strategy or implicitly through infoNCE loss. Both are effective for generating uniform embedding distribution~\citep{wang2021understanding} and thus improve the models.

\subsection{Image Retrieval}
In this section, we focus on validate the generalization capability of our backbone. The experimental results on the revisited Oxford and Paris benchmarks are reported in Table \ref{Image retrieval}. SAIG-D achieves evident performance enhancement compared to ResNet. Especially on the $\mathcal{R}$Paris dataset with images of $224\times224$ input size, SAIG-D improves the mAP by 2.5\% and 3.9\%, compared with the best performance of ResNet models. Its effectiveness for the retrieval task further demonstrates the generalization of our SAIG. The Multi-head Self-Attention module, as the main part of our SAIG, can effectively extract global vision features in image retrieval tasks. Note that we intend to pursue a new flexible and simple network rather than a new state-of-the-art performance with various retrieval tricks.

\begin{table}
\setlength{\abovecaptionskip}{0.2cm}
\centering
\begin{tabular}{c|c|c|c}
\toprule
\multirow{2}{*}{Backbone} & Params & \multirow{2}{*}{GFLOPs} & Top-1 Acc \\ 
 ~&(M) & ~ &(\%) \\\midrule
ResNet-50 & 25.6 & 4.1 & 76.1 \\
DeiT-S & 22.1 & 4.6 & 79.9 \\
T2T-ViT-14 & 22.0 & 6.1 & 80.7 \\
PVT-S  & 24.5 & 3.8 & 79.8 \\
TNT-S & 23.8 & 5.2 & 81.3 \\
Swin-T  & 29.0 & 4.5 & 81.3 \\\midrule
SAIG-S & 9.5 & 3.3 & 77.2\\
SAIG-D & 16.0 & 4.9 & 80.3\\ \bottomrule
\end{tabular}

\caption{Parameters count, FLOPs and ImageNet Top-1 accuracy for baselines ResNet-50~\protect\citep{heResNet2016}, DeiT-S~\protect\citep{DeiT}, T2T-ViT-14~\protect\citep{T2T-ViT2021}, PVT-S~\protect\citep{PVT2021}, TNT-S~\protect\citep{TNT2021}, Swin-T~\protect\citep{Swin-T2021} and Our SAIG. “GFLOPs” is calculated under the size of 224$\times$224.}
\label{Imagenet classification}
\end{table}

\subsection{Pretraining Results} Our backbones are first pre-trained on ImageNet-1k. To demonstrate the effectiveness of our SAIG on vision task, we present its classification accuracy on the ImageNet-1k dataset. Table \ref{Imagenet classification} reports the Top-1 accuracy, parameters, and FLOPs of different backbones and SAIG on ImageNet-1k image classification. Our backbones are trained on ImageNet-1k from scratch. Compared to other backbones, SAIG shows remarkable superiority in terms of model size and gains competitive performance on top-1 accuracy. SAIG-S and SAIG-D achieve 77.2\% and 80.3\% top-1 accuracy on ImageNet-1k, respectively. SAIG-D achieves 4.2\% higher accuracy with 37.5\% fewer parameters compared to ResNet-50. 
Although our SAIG-D is slightly inferior to TNT-S and Swin-T by 1\% in terms of Top-1 accuracy, our SAIG-D has only 67\% and 55\% of the model parameters of TNT-S and Swin-T.
This result implies that SAIG, with a simple structure and fewer parameters, is effective in extracting vision features.

\subsection{Visualization Analysis}

\bmhead{Visualization of Heatmaps} To visualize what is being learnt by our SAIG, we visualize the heatmaps of the proposed backbone. 

\begin{figure}[ht]
	\centering
	\subfigure{
		\begin{minipage}[t]{0.6\linewidth}
			\centering
			{Ground}
		\end{minipage}
		\begin{minipage}[t]{0.3\linewidth}
			\centering
			{Aerial}
		\end{minipage}
		
	}\vspace{-3mm}
	
	{\rotatebox{90}{
	\centering
	\scriptsize{Input}}}
	\centering
	\subfigure{
		\begin{minipage}[t]{0.6\linewidth}
			\centering
			\raisebox{-0.15cm}{\includegraphics[height=0.3\linewidth,width=1\linewidth]{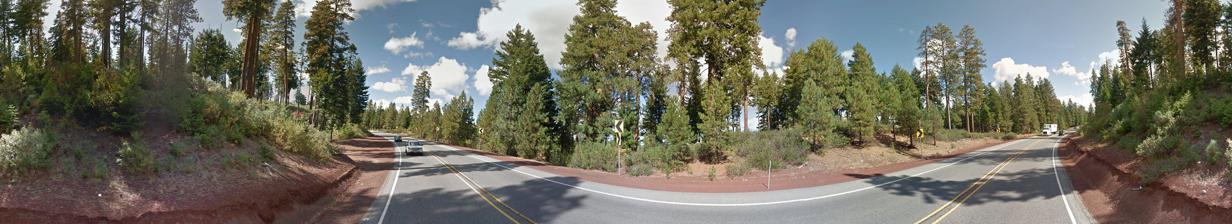}}
		\end{minipage}
		\begin{minipage}[t]{0.3\linewidth}
			\centering
			\raisebox{-0.15cm}{\includegraphics[height=0.6\linewidth,width=0.6\linewidth]{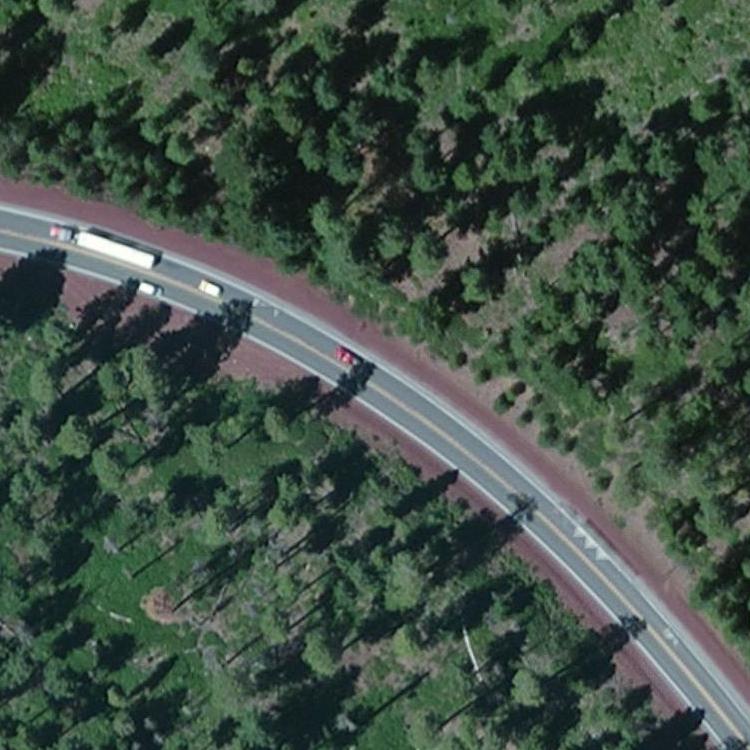}}
		\end{minipage}
	}\vspace{-3mm}
	
	\rotatebox{90}{
	\centering
	\scriptsize{SAIG-S}}
	\centering
	\subfigure{
		\begin{minipage}[t]{0.6\linewidth}
			\centering
			\raisebox{-0.15cm}{\includegraphics[height=0.3\linewidth,width=1\linewidth]{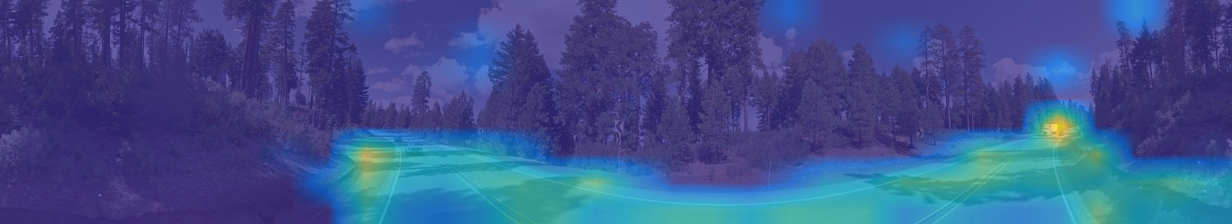}}
		\end{minipage}
		\begin{minipage}[t]{0.3\linewidth}
			\centering
			\raisebox{-0.15cm}{\includegraphics[height=0.6\linewidth,width=0.6\linewidth]{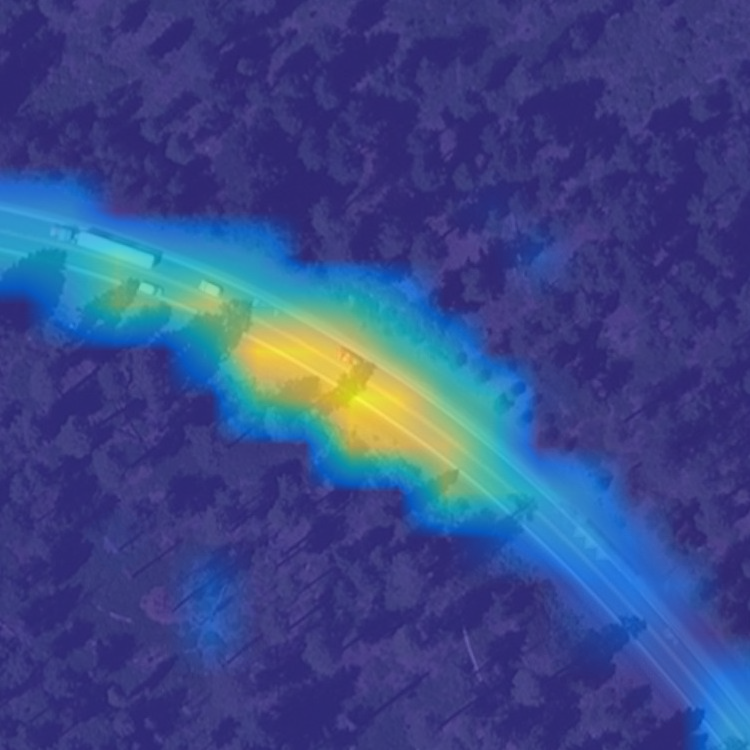}}
		\end{minipage}
	}\vspace{-3mm}
	
	{\rotatebox{90}{
	\centering
	\scriptsize{SAIG-D}}}
	\centering
	\subfigure{
		\begin{minipage}[t]{0.6\linewidth}
			\centering
			\raisebox{-0.15cm}{\includegraphics[height=0.3\linewidth,width=1\linewidth]{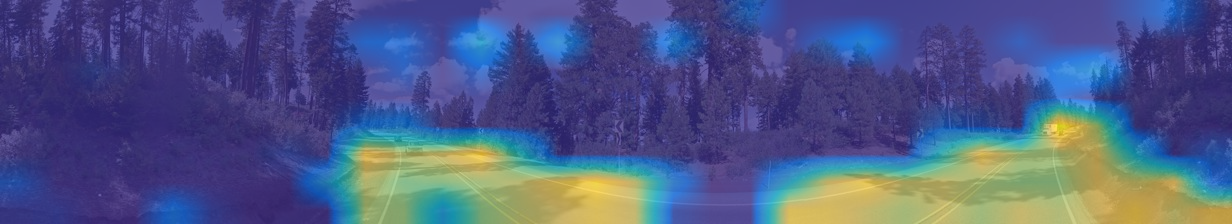}}
		\end{minipage}
		\begin{minipage}[t]{0.3\linewidth}
			\centering
			\raisebox{-0.15cm}{\includegraphics[height=0.6\linewidth,width=0.6\linewidth]{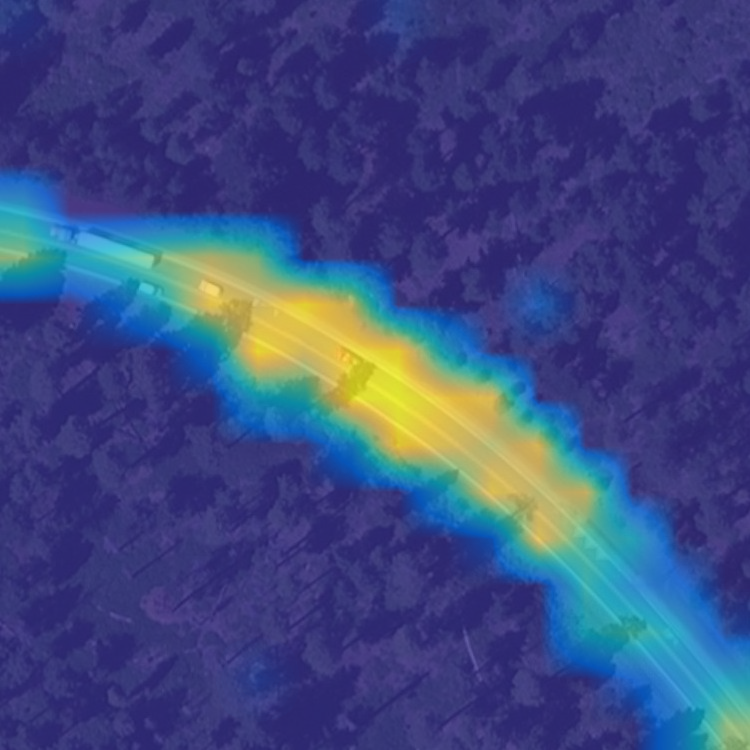}}
		\end{minipage}
	}

	\caption{Visualization of heatmaps in Cross-View Geo-localization. Top: input images. Middle: heatmaps in SAIG-S. Bottom: heatmaps in SAIG-D.} 
	\label{heatmap of cross view}
\end{figure}

Figure \ref{heatmap of cross view} shows the heatmaps in different variants of SAIG. We notice that SAIG-S focuses on the road information, while SAIG-D pays more attention to contextual information such as roads, roadbeds, and trees, which are more reasonable to cross-view geo-localization. 

\begin{figure}[!htbp]
	
	\centering
	\subfigure{
		\begin{minipage}[t]{0.235\linewidth}
			\centering
			\raisebox{-0.15cm}{\includegraphics[height=1\linewidth,width=1\linewidth]{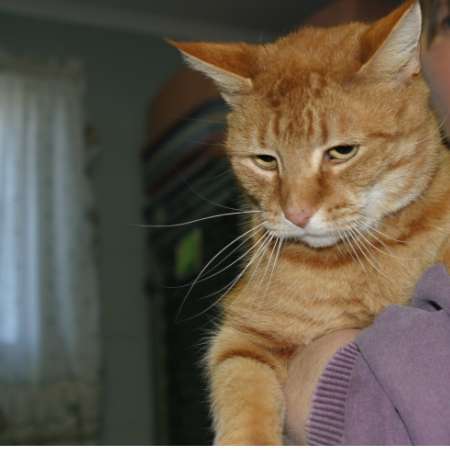}}
		\end{minipage}
		\begin{minipage}[t]{0.235\linewidth}
			\centering
			\raisebox{-0.15cm}{\includegraphics[height=1\linewidth,width=1\linewidth]{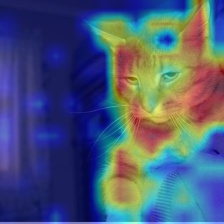}}
		\end{minipage}
		\begin{minipage}[t]{0.235\linewidth}
			\centering
			\raisebox{-0.15cm}{\includegraphics[height=1\linewidth,width=1\linewidth]{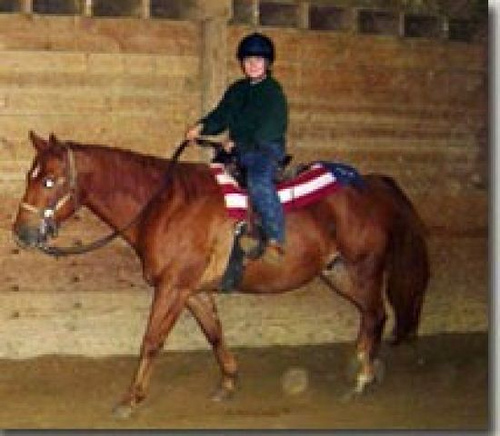}}
		\end{minipage}
		\begin{minipage}[t]{0.235\linewidth}
			\centering
			\raisebox{-0.15cm}{\includegraphics[height=1\linewidth,width=1\linewidth]{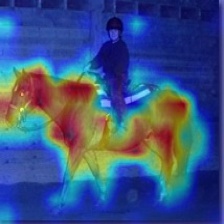}}
		\end{minipage}
	}\vspace{-3mm}
	
	\centering
	\subfigure{
		\begin{minipage}[t]{0.235\linewidth}
			\centering
			\raisebox{-0.15cm}{\includegraphics[height=1\linewidth,width=1\linewidth]{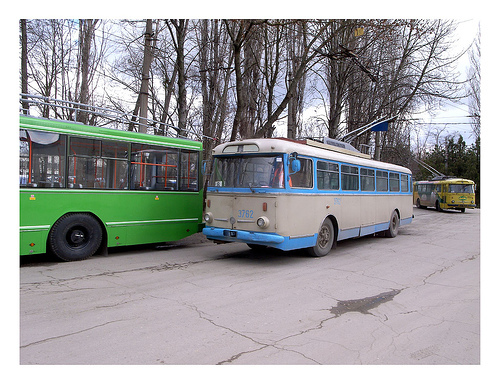}}
		\end{minipage}
		\begin{minipage}[t]{0.235\linewidth}
			\centering
			\raisebox{-0.15cm}{\includegraphics[height=1\linewidth,width=1\linewidth]{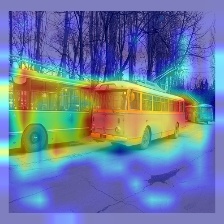}}
		\end{minipage}
		\begin{minipage}[t]{0.235\linewidth}
			\centering
			\raisebox{-0.15cm}{\includegraphics[height=1\linewidth,width=1\linewidth]{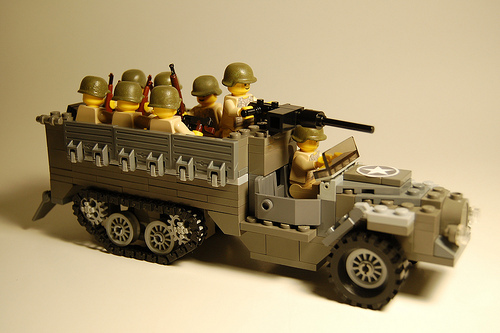}}
		\end{minipage}
		\begin{minipage}[t]{0.235\linewidth}
			\centering
			\raisebox{-0.15cm}{\includegraphics[height=1\linewidth,width=1\linewidth]{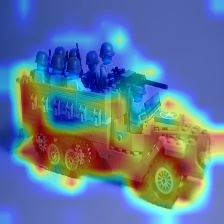}}
		\end{minipage}
	}\vspace{-3mm}
	
	\centering
	\subfigure{
		\begin{minipage}[t]{0.235\linewidth}
			\centering
			\raisebox{-0.15cm}{\includegraphics[height=1\linewidth,width=1\linewidth]{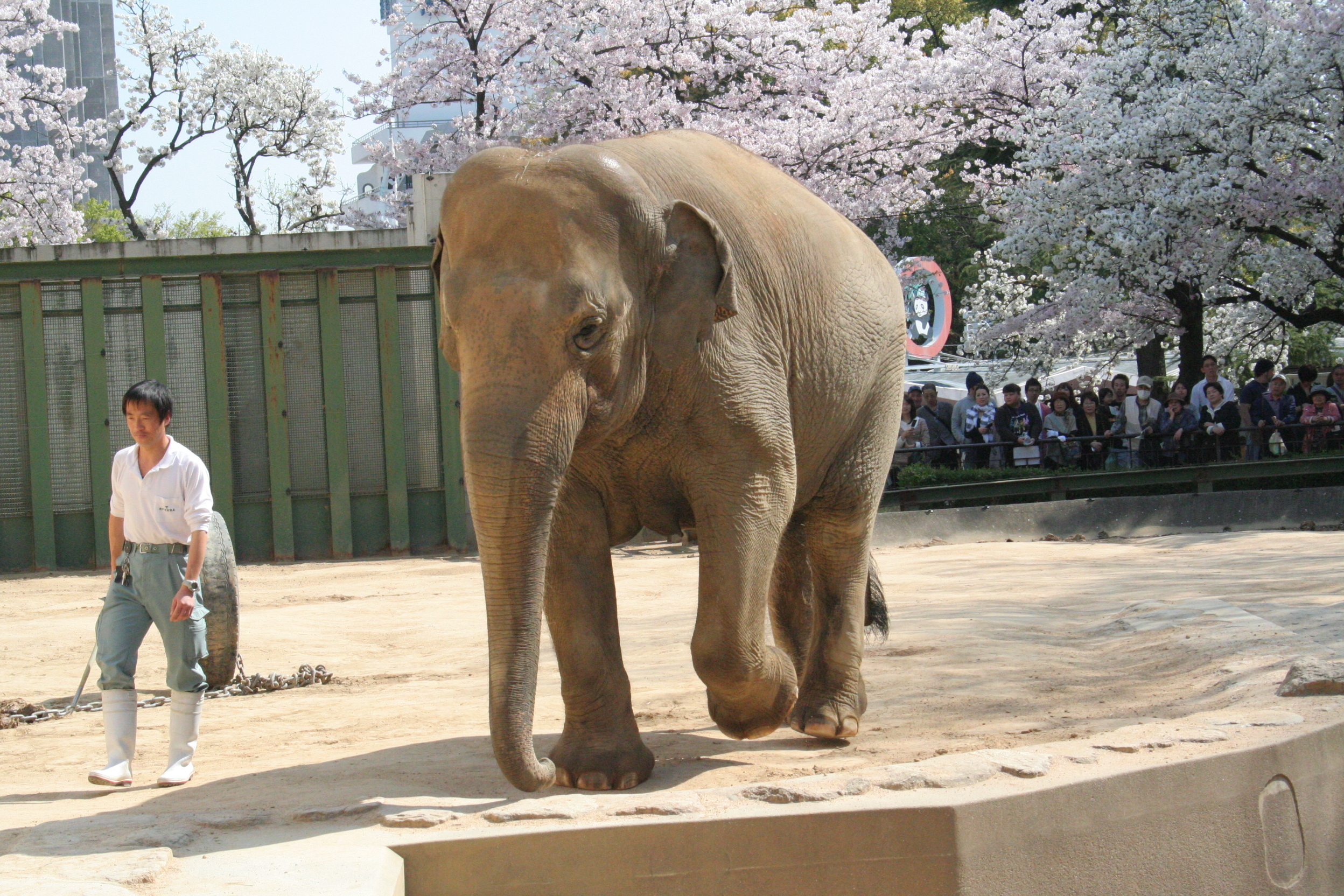}}
		\end{minipage}
		\begin{minipage}[t]{0.235\linewidth}
			\centering
			\raisebox{-0.15cm}{\includegraphics[height=1\linewidth,width=1\linewidth]{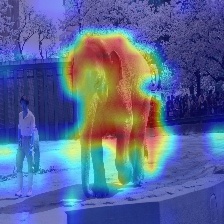}}
		\end{minipage}
		\begin{minipage}[t]{0.235\linewidth}
			\centering
			\raisebox{-0.15cm}{\includegraphics[height=1\linewidth,width=1\linewidth]{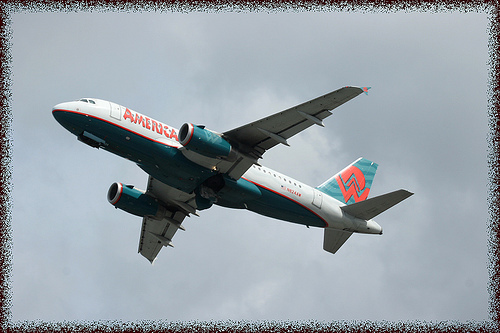}}
		\end{minipage}
		\begin{minipage}[t]{0.235\linewidth}
			\centering
			\raisebox{-0.15cm}{\includegraphics[height=1\linewidth,width=1\linewidth]{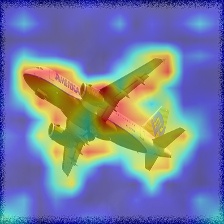}}
		\end{minipage}
	}
	
	\caption{Visualization of heatmaps in Image Classification with SAIG-D backbone.} 
	\label{heatmap of classification}
\end{figure}

In Figure \ref{heatmap of classification}, we shows the heatmaps generated by grad-CAM \citep{selvaraju2017grad} of SAIG-D backbone in image classification. SAIG-D concentrates on image regions that are highly relevant to the category and integrates semantic information globally, which demonstrate that our SAIG is effective in extracting vision features.

\begin{figure*}

	\centering
	\subfigure{
		\begin{minipage}[t]{0.286\linewidth}
			\centering
			\raisebox{-0.15cm}{\includegraphics[height=0.5\linewidth,width=1\linewidth]{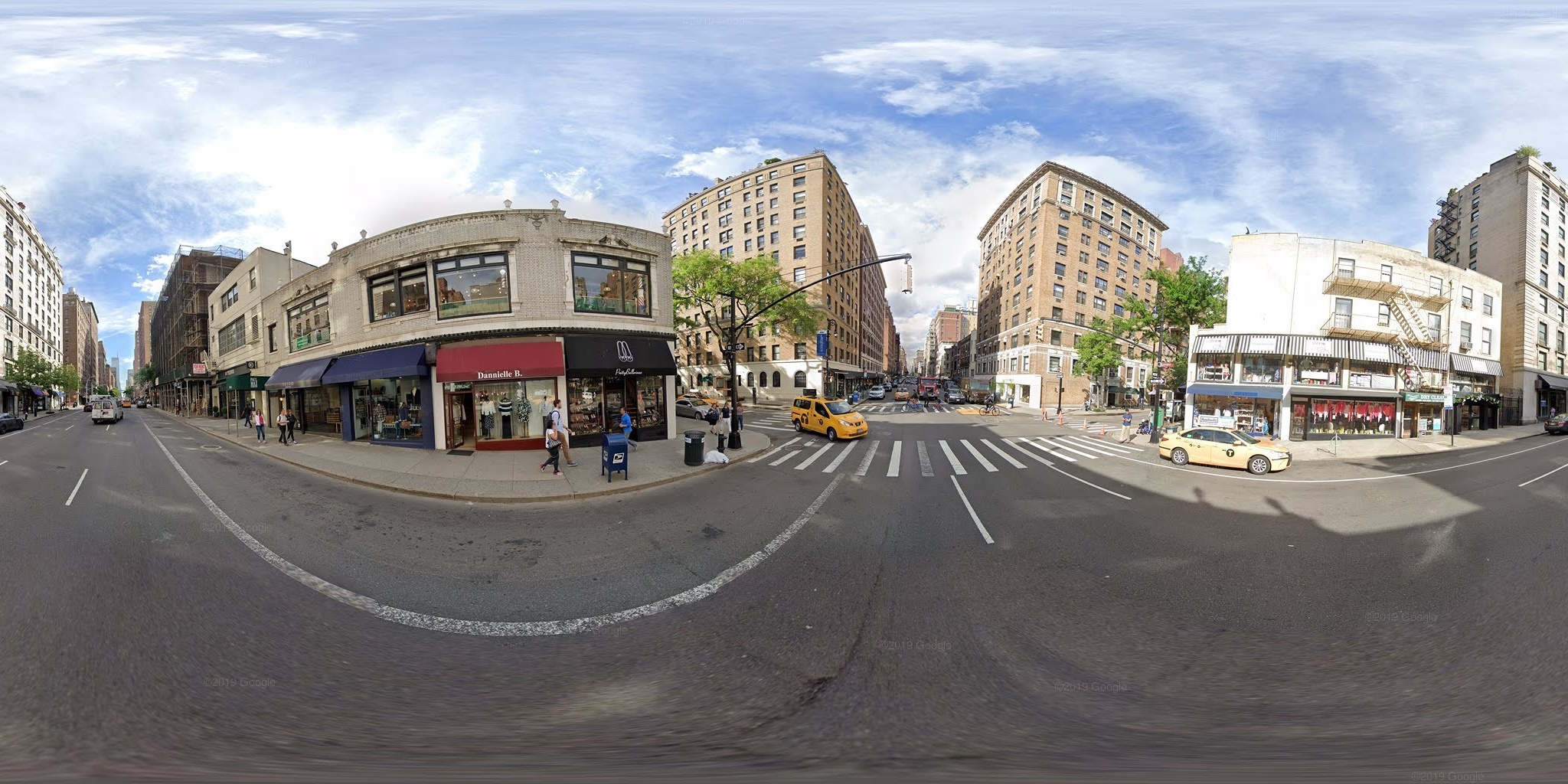}}
		\end{minipage}
		
		\begin{minipage}[t]{0.143\linewidth}
			\centering
			\raisebox{-0.15cm}{\includegraphics[height=1\linewidth,width=1\linewidth]{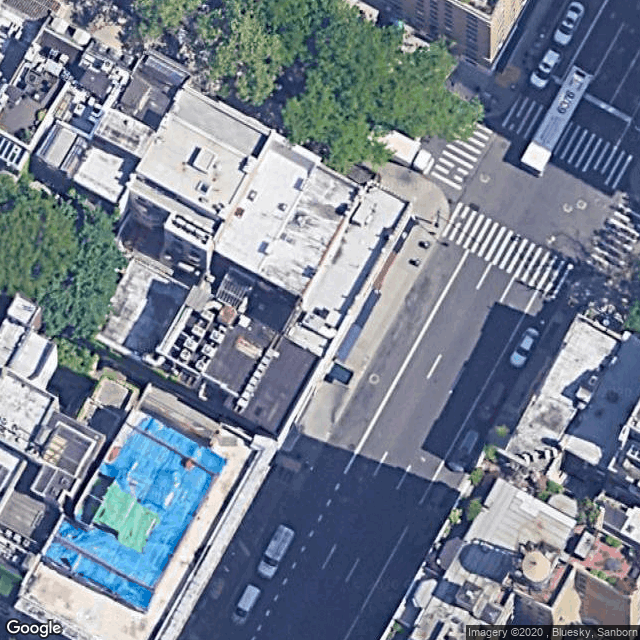}}
		\end{minipage}
		
		\begin{minipage}[t]{0.143\linewidth}
			\centering
			\raisebox{-0.15cm}{\includegraphics[height=1\linewidth,width=1\linewidth]{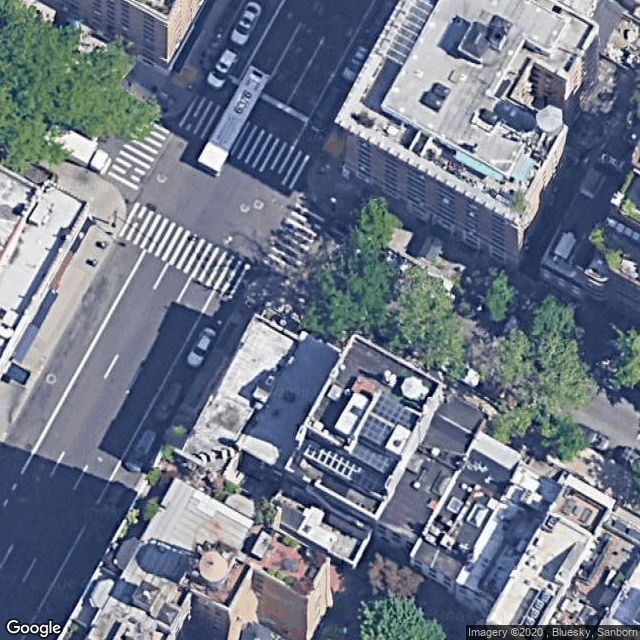}}
		\end{minipage}
		
		\begin{minipage}[t]{0.143\linewidth}
			\centering
			\raisebox{-0.15cm}{\includegraphics[height=1\linewidth,width=1\linewidth]{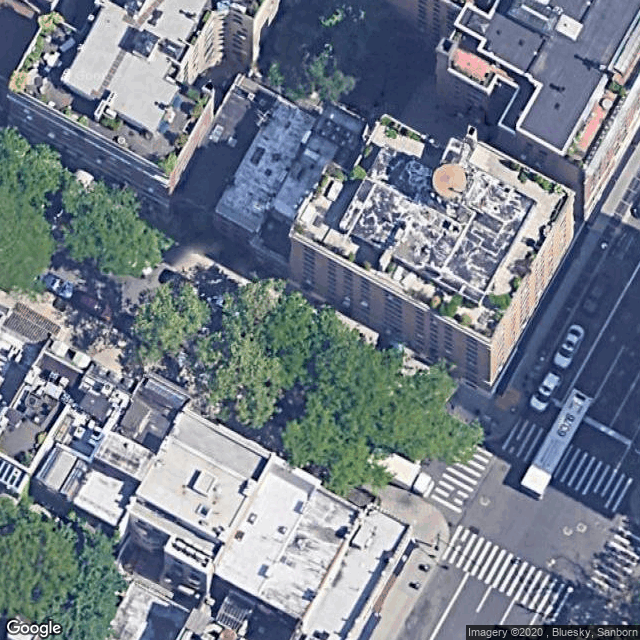}}
		\end{minipage}
		
		\begin{minipage}[t]{0.143\linewidth}
			\centering
			\raisebox{-0.15cm}{\includegraphics[height=1\linewidth,width=1\linewidth]{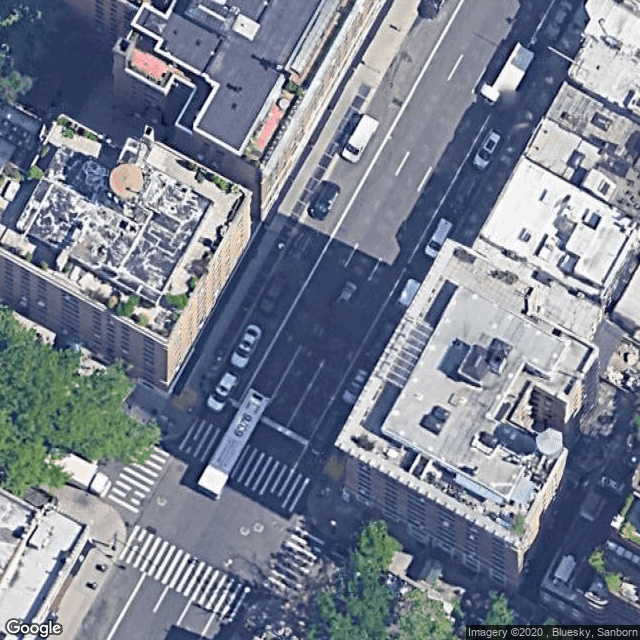}}
		\end{minipage}
	}\vspace{-3mm}
	\centering
	\subfigure{
		\begin{minipage}[t]{0.286\linewidth}
			\centering
			\raisebox{-0.15cm}{\includegraphics[height=0.5\linewidth,width=1\linewidth]{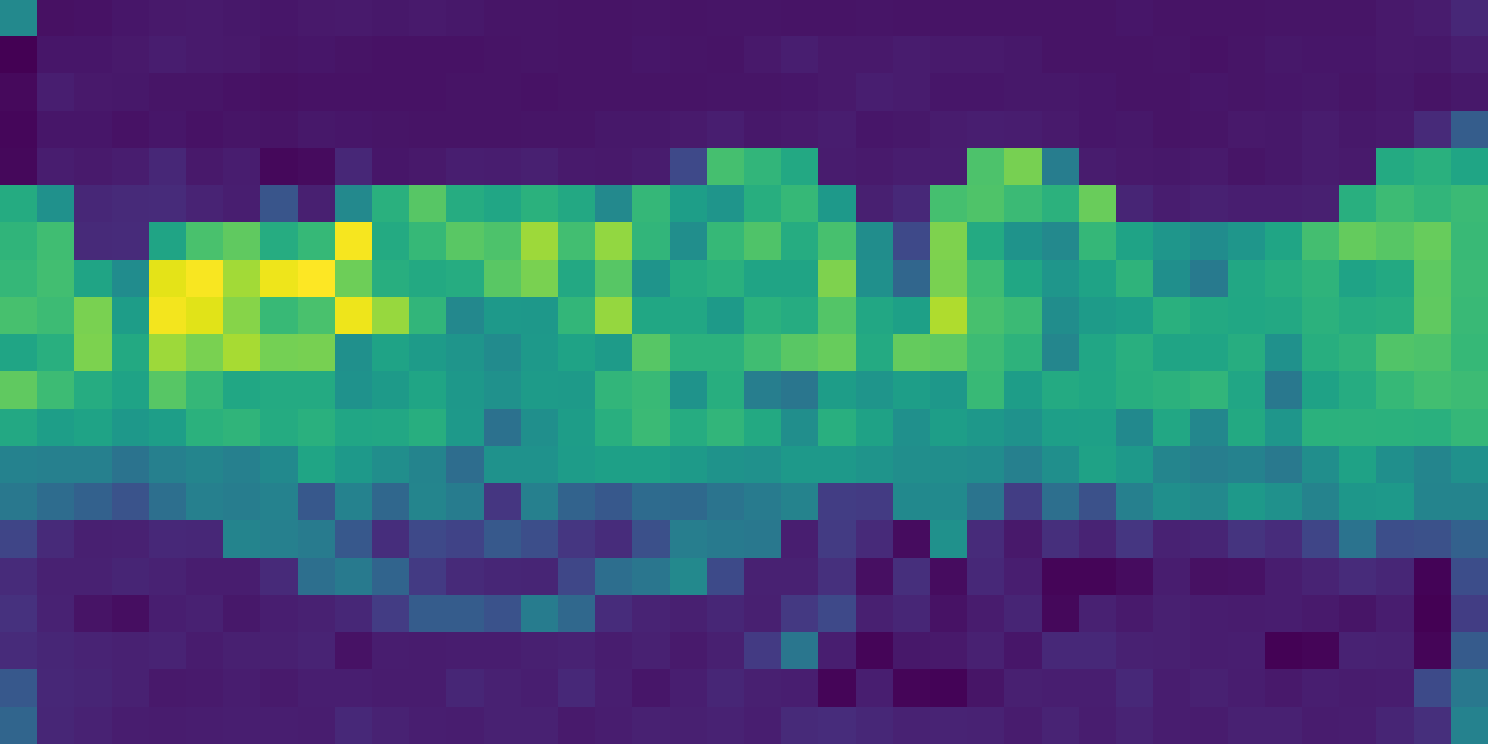}}
		\end{minipage}
		
		\begin{minipage}[t]{0.143\linewidth}
			\centering
			\raisebox{-0.15cm}{\includegraphics[height=1\linewidth,width=1\linewidth]{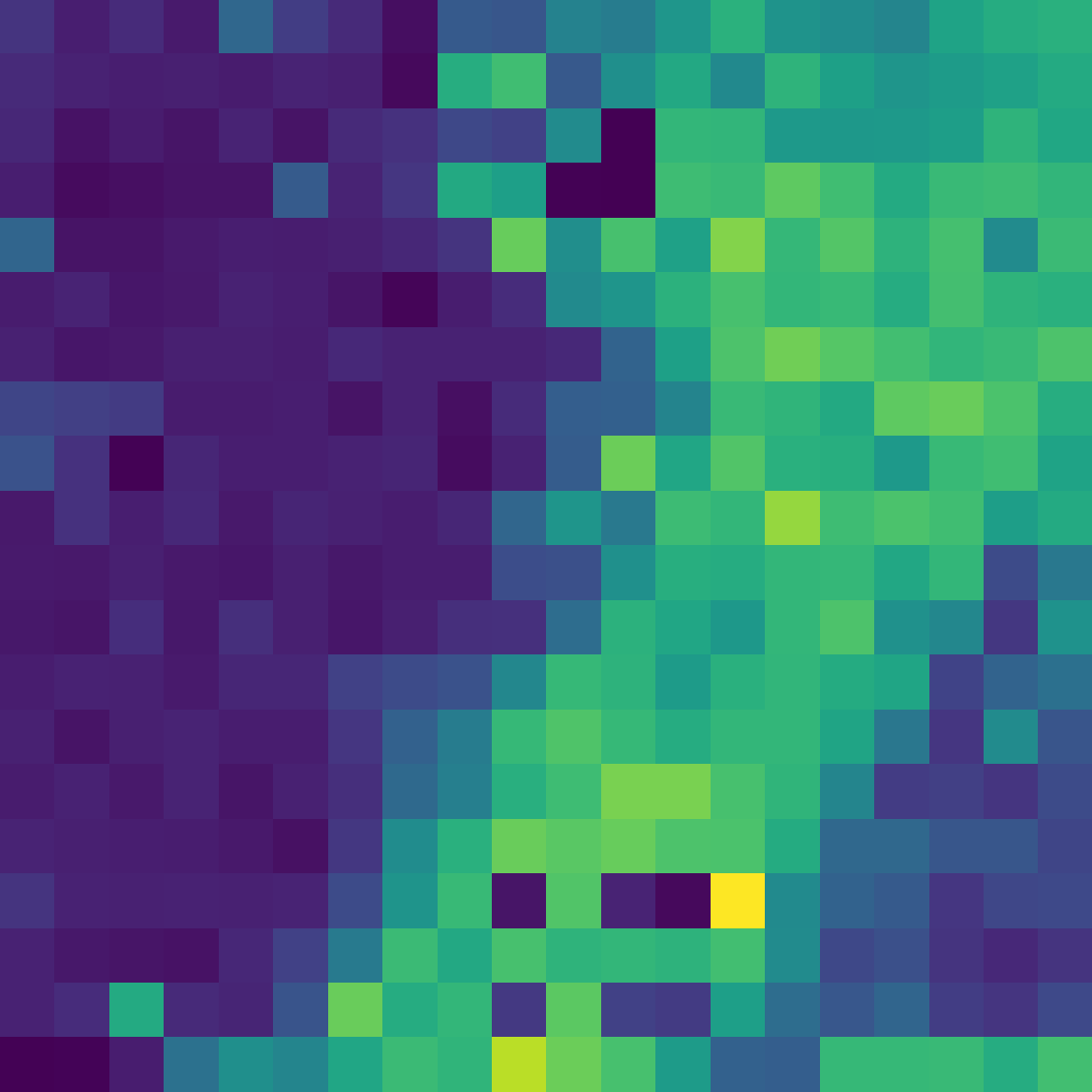}}
		\end{minipage}
		
		\begin{minipage}[t]{0.143\linewidth}
			\centering
			\raisebox{-0.15cm}{\includegraphics[height=1\linewidth,width=1\linewidth]{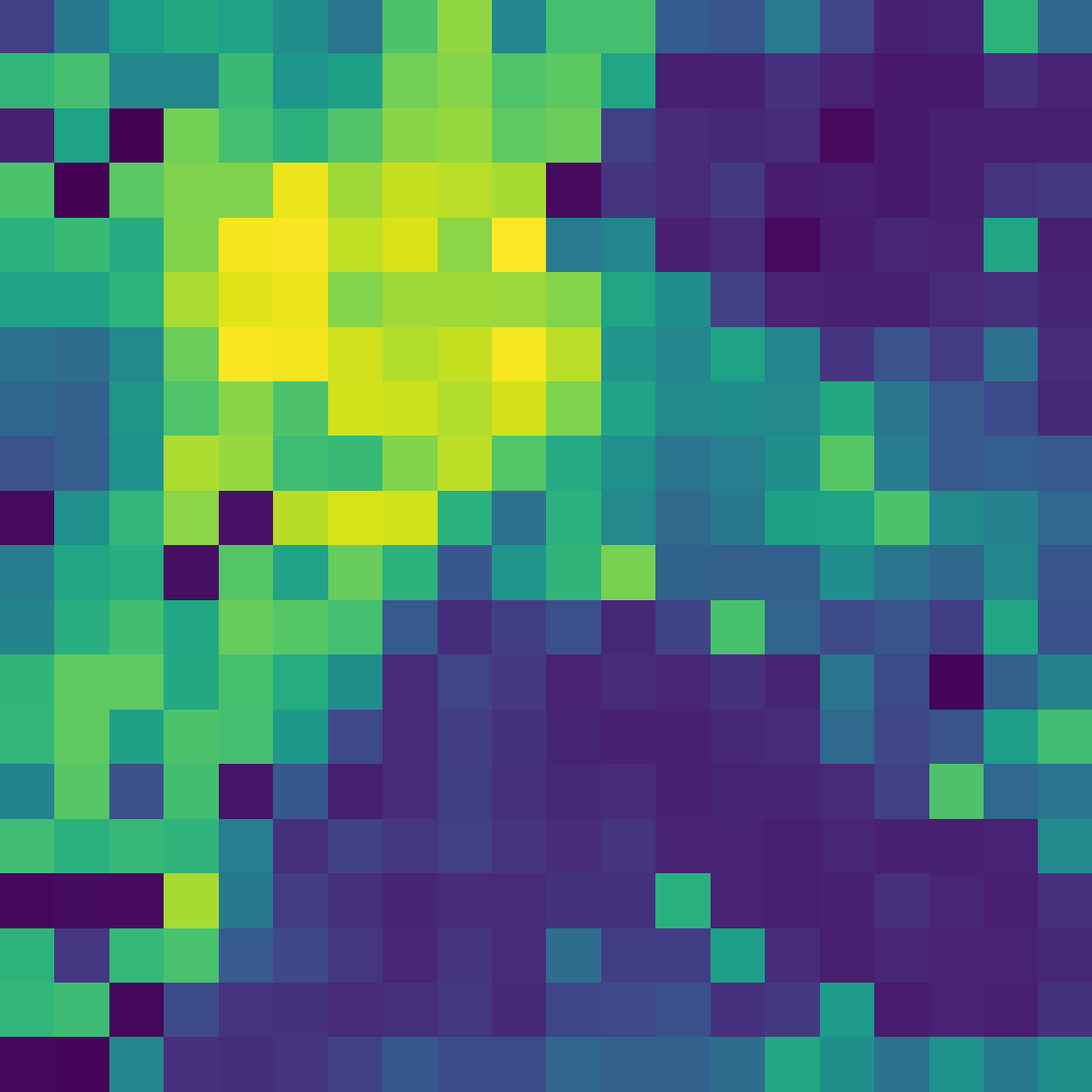}}
		\end{minipage}
		
		\begin{minipage}[t]{0.143\linewidth}
			\centering
			\raisebox{-0.15cm}{\includegraphics[height=1\linewidth,width=1\linewidth]{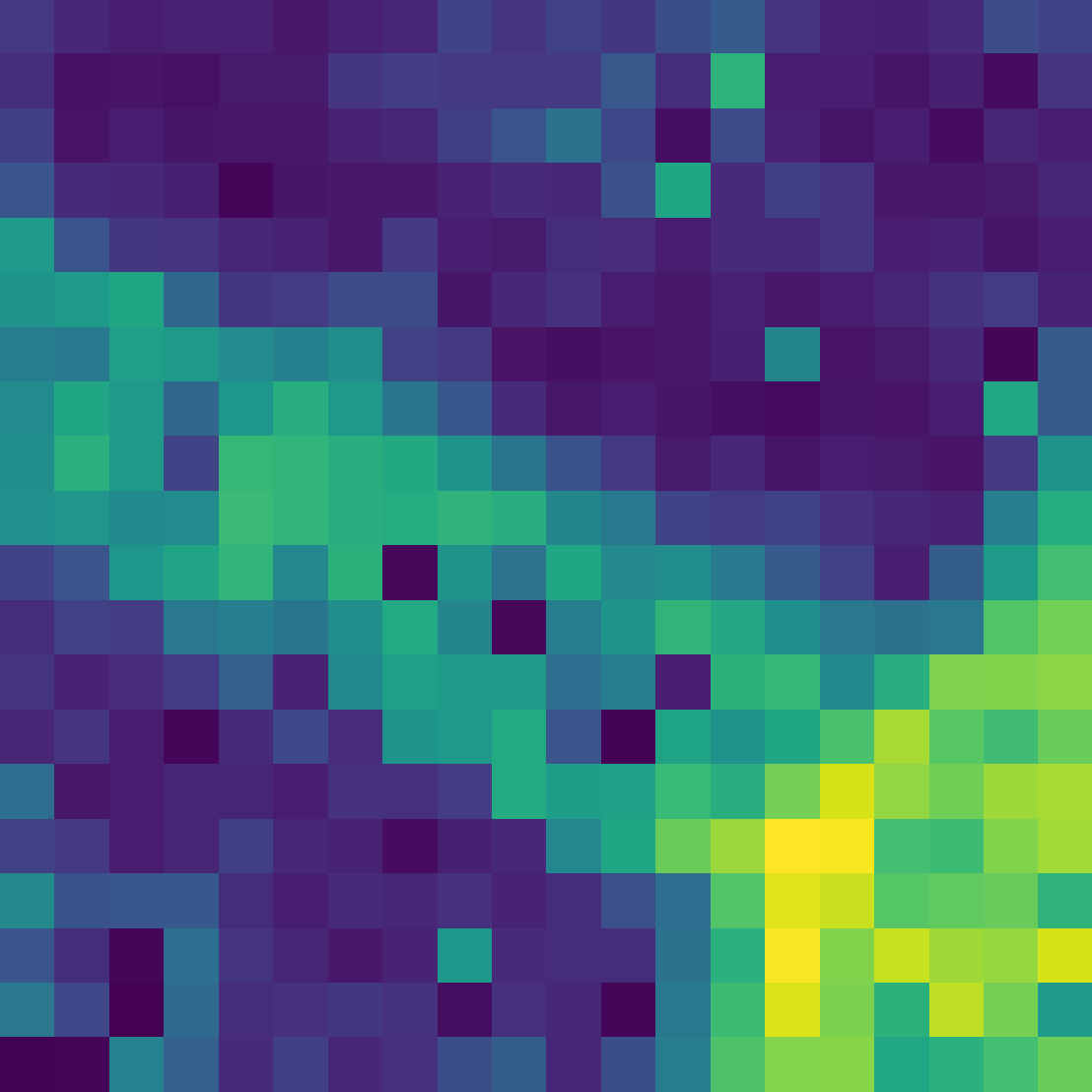}}
		\end{minipage}
		
		\begin{minipage}[t]{0.143\linewidth}
			\centering
			\raisebox{-0.15cm}{\includegraphics[height=1\linewidth,width=1\linewidth]{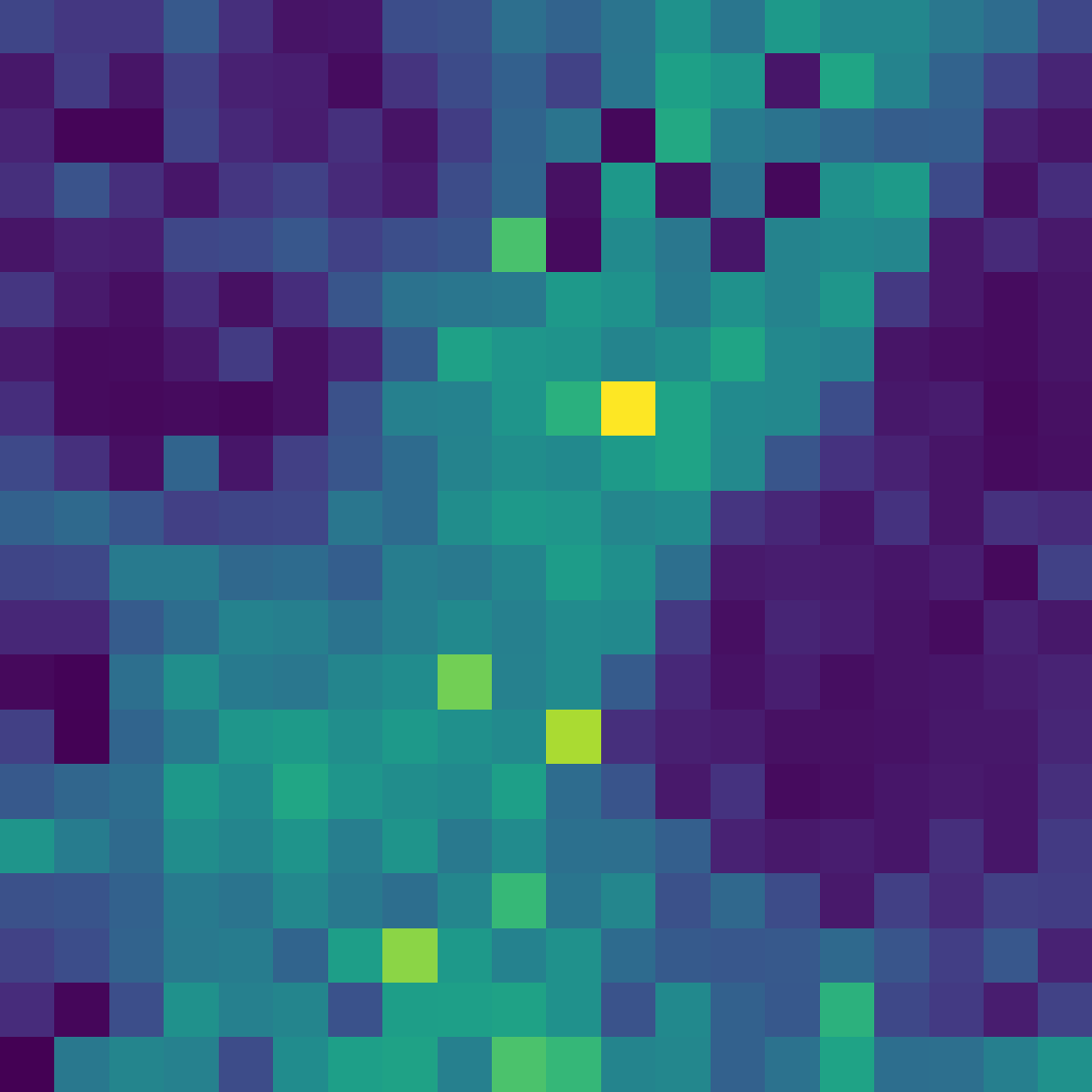}}
		\end{minipage}
	}\vspace{-3mm}

	\centering
	\subfigure{
		\begin{minipage}[t]{0.286\linewidth}
			\centering
			\raisebox{-0.15cm}{\includegraphics[height=0.5\linewidth,width=1\linewidth]{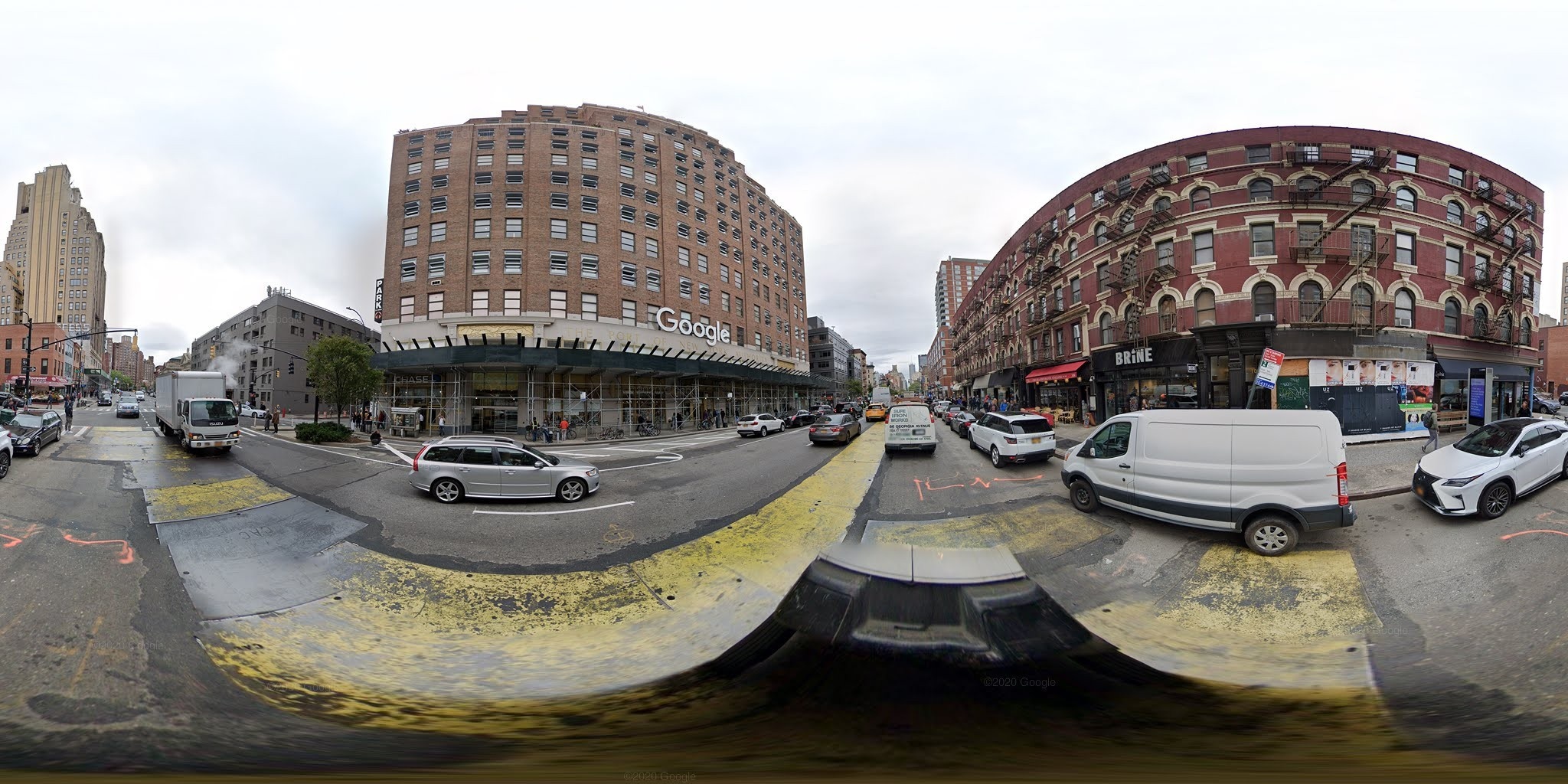}}
		\end{minipage}
		
		\begin{minipage}[t]{0.143\linewidth}
			\centering
			\raisebox{-0.15cm}{\includegraphics[height=1\linewidth,width=1\linewidth]{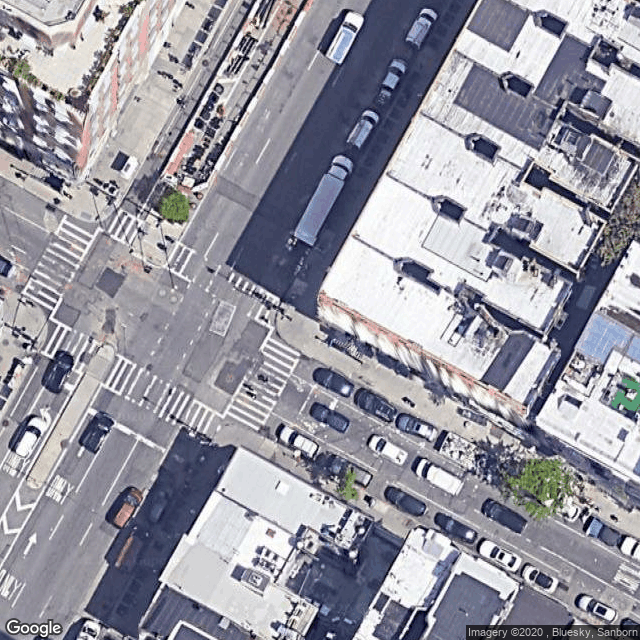}}
		\end{minipage}
		
		\begin{minipage}[t]{0.143\linewidth}
			\centering
			\raisebox{-0.15cm}{\includegraphics[height=1\linewidth,width=1\linewidth]{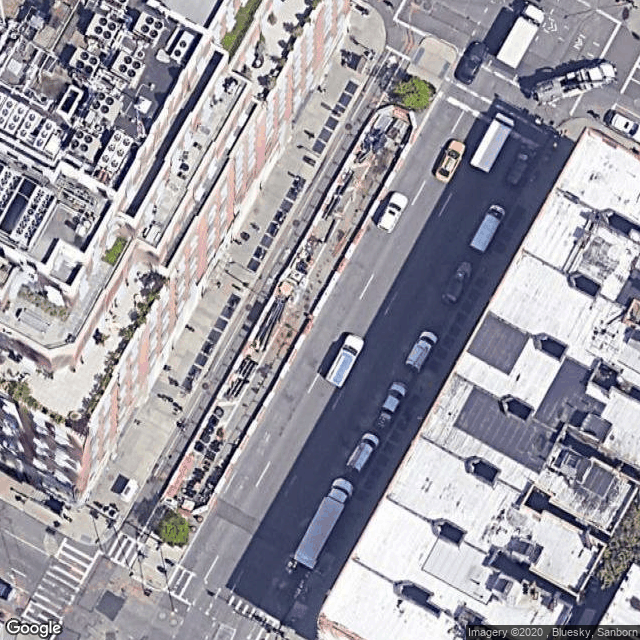}}
		\end{minipage}
		
		\begin{minipage}[t]{0.143\linewidth}
			\centering
			\raisebox{-0.15cm}{\includegraphics[height=1\linewidth,width=1\linewidth]{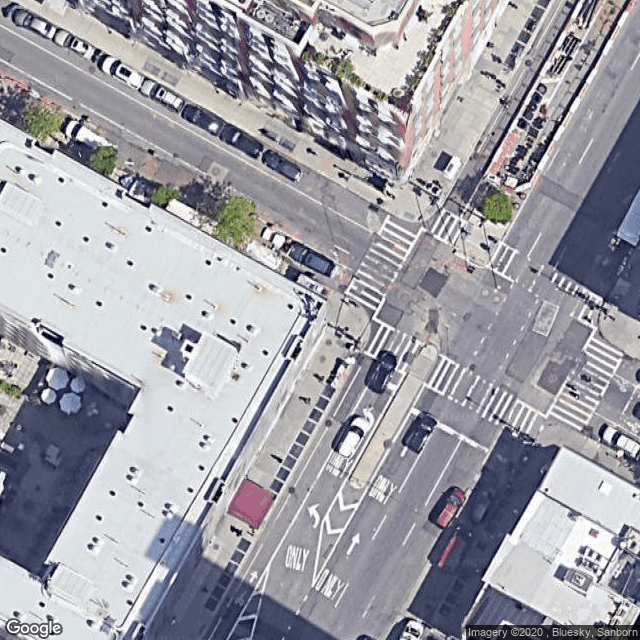}}
		\end{minipage}
		
		\begin{minipage}[t]{0.143\linewidth}
			\centering
			\raisebox{-0.15cm}{\includegraphics[height=1\linewidth,width=1\linewidth]{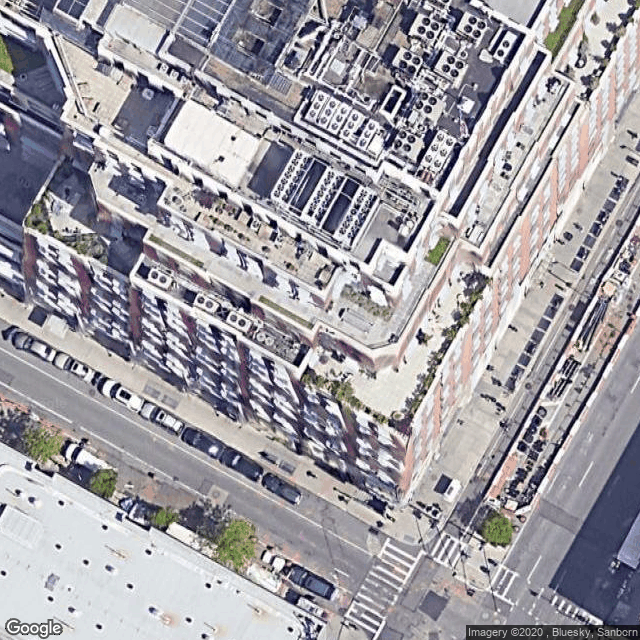}}
		\end{minipage}
	}\vspace{-3mm}
	\centering
	\subfigure{
		\begin{minipage}[t]{0.286\linewidth}
			\centering
			\raisebox{-0.15cm}{\includegraphics[height=0.5\linewidth,width=1\linewidth]{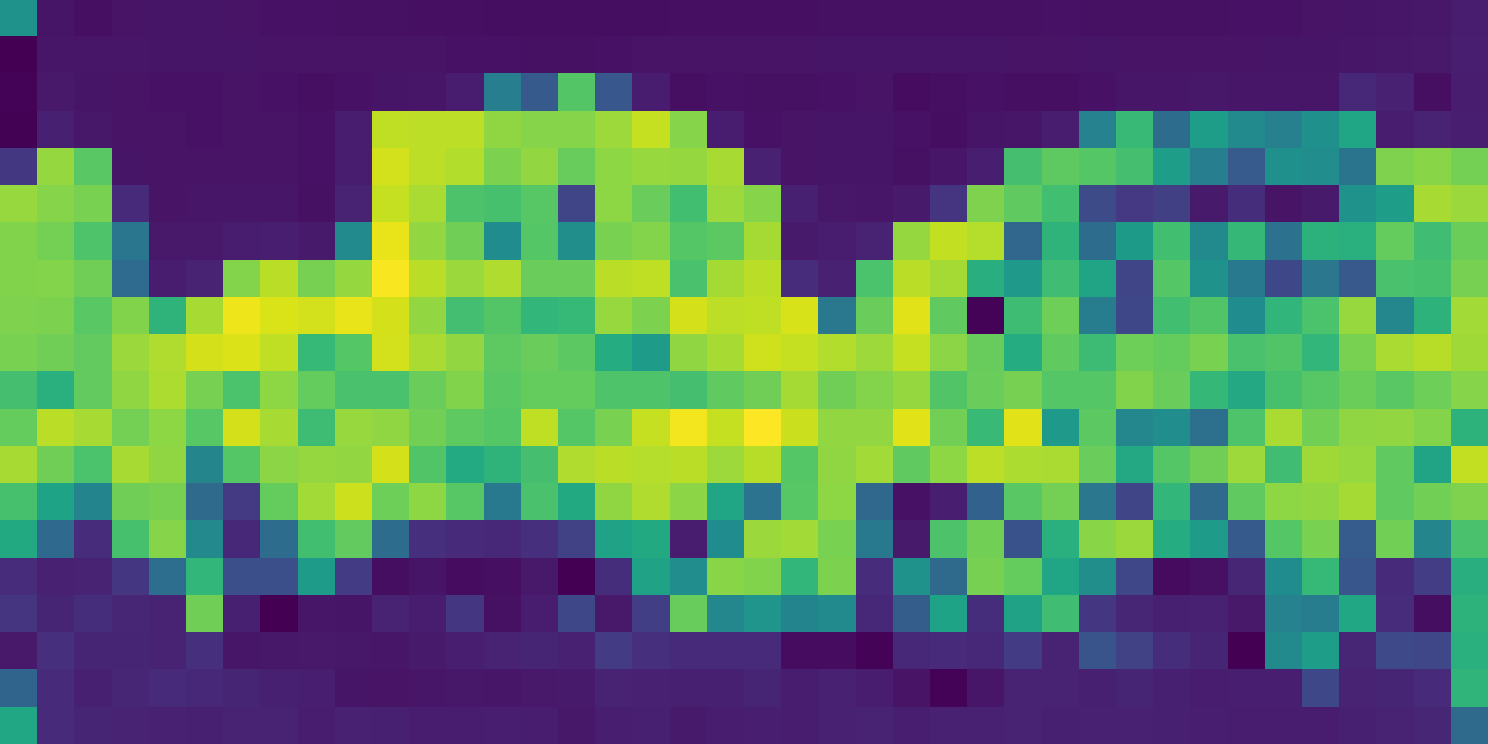}}
		\end{minipage}
		
		\begin{minipage}[t]{0.143\linewidth}
			\centering
			\raisebox{-0.15cm}{\includegraphics[height=1\linewidth,width=1\linewidth]{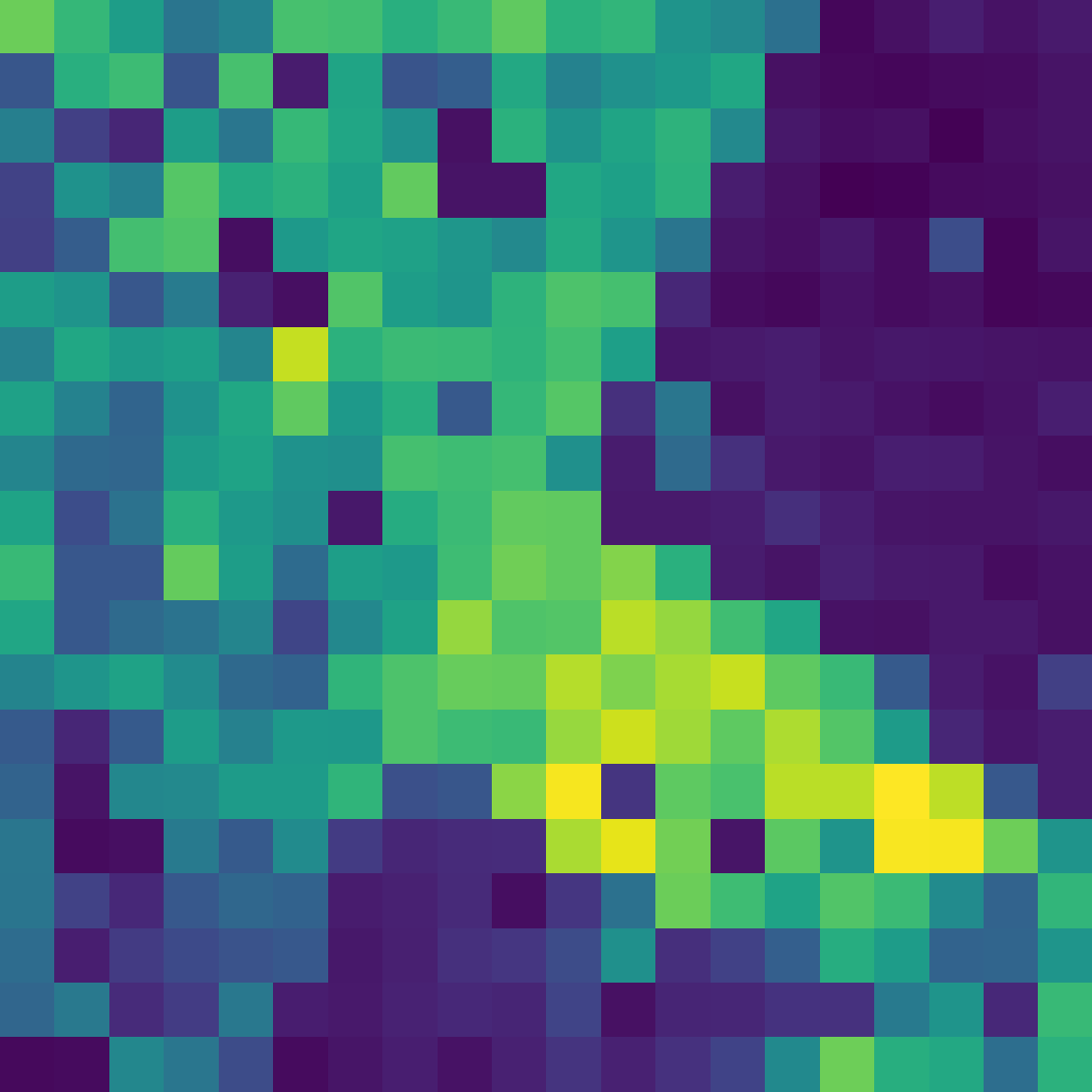}}
		\end{minipage}
		
		\begin{minipage}[t]{0.143\linewidth}
			\centering
			\raisebox{-0.15cm}{\includegraphics[height=1\linewidth,width=1\linewidth]{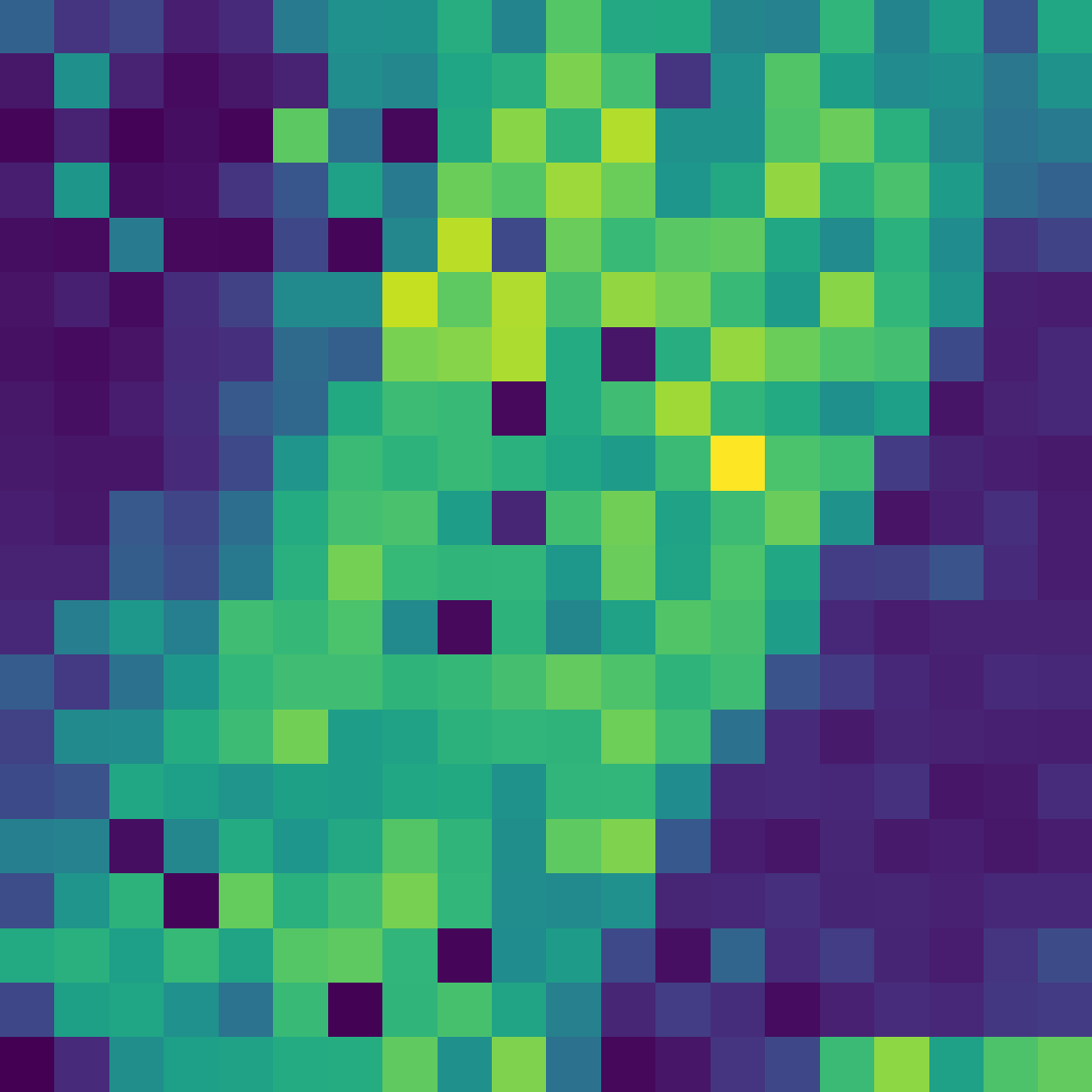}}
		\end{minipage}
		
		\begin{minipage}[t]{0.143\linewidth}
			\centering
			\raisebox{-0.15cm}{\includegraphics[height=1\linewidth,width=1\linewidth]{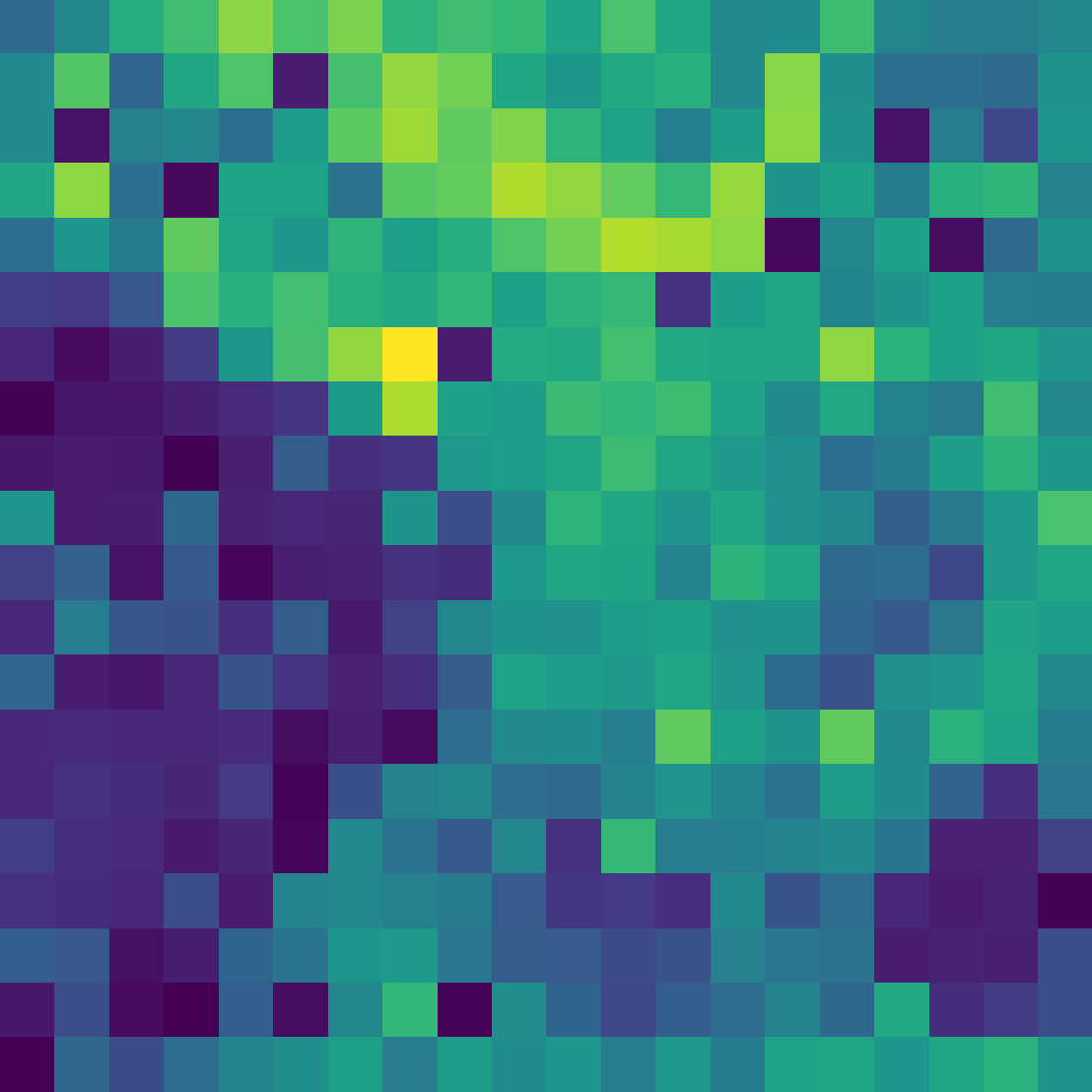}}
		\end{minipage}
		
		\begin{minipage}[t]{0.143\linewidth}
			\centering
			\raisebox{-0.15cm}{\includegraphics[height=1\linewidth,width=1\linewidth]{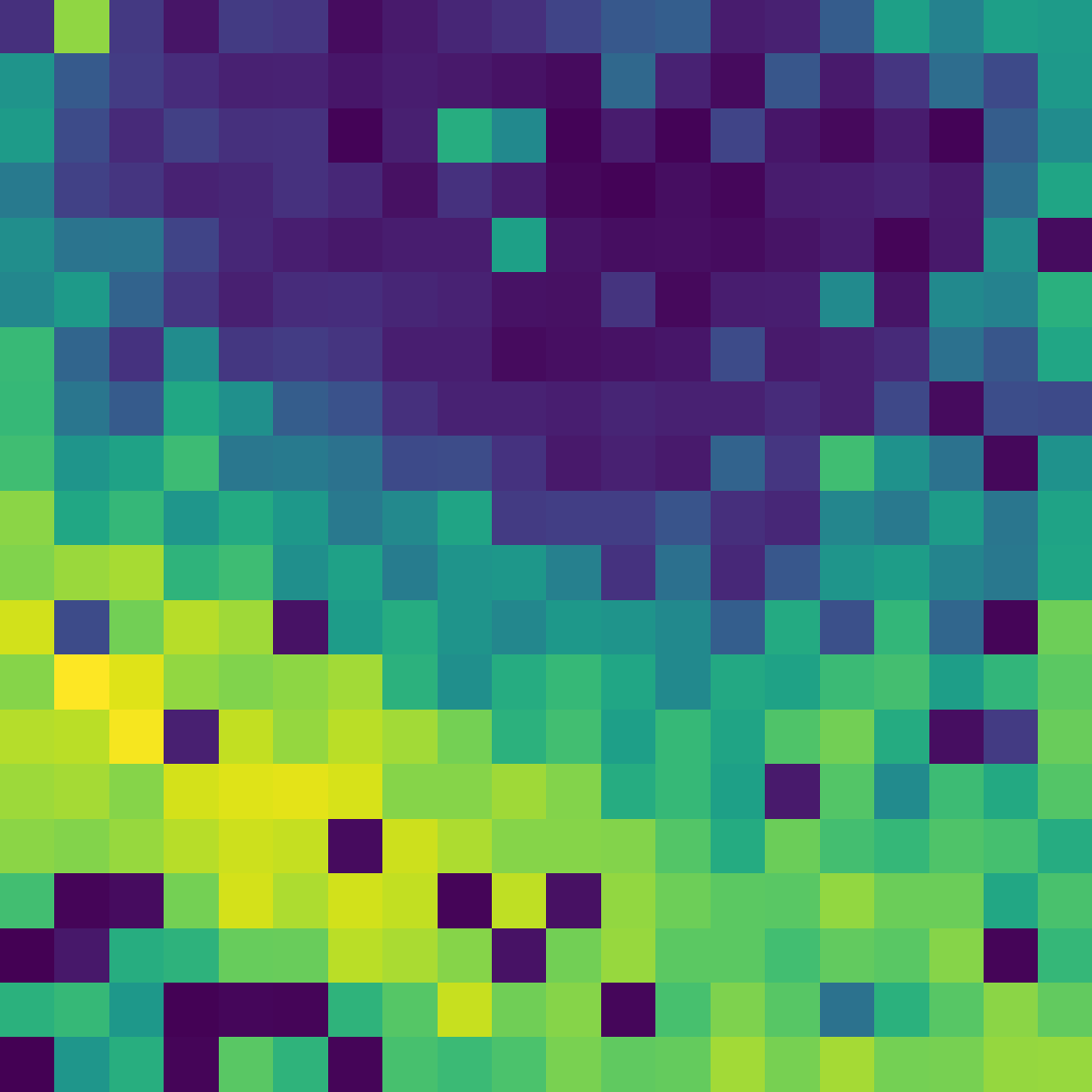}}
		\end{minipage}
	}\vspace{-3mm}
	
	\centering
	\subfigure{
		\begin{minipage}[t]{0.286\linewidth}
			\centering
			\raisebox{-0.15cm}{\includegraphics[height=0.5\linewidth,width=1\linewidth]{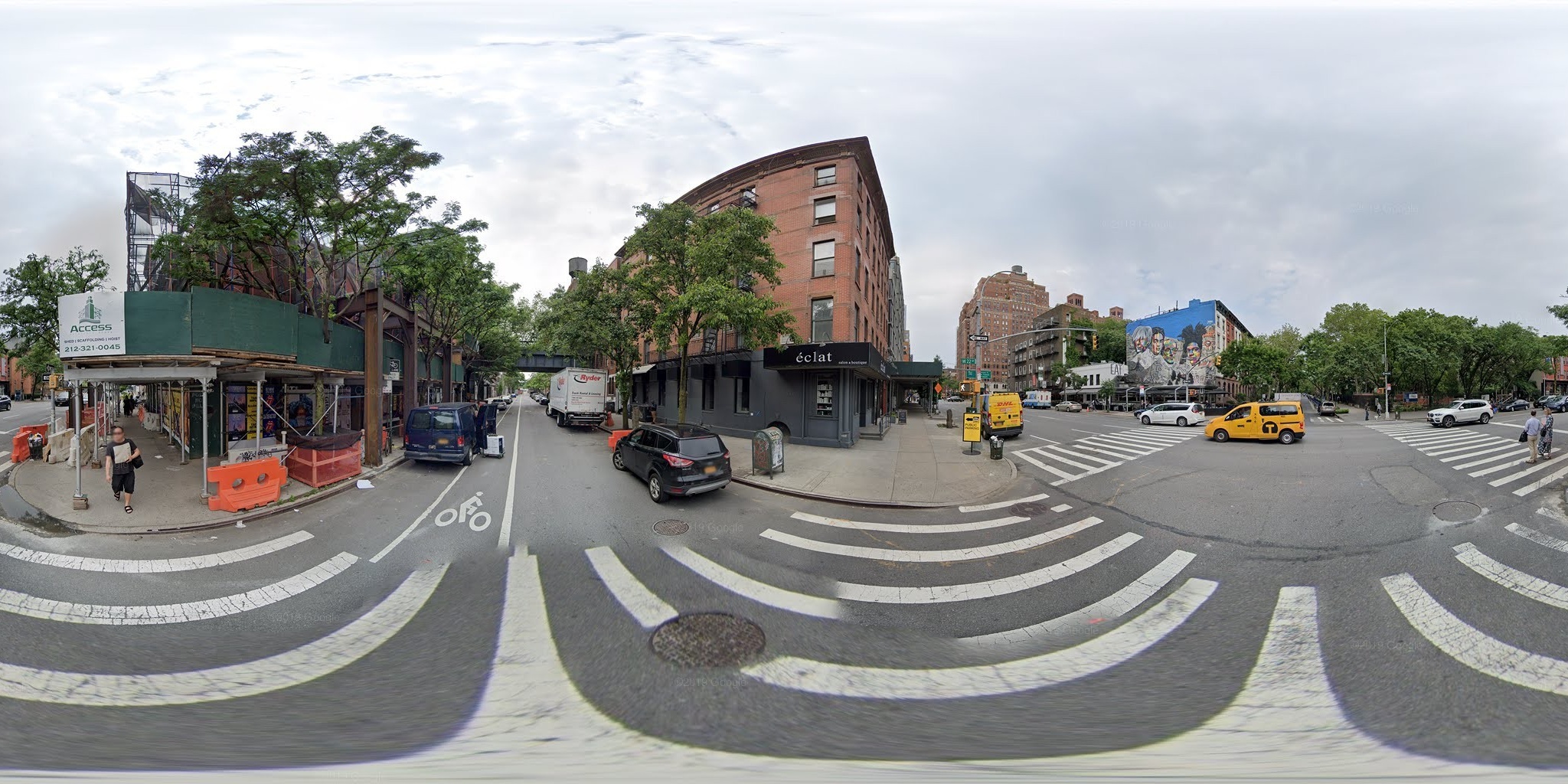}}
		\end{minipage}
		
		\begin{minipage}[t]{0.143\linewidth}
			\centering
			\raisebox{-0.15cm}{\includegraphics[height=1\linewidth,width=1\linewidth]{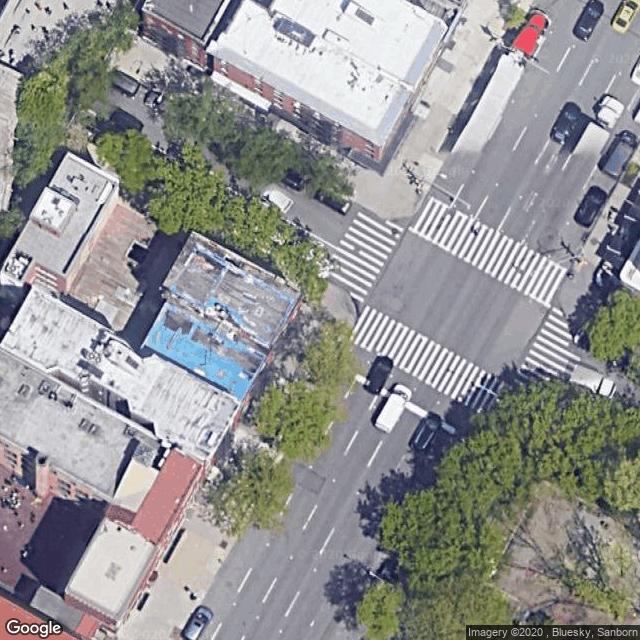}}
		\end{minipage}
		
		\begin{minipage}[t]{0.143\linewidth}
			\centering
			\raisebox{-0.15cm}{\includegraphics[height=1\linewidth,width=1\linewidth]{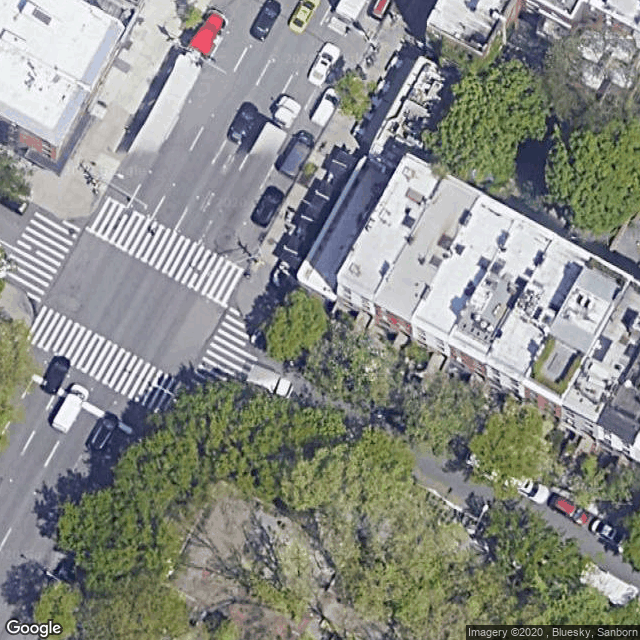}}
		\end{minipage}
		
		\begin{minipage}[t]{0.143\linewidth}
			\centering
			\raisebox{-0.15cm}{\includegraphics[height=1\linewidth,width=1\linewidth]{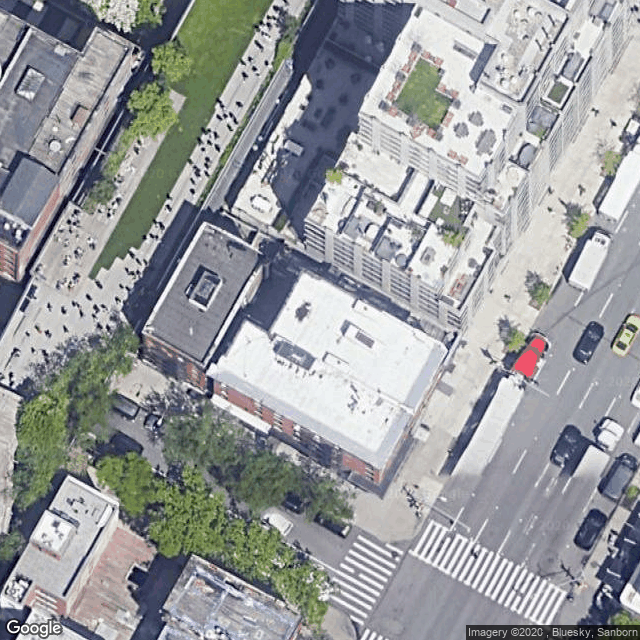}}
		\end{minipage}
		
		\begin{minipage}[t]{0.143\linewidth}
			\centering
			\raisebox{-0.15cm}{\includegraphics[height=1\linewidth,width=1\linewidth]{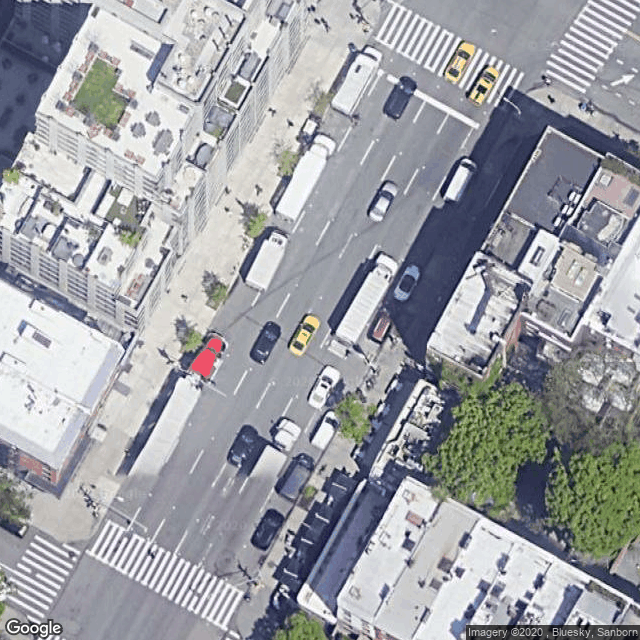}}
		\end{minipage}
	}\vspace{-3mm}
	\centering
	\subfigure{
		\begin{minipage}[t]{0.286\linewidth}
			\centering
			\raisebox{-0.15cm}{\includegraphics[height=0.5\linewidth,width=1\linewidth]{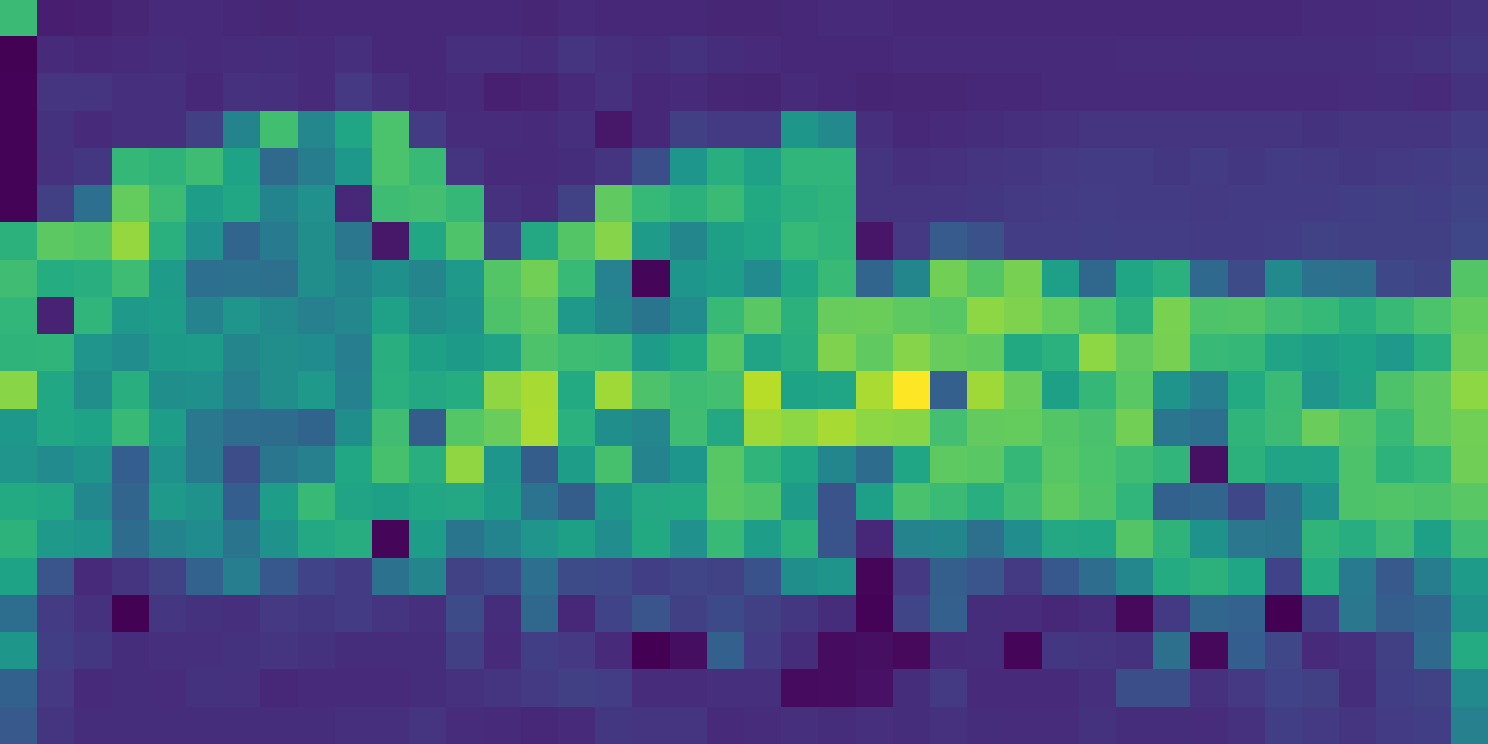}}
		\end{minipage}
		
		\begin{minipage}[t]{0.143\linewidth}
			\centering
			\raisebox{-0.15cm}{\includegraphics[height=1\linewidth,width=1\linewidth]{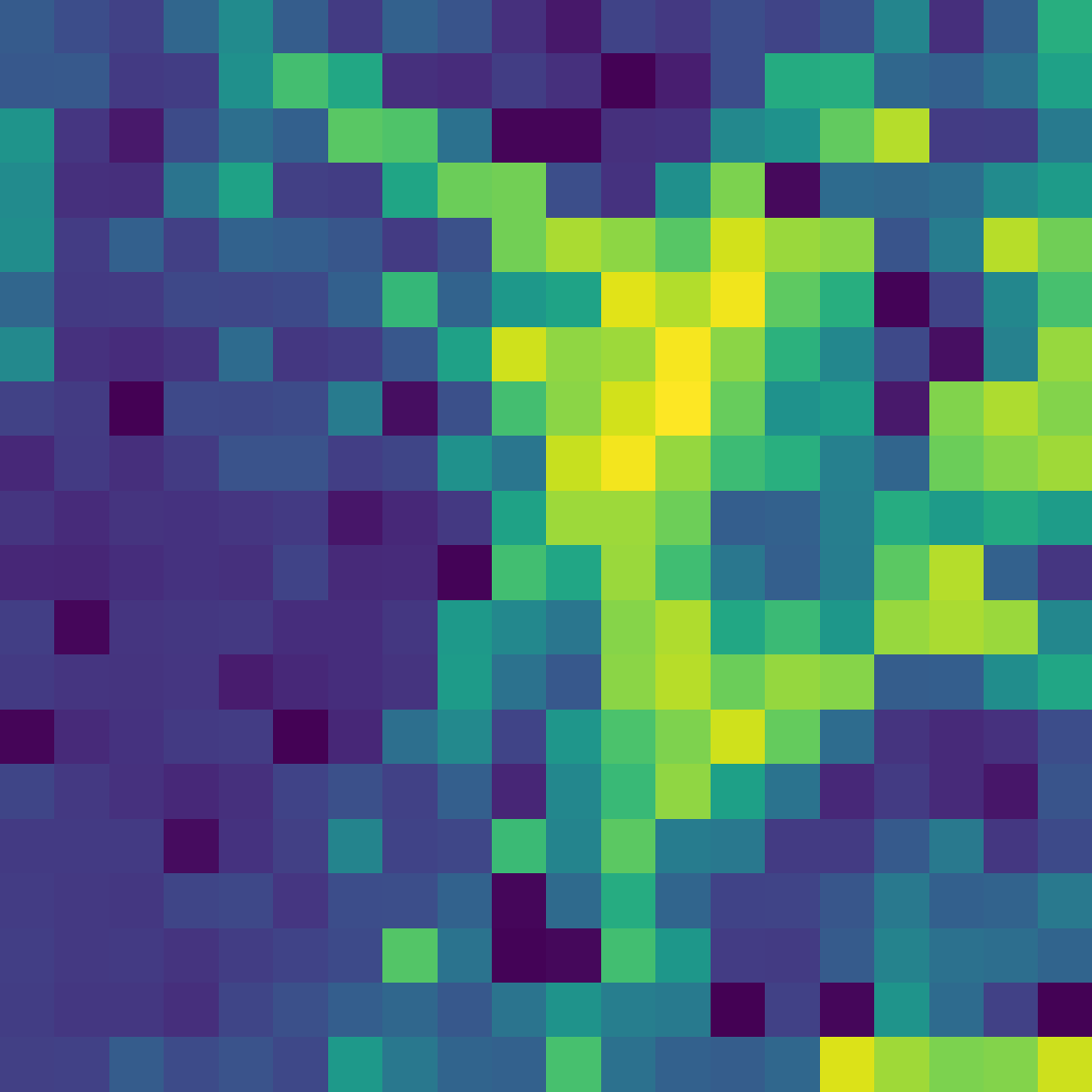}}
		\end{minipage}
		
		\begin{minipage}[t]{0.143\linewidth}
			\centering
			\raisebox{-0.15cm}{\includegraphics[height=1\linewidth,width=1\linewidth]{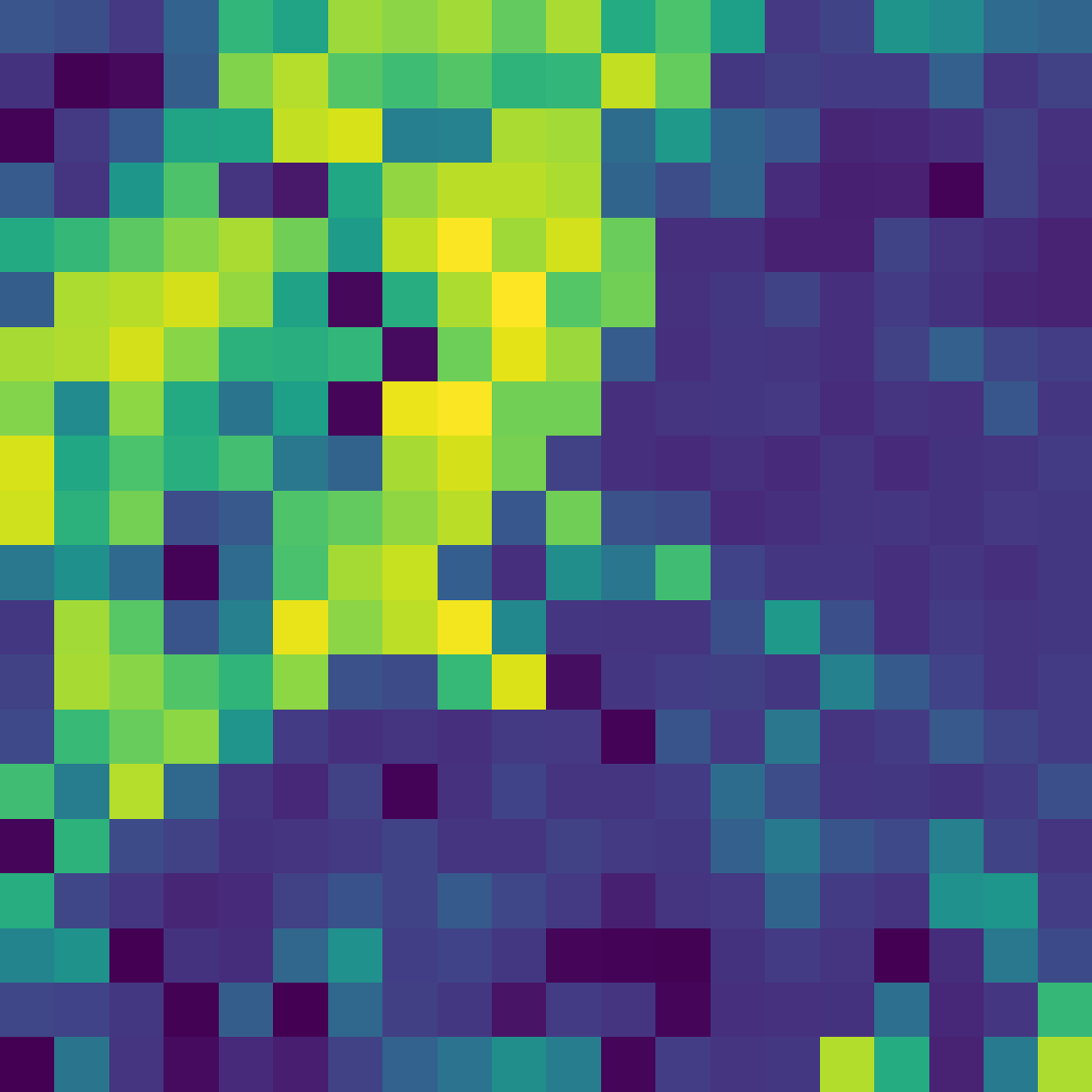}}
		\end{minipage}
		
		\begin{minipage}[t]{0.143\linewidth}
			\centering
			\raisebox{-0.15cm}{\includegraphics[height=1\linewidth,width=1\linewidth]{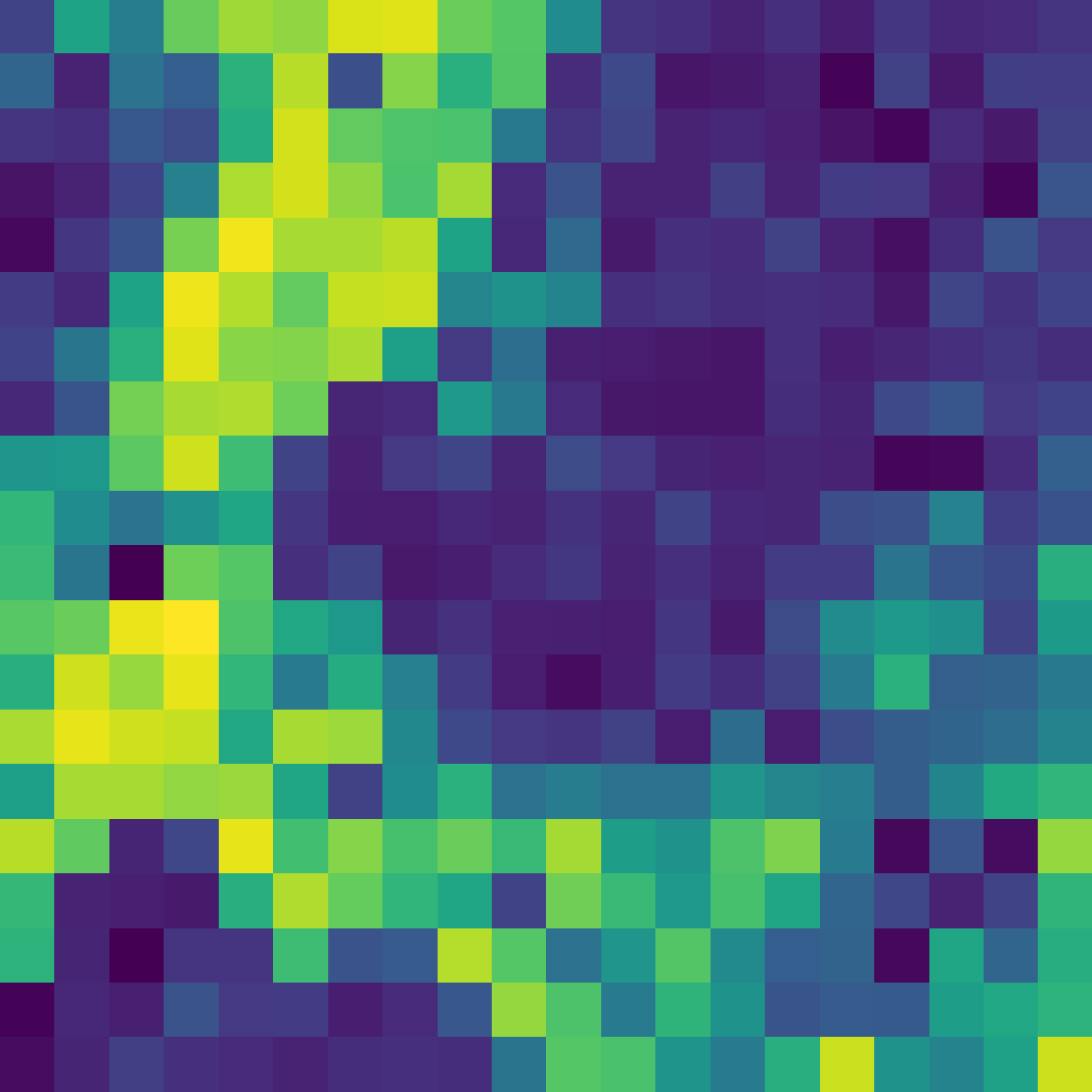}}
		\end{minipage}
		
		\begin{minipage}[t]{0.143\linewidth}
			\centering
			\raisebox{-0.15cm}{\includegraphics[height=1\linewidth,width=1\linewidth]{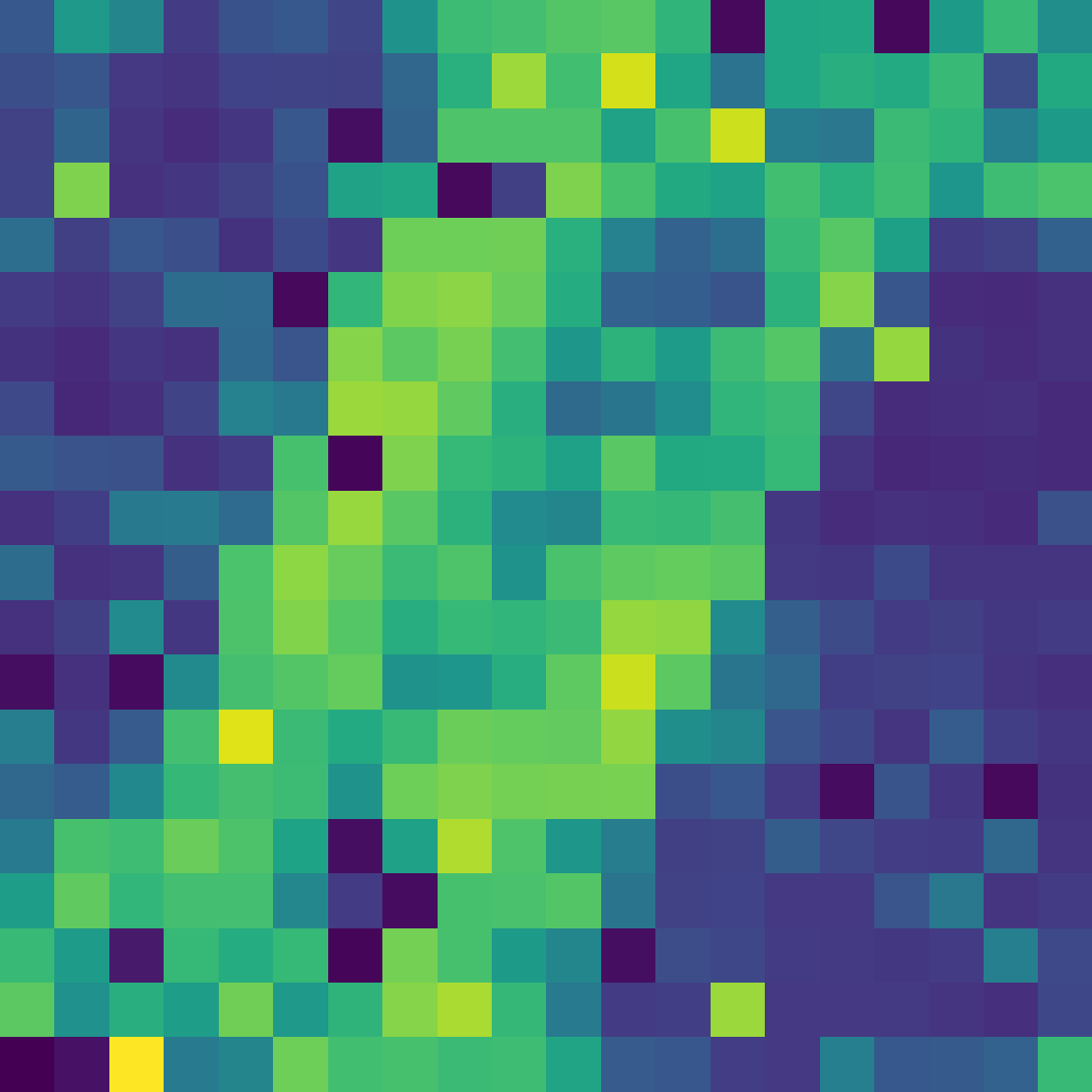}}
		\end{minipage}
	}\vspace{-3mm}
	
	\centering
	\subfigure{
		\begin{minipage}[t]{0.286\linewidth}
			\centering
			\raisebox{-0.15cm}{\includegraphics[height=0.5\linewidth,width=1\linewidth]{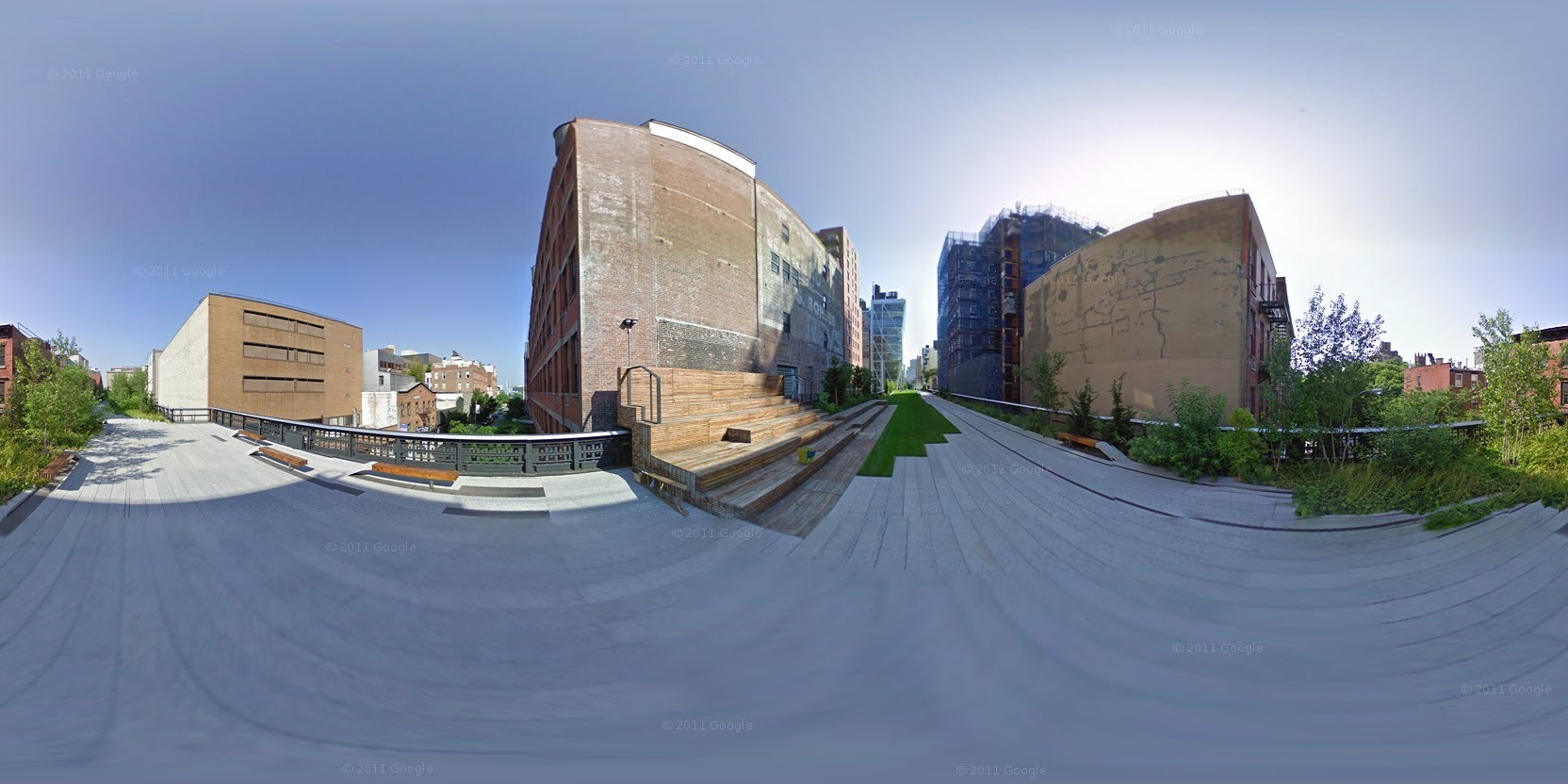}}
		\end{minipage}
		
		\begin{minipage}[t]{0.143\linewidth}
			\centering
			\raisebox{-0.15cm}{\includegraphics[height=1\linewidth,width=1\linewidth]{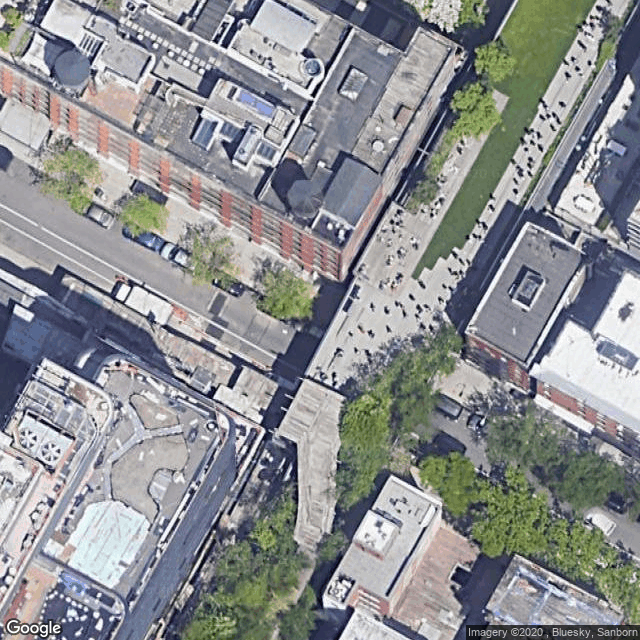}}
		\end{minipage}
		
		\begin{minipage}[t]{0.143\linewidth}
			\centering
			\raisebox{-0.15cm}{\includegraphics[height=1\linewidth,width=1\linewidth]{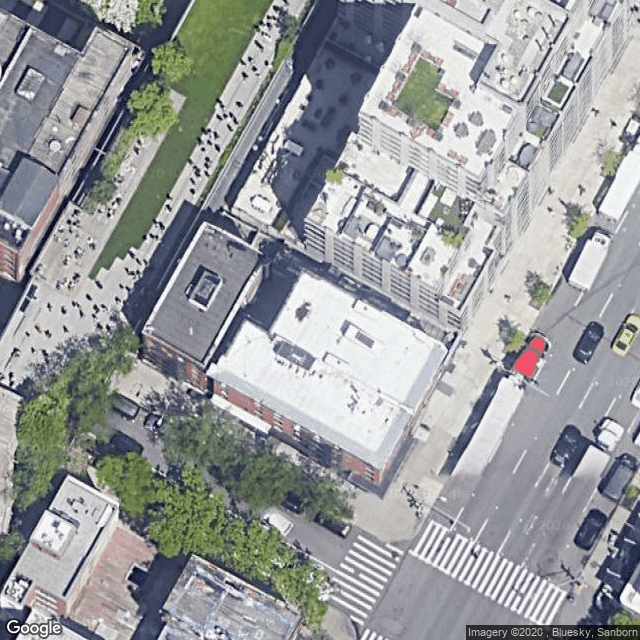}}
		\end{minipage}
		
		\begin{minipage}[t]{0.143\linewidth}
			\centering
			\raisebox{-0.15cm}{\includegraphics[height=1\linewidth,width=1\linewidth]{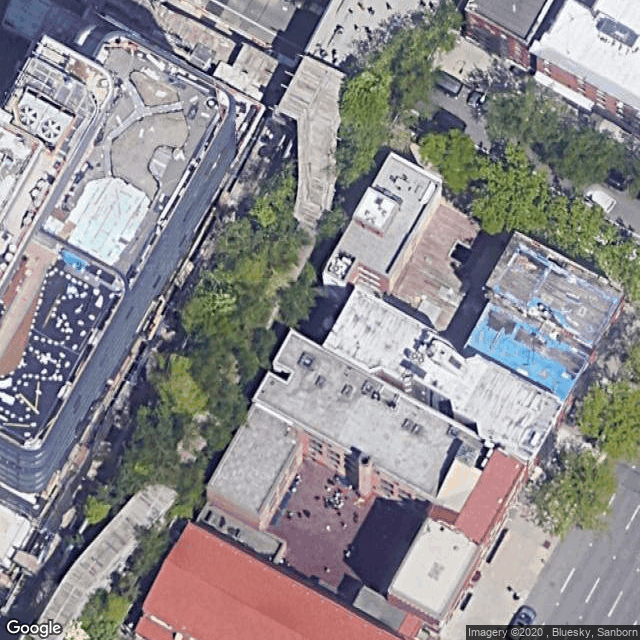}}
		\end{minipage}
		
		\begin{minipage}[t]{0.143\linewidth}
			\centering
			\raisebox{-0.15cm}{\includegraphics[height=1\linewidth,width=1\linewidth]{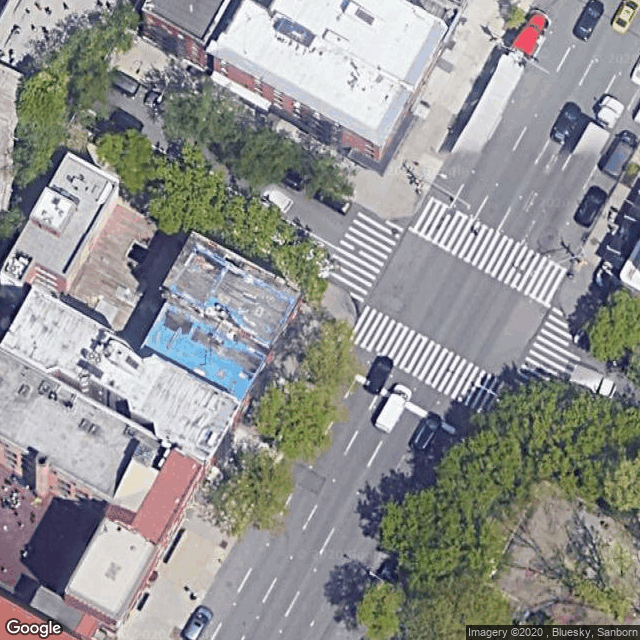}}
		\end{minipage}
	}\vspace{-3mm}
	\centering
	\subfigure{
		\begin{minipage}[t]{0.286\linewidth}
			\centering
			\raisebox{-0.15cm}{\includegraphics[height=0.5\linewidth,width=1\linewidth]{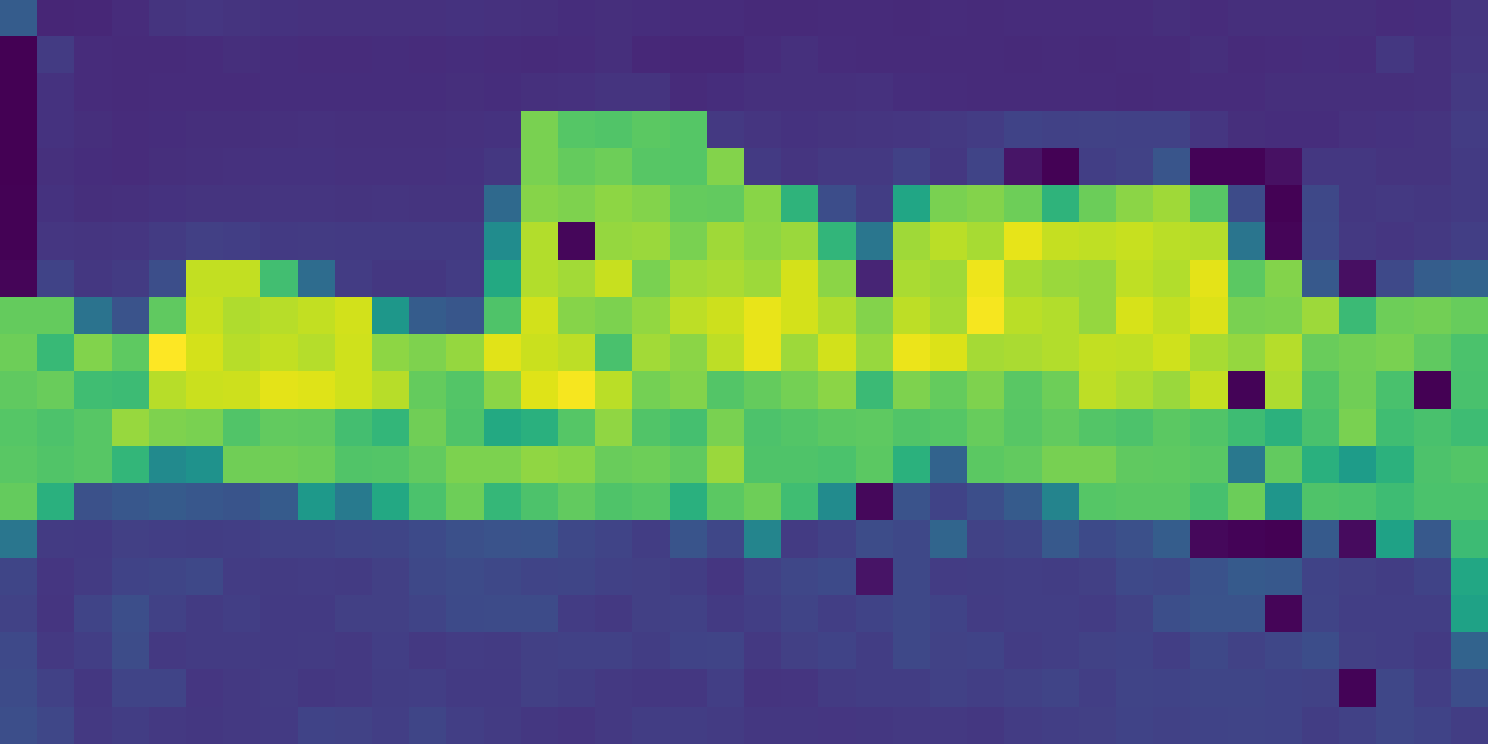}}
		\end{minipage}
		
		\begin{minipage}[t]{0.143\linewidth}
			\centering
			\raisebox{-0.15cm}{\includegraphics[height=1\linewidth,width=1\linewidth]{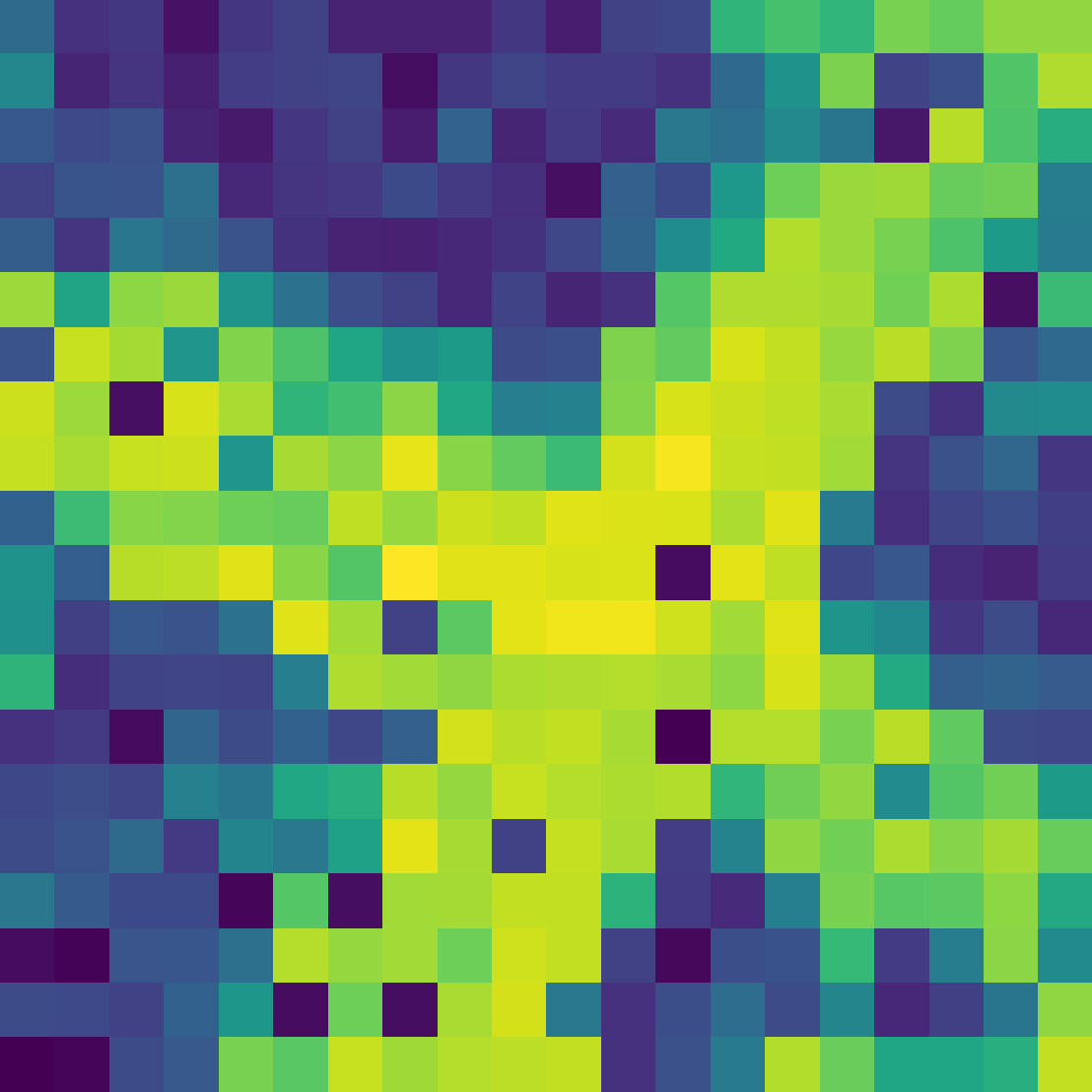}}
		\end{minipage}
		
		\begin{minipage}[t]{0.143\linewidth}
			\centering
			\raisebox{-0.15cm}{\includegraphics[height=1\linewidth,width=1\linewidth]{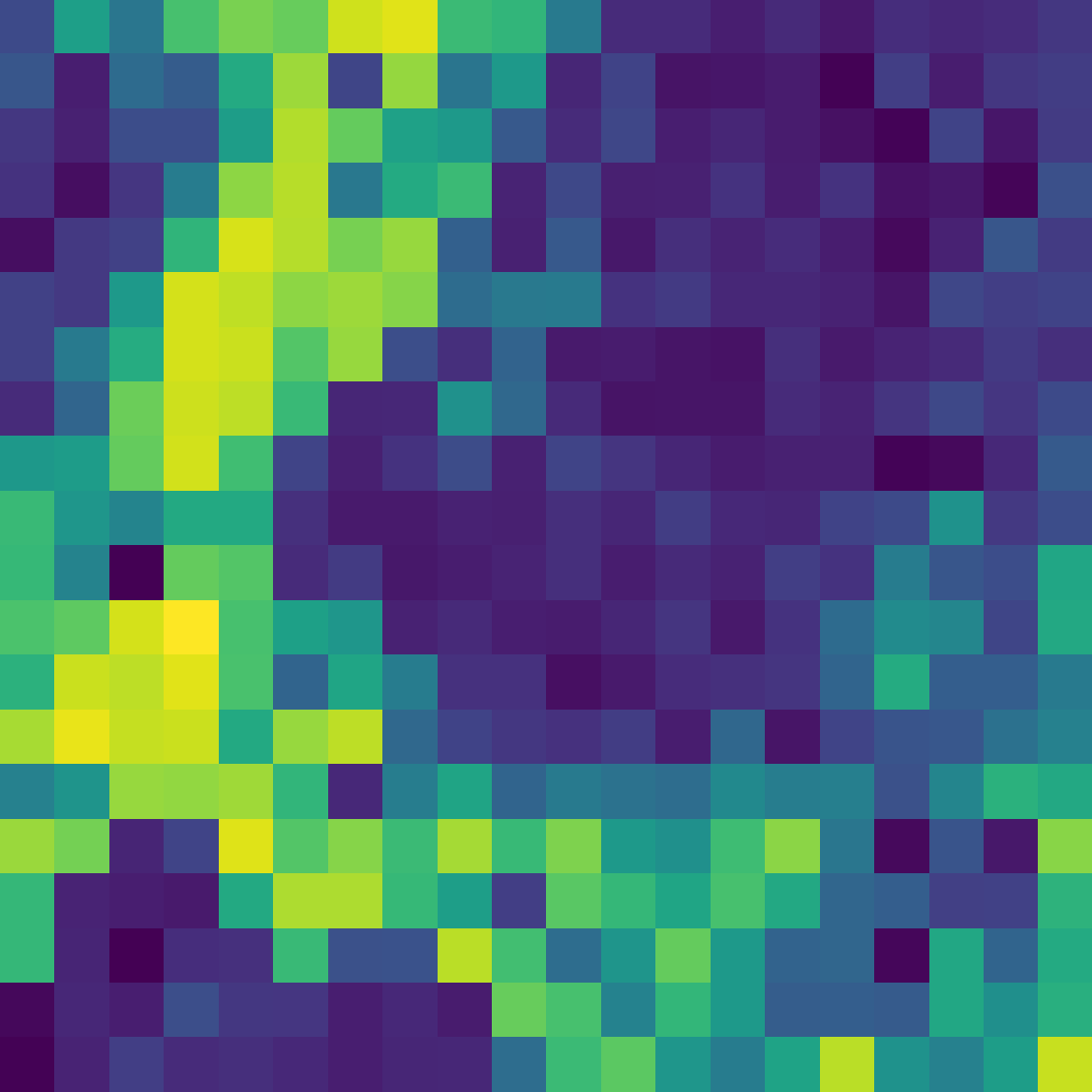}}
		\end{minipage}
		
		\begin{minipage}[t]{0.143\linewidth}
			\centering
			\raisebox{-0.15cm}{\includegraphics[height=1\linewidth,width=1\linewidth]{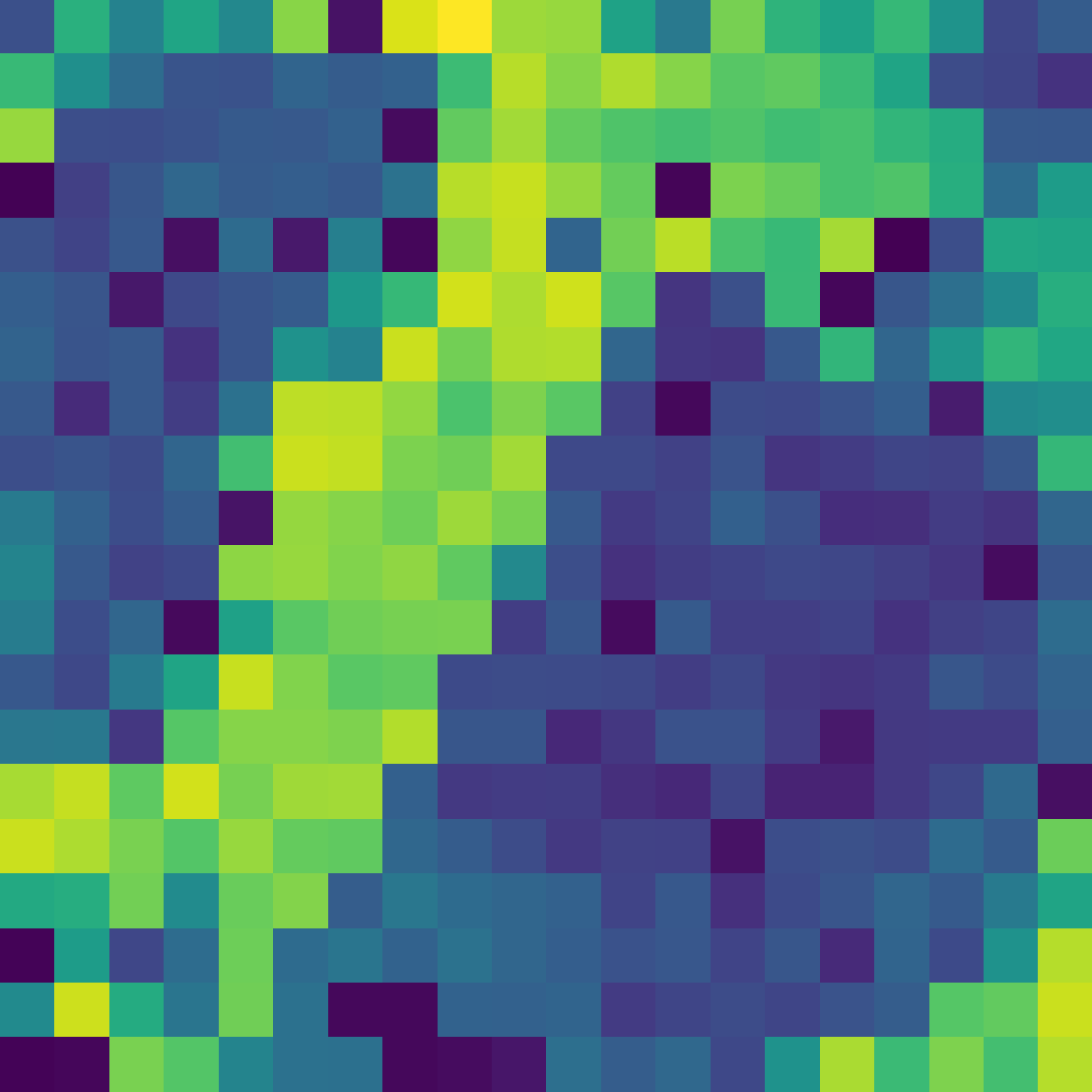}}
		\end{minipage}
		
		\begin{minipage}[t]{0.143\linewidth}
			\centering
			\raisebox{-0.15cm}{\includegraphics[height=1\linewidth,width=1\linewidth]{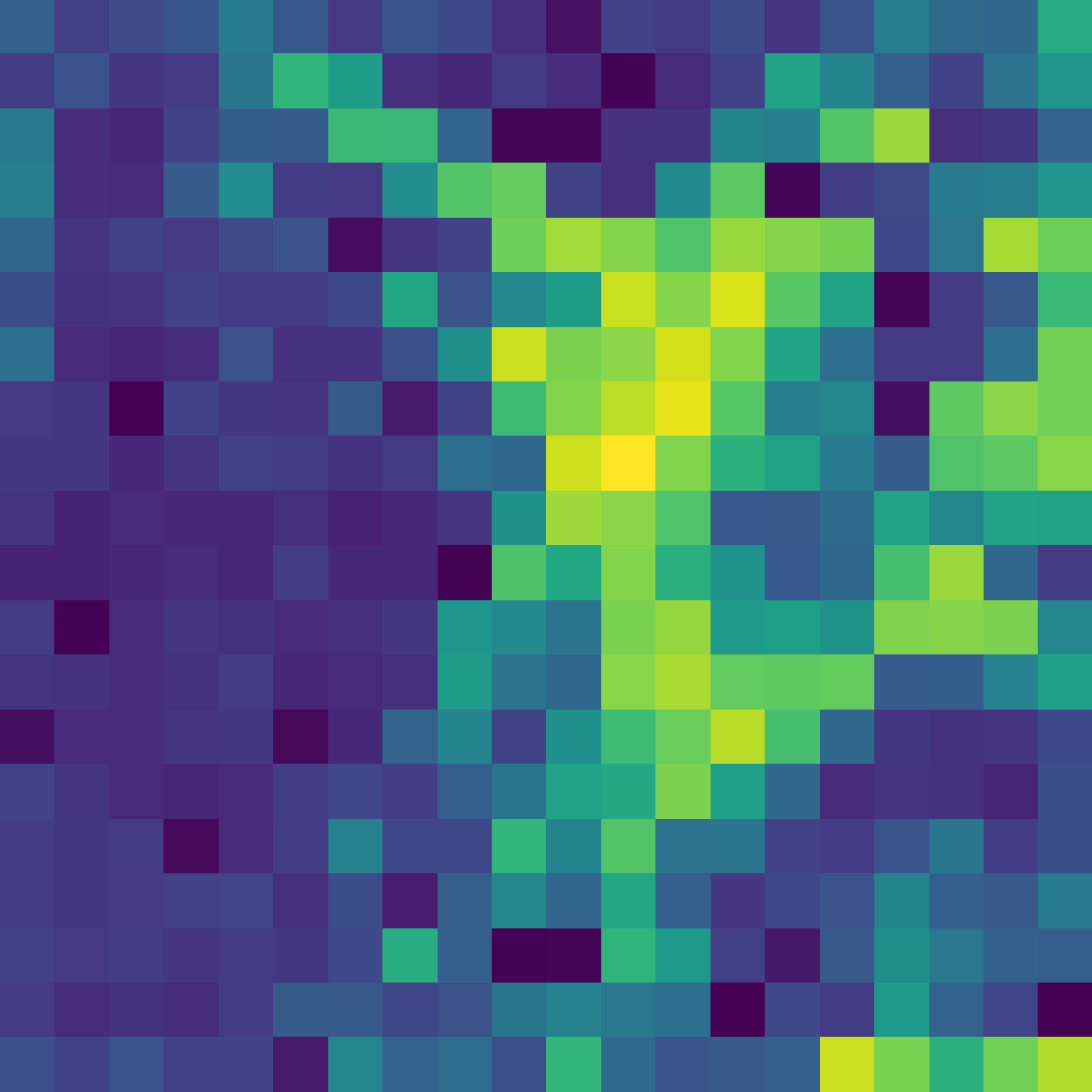}}
		\end{minipage}
	}
	
	\caption{\textbf{Visualization of heatmaps on VIGOR with SAIG-D.} Each row contains a ground view image, a positive aerial image and three semi-positive aerial images. All the images are from VIGOR-test that SAIG has not seen at training. } 
	\label{heatmap in VIGOR}
\end{figure*}

Figure \ref{heatmap in VIGOR} shows the heatmap on VIGOR. The first column is the ground view image and the second column is positive aerial view image. Moreover, the semi-positive aerial view images are also presented from the third to the fifth columns. The corresponding heatmaps are given below each row of images. It can be seen in these heatmaps that SAIG focuses very much on the roads as well as the corner points of buildings. In addition, some supervised information provided by plants and forest is potentially beneficial for retrieval. Since the non-ortho images has a slanted camera angle, some surface texture information from the buildings may also be learned by the network. 
Note that on VIGOR, the semi-positive aerial images correlate to multiple ground view image. For example, there are duplicate aerial images in rows 5 and 7, but the models may prefer a particular region and ignore those in other positive samples.

\begin{figure*}
    
        \centering
        \subfigure[CVUSA]{
        \label{Retrieval examples:A}
        \includegraphics[width=0.9\textwidth]{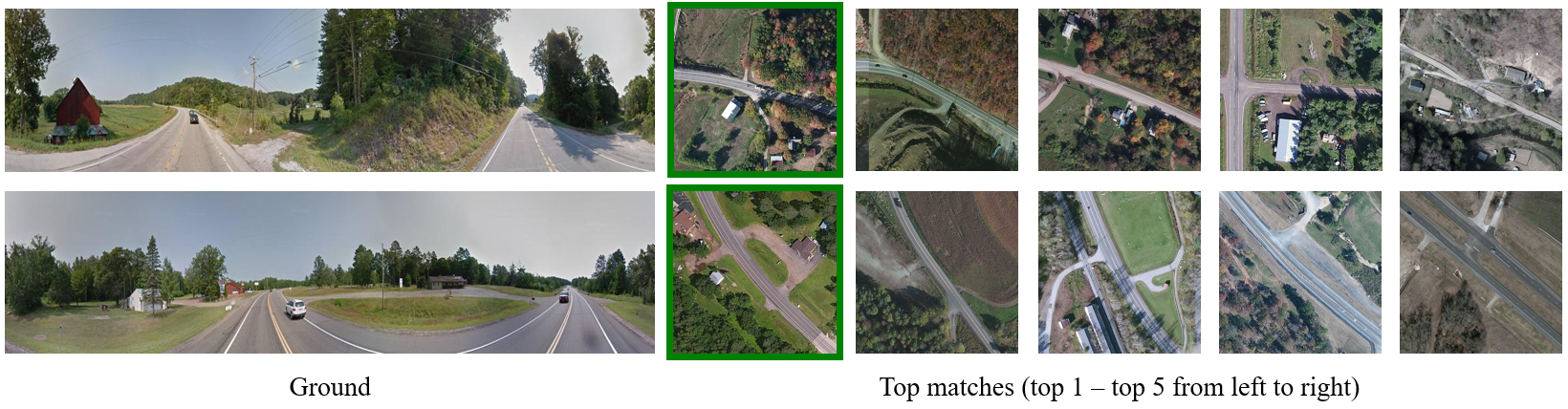} 
        }\vspace{-3mm}
        
        \centering
        \subfigure[CVACT]{
        \label{Retrieval examples:B}
        \includegraphics[width=0.9\textwidth]{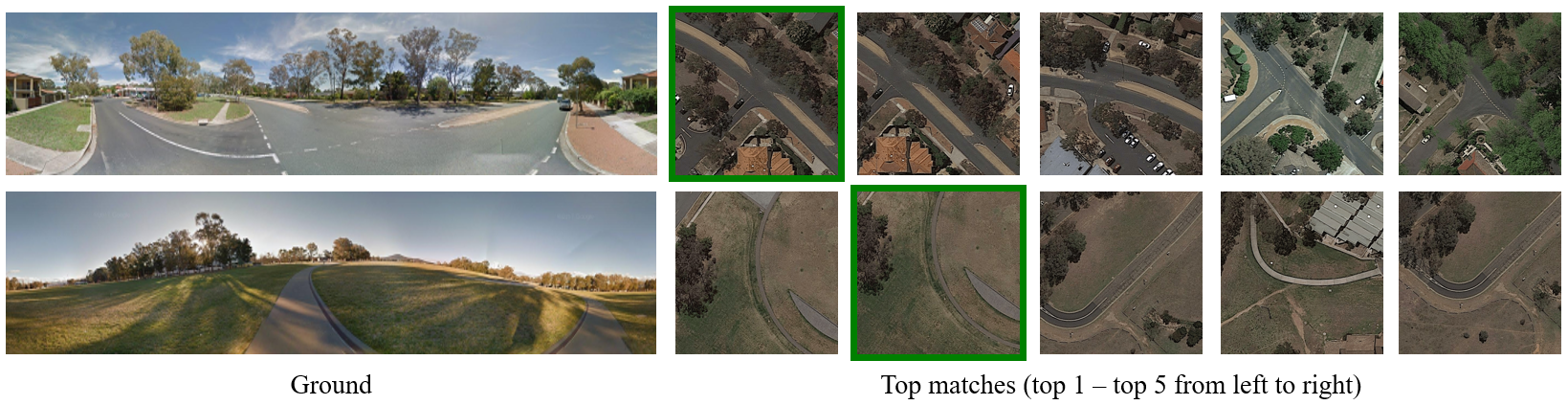} 
        }\vspace{-3mm}
        
        \centering
        \subfigure[VIGOR]{
        \label{Retrieval examples:C}
        \includegraphics[width=0.9\textwidth]{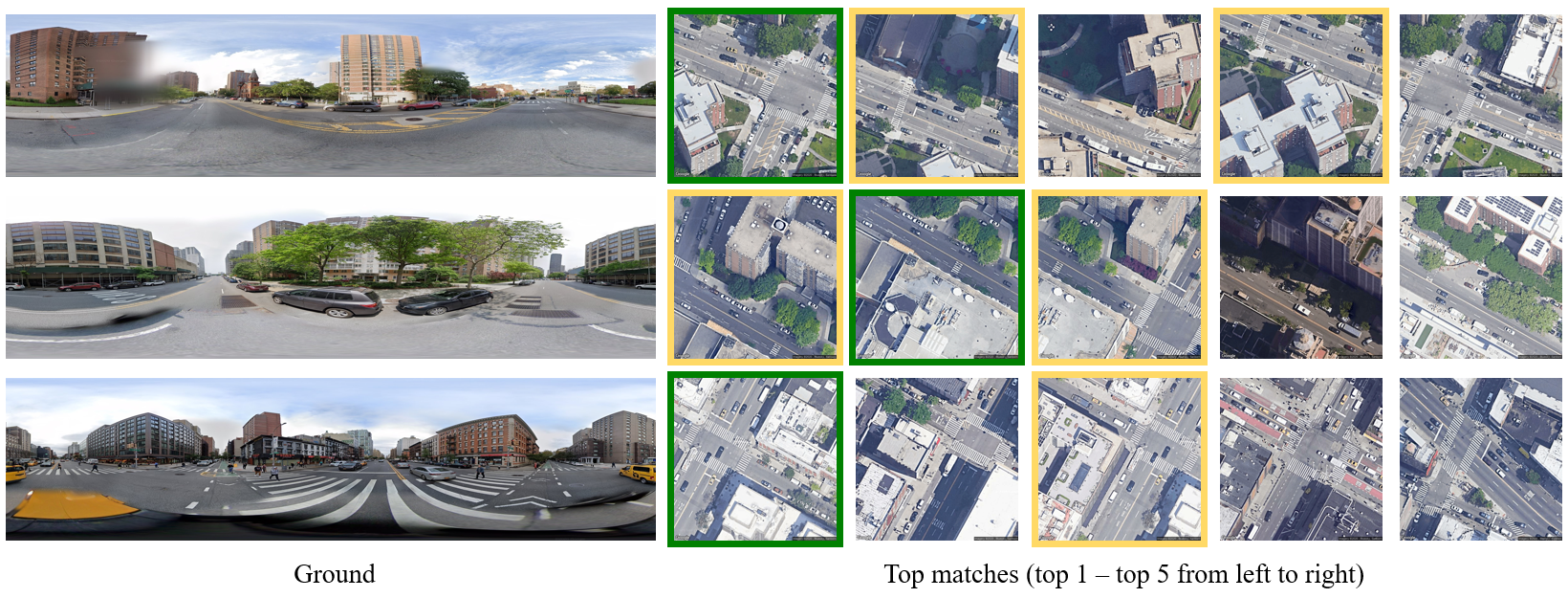} 
        }\vspace{-3mm}
        
        \centering
        \subfigure[University-1652]{
        \label{Retrieval examples:D}
        \includegraphics[width=0.9\textwidth]{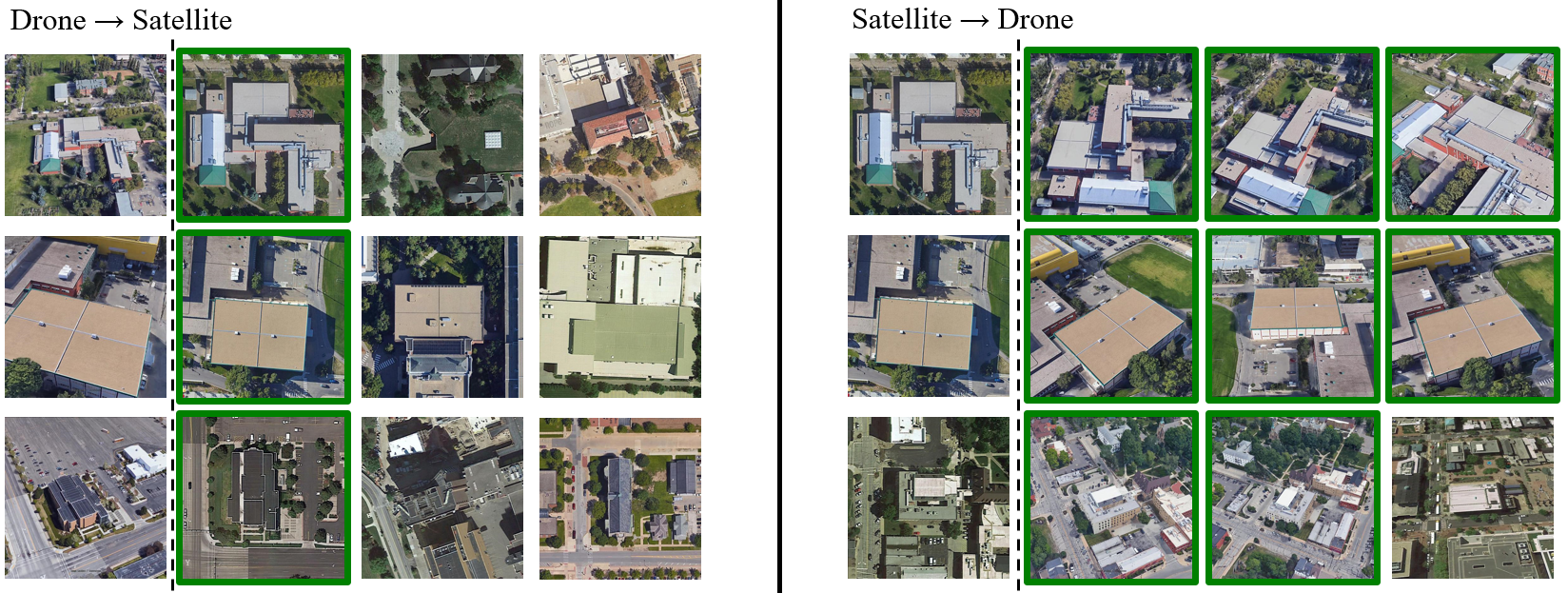} 
        }\vspace{-3mm}
        \subfigure{\includegraphics[width=0.7\textwidth]{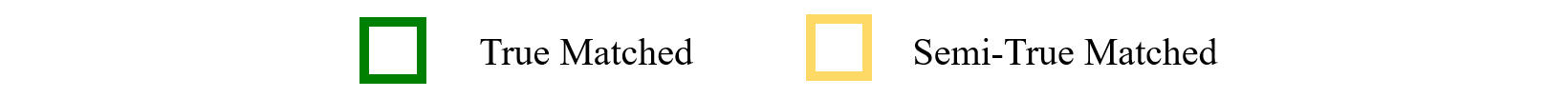}}
        \DeclareGraphicsExtensions.
        
        \caption{Image retrieval examples on (a) CVUSA, (b) CVACT, (c) VIGOR and (d) University-1652.}
\label{Retrieval examples}

\end{figure*}

\bmhead{Examples of Retrieval}We present top 5 retrieval results using SAIG-D backbone with 384-d features on CVUSA, CVACT, VIGOR, and University-1652 datasets in Figure \ref{Retrieval examples}. The ground-truth results are in the green border, and semi-positive samples are in the yellow border. From these examples, we can observe that SAIG performs well across four datasets, including the remote countryside and urban areas. Although the ground and aerial image pairs show entirely different views, SAIG still retrieve the corresponding images based on the content and global correspondence in the images. As observed in Figure \ref{Retrieval examples:C}, although we ignore the semi-positive samples at training, SAIG succeeds in assigning high rankings to these semi-positive samples. It suggests that SAIG can mine images that are similar in content. 
As shown in the right of Figure \ref{Retrieval examples:D}, given the satellite-view query image, these drone-view images depict the same place are retrieved by our SAIG backbone for the drone navigation tasks, even if these images with different flight altitudes and angles.
We show the top-3 retrieval result, where the true matched drone view images are highlighted by the green boxes. Regarding the mis-retrieved result like the top-1 image in the second row of Figure \ref{Retrieval examples:B}, we argue this is due to the fact that there are very subtle difference between the retrieved aerial image and the ground truth, such that our backbone confuses them. To summarize, these above retrieval results from different kinds of datasets reflect the effectiveness and generalization of our SAIG backbone.

\section{Conclusion}
Previous cross-view geo-localization works usually exploit feature aggregation layers attached to CNN networks relying on the alignment information. However, we propose and verify that a simple backbone with the ability of global information modeling is superior for generalization. It can achieve favorable performance without leveraging the alignment information. Specifically, we introduce overlapping patchify features by employing the convolutional stem, while capture global interactions with the Multi-head Self-attention module. In addition, we propose a new feature aggregation module that enhances the model and facilitates the convergence at training. Extensive experiments are conducted on multiple cross-view benchmarks with various settings and image retrieval datasets. The results validate the simplicity, effectiveness and generalizability of the proposed cross-view SAIG backbone.

\section*{Data Availability Statement}
Six datasets supporting the findings of this study are available with the permission of the dataset authors. The links to request these datasets are as follows.
\begin{enumerate}[(1)]
\item CVUSA: \href{https://github.com/viibridges/crossnet}{https://github.com/viibridges/crossnet}
\item CVACT: \href{https://github.com/Liumouliu/OriCNN}{https://github.com/Liumouliu/OriCNN}
\item VIGOR: \href{https://github.com/Jeff-Zilence/VIGOR}{https://github.com/Jeff-Zilence/VIGOR}
\item University-1652: \href{https://github.com/layumi/University1652-Baseline}{https://github.com/layumi/University1652-Baseline}
\item ILSVRC-2012 ImageNet: \href{https://www.image-net.org/}{https://www.image-net.org/}
\item $\mathcal{R}$Oxford and $\mathcal{R}$Paris: \href{https://github.com/filipradenovic/revisitop}{https://github.com/filipradenovic/revisitop}
\end{enumerate}


\bibliography{sn-bibliography}


\end{document}